%% file: arxiv.tex
\documentclass{article}

\usepackage{arxiv}
\usepackage{tabularx}
\usepackage[utf8]{inputenc} % allow utf-8 input
\usepackage[T1]{fontenc}    % use 8-bit T1 fonts
\usepackage[pagebackref=true,breaklinks=true,colorlinks,bookmarks=false]{hyperref}
\hypersetup{linkcolor=[rgb]{0.8,0.1,0.1}}
\hypersetup{citecolor=[rgb]{0.4,0.15,0.95}}

\usepackage{url}            
\usepackage{booktabs}       
\usepackage{amsfonts}       % blackboard math symbols
\usepackage[subfigure]{tocloft}
\usepackage{nicefrac}       % compact symbols for 1/2, etc.
\usepackage{microtype}      % microtypography
\usepackage{xcolor}         % colors

\usepackage{graphicx}      
\usepackage{subfigure}
\usepackage[toc,page,header]{appendix}
\usepackage{minitoc}
\usepackage{makecell}
\usepackage{adjustbox}
\usepackage{colortbl}
\definecolor{Gray}{gray}{0.92}
\usepackage{amsmath}
\usepackage{amssymb}
\usepackage{mathtools}
\usepackage{amsthm}
\usepackage{multirow}
\usepackage{wrapfig}
\usepackage{enumitem}

\newcommand{\x}[0]{\textbf{x}}

\DeclareMathOperator*{\argmin}{arg\,min}

\newcommand{\ttl}[0]{Exploring Transformer Backbones for\\
Heterogeneous Treatment Effect Estimation}

\newcommand{\ie}[0]{i.e., }
\newcommand{\eg}[0]{e.g., }

\newcommand{\Oh}{\mathcal{O}}
\newcommand{\abbr}[0]{TransTEE} %\xspace
\newcommand{\indep}{\rotatebox[origin=c]{90}{$\models$}}

\usepackage[capitalize,noabbrev]{cleveref}

\theoremstyle{plain}
\usepackage{mdframed}
\newmdtheoremenv[linewidth=0pt,innerleftmargin=4pt,innerrightmargin=4pt]{prop}{Proposition}
\newtheorem{theorem}{Theorem}[section]

\theoremstyle{definition}

\newtheorem{assump}[theorem]{Assumption}
\theoremstyle{remark}

\newcommand\blfootnote[1]{%
  \begingroup
  \renewcommand\thefootnote{}\footnote{#1}%
  \addtocounter{footnote}{-1}%
  \endgroup
}
\usepackage[textsize=tiny]{todonotes}
\title{\ttl}

\author{%
  Yi-Fan Zhang$^{1,*}$ 
  Hanlin Zhang$^{2,*}$ 
  Zachary C. Lipton$^{2}$ 
  Li Erran Li$^{3}$ 
  Eric P. Xing$^{2,4}$ 
}

\begin{document}
% \doparttoc % Tell to minitoc to generate a toc for the parts
% \faketableofcontents
\addtocontents{toc}{\protect\setcounter{tocdepth}{0}} 
\maketitle
\blfootnote{$^*$Equal contribution $^{1}$Chinese Academy of Science $^{2}$Carnegie Mellon University $^{3}$AWS AI, Amazon $^{4}$MBZUAI. Correspondence to: Hanlin Zhang <hanlinzh@cs.cmu.edu $>$.}

\begin{abstract}
\input{arxiv/sections/00_abstract}
\end{abstract}

\section{Introduction}\label{sec:intro}
\input{arxiv/sections/01_intro}
\section{Related Work}
\label{sec:related}
\input{arxiv/sections/02_related}

\section{Problem Statement and Assumptions} %sub
\input{arxiv/sections/03_problem}

\section{\abbr: Transformers as Treatment Effect Estimators} %sub

\input{arxiv/sections/04_transformers}
% \section{Propensity Score Modeling}
% \input{arxiv/sections/05_propensity}
\vspace{2mm}
\section{Experimental Results}\label{sec:exp}
\input{arxiv/sections/06_experiments}
\section{Concluding Remarks}
\input{arxiv/sections/07_conclusions}

\section*{Acknowledgements}
We thank Edward H. Kennedy and Hongyi Wang for helpful discussions and feedback.


\bibliography{ref.bib} 
\bibliographystyle{plain}

%%%%%%%%%%%%%%%%%%%%%%%%%%%%%%%%%%%%%%%%%%%%%%%%%%%%%%%%%%%%
\newpage
\appendix
%\part{\Large\centering{-- Appendix --}} %\ttl \\
\begin{center}
{\LARGE \textbf{Appendix}}
\end{center}

{
  \hypersetup{hidelinks}
  \tableofcontents
  %\parttoc
  \noindent\hrulefill
}

\addtocontents{toc}{\protect\setcounter{tocdepth}{2}} 
\clearpage
\newpage

% \renewcommand \thepart{} 
% \renewcommand \partname{}
% \parttoc % Insert the appendix TOC

\input{arxiv/sections/appendix}

\end{document}

%% file: arxiv/sections/00_abstract.tex
Previous works on Treatment Effect Estimation (TEE) are not in widespread use because they are predominantly theoretical, where strong parametric assumptions are made but untractable for practical application. Recent work uses multilayer perceptron (MLP) for modeling casual relationships, however, MLPs lag far behind recent advances in ML methodology, which limits their applicability and generalizability. To extend beyond the single domain formulation and towards more realistic learning scenarios, we explore model design spaces beyond MLPs, i.e., transformer backbones, which provide flexibility where attention layers govern interactions among treatments and covariates to exploit structural similarities of potential outcomes for confounding control. Through careful model design, \underline{\textbf{Trans}}formers as \underline{\textbf{T}}reatment \underline{\textbf{E}}ffect \underline{\textbf{E}}stimators (TransTEE) is proposed. We show empirically that TransTEE~ can: (1) serve as a general purpose treatment effect estimator that significantly outperforms competitive baselines in a variety of challenging TEE problems (e.g., discrete, continuous, structured, or dosage-associated treatments) and is applicable to both when covariates are tabular and when they consist of structural data (e.g., texts, graphs); (2) yield multiple advantages: compatibility with propensity score modeling, parameter efficiency, robustness to continuous treatment value distribution shifts, explainable in covariate adjustment, and real-world utility in auditing pre-trained language models. \footnote{Code repository: \url{https://github.com/hlzhang109/TransTEE}}

%% file: arxiv/sections/01_intro.tex
\vspace{-1.6mm}
One of the fundamental tasks in causal inference 
is to estimate treatment effects given covariates, treatments and outcomes.
Treatment effect estimation is a central problem of interest
in clinical healthcare
and social science \citep{imbens2015causal}, 
as well as econometrics \citep{wooldridge2015introductory}. 
Under certain conditions \citep{rosenbaum1983central},
the task can be framed as a particular type
of missing data problem,
whose structure is fundamentally different 
in key ways from supervised learning 
and entails a more complex set of covariate and treatment representation choices. 

Previous works in statistics leverage parametric models \citep{imbens2015causal, wager2018estimation, kunzel2019metalearners, foster2019orthogonal} to estimate heterogeneous treatment effects.
To improve their utility, feed-forward neural networks 
have been adapted to model causal relationships 
and estimate treatment effects
\citep{yoon2018ganite, bica2020scigan, schwab2020drnet, nie2021vcnet, curth2021on},
in part due to their flexibility in modeling nonlinear functions \citep{hornik1989multilayer} 
and high-dimensional input \citep{johansson2016learning}. 
Among them, the specialized NN's architecture 
plays a key role in learning representations 
for counterfactual inference \citep{alaa2018limits, curth2021on} 
such that treatment variables and covariates are well distinguished \citep{shalit2017tarnet}. 

Despite these encouraging results, 
several key challenges make it difficult
to adopt these methods as
% general-purpose 
% treatment effect estimators.
standard tools for 
treatment effect estimation.
% Fundamentally, current works based on subnetworks are not well equipped with suitable inductive biases 
Most current works based on subnetworks do not sufficiently exploit 
the structural similarities 
of potential outcomes for heterogeneous TEE\footnote{For example, $\mathbb{E}[Y(1)-Y(0)|X]$ is often of a much simpler form to estimate than either $\mathbb{E}[Y(1)|X]$ or $\mathbb{E}[Y(0)|X]$, due to inherent similarities between $Y(1)$ and $Y(0)$.} and accounting for them needs complicated regularizations, reparametrization or multi-task architectures that are problem-specific \citep{curth2021on}. 
Moreover, they heavily rely on their treatment-specific designs and cannot be easily extended beyond the narrow context in which they are originally. For example, they have poor practicality and generalizability when high-dimensional structural data (e.g., texts and graphs) are given as input \citep{kaddour2021causal}.
Besides, those MLP-based models currently lag far behind
recent advances in machine learning methodology, which are prone to issues of scale, expressivity and flexibility. Specifically, those side limitations include parameter inefficiency (Table \ref{tab:diffs}), and brittleness under different scenarios, such as when treatments shift slightly from the training distribution.
The above limitations clearly show a 
%erran8: clear need
pressing need
for an effective and practical framework to estimate treatment effects.

\begin{wrapfigure}{r}{0.45\textwidth}
    \begin{center}
    \includegraphics[width=0.45\textwidth]{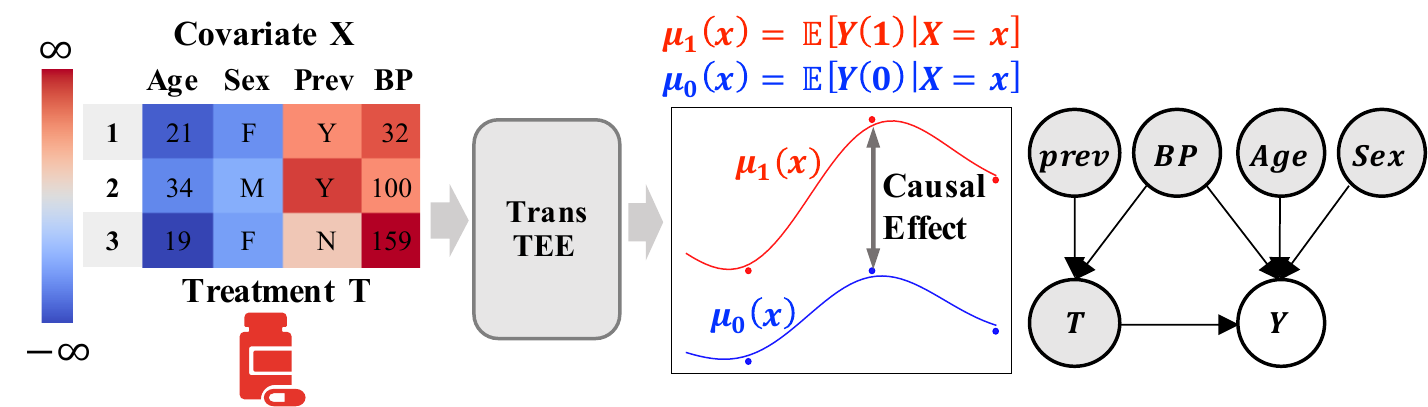}
    \end{center}
    \caption{\textbf{A motivating example} with a corresponding causal graph.
\textbf{Prev} denotes previous infection condition 
and \textbf{BP} denotes blood pressure. 
\abbr~adjusts an appropriate covariate set 
$\{\textbf{Prev}, \textbf{BP}\}$ 
with attention which is visualized via a heatmap.}
    \label{fig:teaser1}
\end{wrapfigure}

In this work,
we explore recent advanced models in the deep learning community to enhance model design for TEE tasks. Specifically, the core idea of our approach consists of three parts: as an S-learner, \abbr~embeds all treatments and covariates, which avoids multi-task architecture and shows improved flexibility and robustness to continuous treatment value distribution shifts; attention mechanisms are used for modeling treatment interaction and treatment-covariate interaction. In this way, \abbr~enables adaptive covariate selection 
\citep{de2011covariate, vanderweele2019principles} to infer causal effects.
For example, one can observe in Figure \ref{fig:teaser1} 
that both pre-treatment covariates and confounders 
are appropriately adjusted with higher weights, 
which recovers the ``disjunctive cause criterion'' \citep{de2011covariate} 
that accounts for those two kinds of covariates 
and is helpful for ensuring the plausibility 
of the conditional ignorability assumption 
when complete knowledge of a causal graph is not available. 
This recipe also gives improved versatility 
when working with heterogeneous treatments types
(Figure \ref{fig:framework}).

Our first contribution 
shows
that transformer backbones, equipped with proper design choices, can be effective and versatile treatment effect estimators under the Rubin-Neyman potential outcomes framework. \abbr~is empirically verified to be 
(i) a flexible framework applicable for a wide range of TEE settings; 
(ii) compatible and effective with propensity score modeling; 
(iii) parameter-efficient; 
(iv) explainable in covariate adjustment; 
(v) robust under continuous treatment shifts; 
(vi) useful for debugging pre-trained language models (LMs) to promote favorable social outcomes.

Moreover, comprehensive experiments on six benchmarks with four types of treatments are conducted to verify the effectiveness of \abbr~in estimating treatment effects. We show that \abbr~produces covariate adjustment interpretation and significant performance gains given discrete, continuous or structured treatments on popular benchmarks including IHDP, News, TCGA. We introduce a new surrogate modeling task to expand the scope of TEE beyond semi-synthetic evaluation and show that \abbr~ is effective in real-world applications such as auditing fair predictions of LMs.

\begin{table*}[t]
\caption{\textbf{Comparison of existing works
and \abbr~in terms of parameter complexity.}
$n$ is the number of treatments. 
$B_T,B_D$ are the number of branches 
for approximating continuous treatment and dosage.
Treatment interaction means explicitly modeling 
collective effects of multiple treatments.
\abbr~is general for all the factors.}\label{tab:diffs} 
\vspace{-0.2cm}
\begin{center}
\begin{small}
\begin{sc}
\resizebox{\textwidth}{!}{
\begin{tabular}{@{}ccccc@{}}
\toprule
Methods      & Discrete Treatment       & Continuous Treatment & Treatment Interaction & Dosage    \\ \midrule
TARNet \citep{shalit2017tarnet}      & $\Oh(n)$ &     &                   &              \\ % & ICML 2017 
Perfect Match \citep{schwab2018perfect}    & $\Oh(n)$ &     &      $\Oh(2^T)$             &              \\ % & Arxiv 2018 
Dragonnet \citep{shi2019dragonnet}  & $\Oh(n)$ & &                     &             \\ % & NeurIPS 2019
DRNet \citep{schwab2020drnet}         & $\Oh(n)$ &  &                     & $\Oh(TB_D)$ \\ % & AAAI 2020
SCIGAN \citep{bica2020scigan}      & $\Oh(n)$ &  &                     & $\Oh(TB_D)$ \\ % & NeurIPS 2020
VCNet \citep{nie2021vcnet}       & $\Oh(1)$ &  $\Oh(1)$         &                     &              \\ %& ICLR 2021 
NCoRE \citep{parbhoo2021ncore}    & $\Oh(n)$ &  $\Oh(B_T)$ & $\Oh(n)$        &              \\ %  & Arxiv 2021
FlexTENet \citep{curth2021on}      &$\Oh(n)$ &   &        &              \\ %   & NeurIPS 2021
Ours  & $\Oh(1)$ & $\Oh(1)$ & $\Oh(1)$  & $\Oh(1)$ \\ \bottomrule 
\end{tabular}}
\end{sc}
\end{small}
\end{center}

\end{table*}

%% file: arxiv/sections/02_related.tex
\vspace{-2.6mm}

\textbf{Neural Treatment Effect Estimation.}
There are many recent works on adapting neural networks to learn counterfactual representations for treatment effect estimation \citep{johansson2016learning, shalit2017tarnet, louizos2017causal, yoon2018ganite, bica2020scigan, schwab2020drnet, nie2021vcnet, curth2021on}. To mitigate the imbalance of covariate representations across treatment groups, various approaches are proposed including optimizing distributional divergence (e.g. IPM including MMD, Wasserstein distance), entropy balancing \citep{zeng2020double} (converges to JSD between groups), counterfactual variance \citep{zhang2020learning}. However, their domain-specific designs make them limited to different treatments as shown in Table \ref{tab:diffs}: methods like VCNet \citep{nie2021vcnet} use a hand-crafted way to map a real-value treatment to an $n$-dimension vector with a constant mapping function, which is hard to converge under shifts of treatments (Table \ref{tab:syn_emb} in Appendix); models like TARNet \citep{shalit2017tarnet} need an accurate estimation of the value interval of treatments. Moreover, previous estimators embed covariates to only one representation space by fully connected layers, tending to lose their connection and interactions \citep{shalit2017tarnet, johansson2020generalization}. And it is non-trivial to adapt to the wider settings given existing ad hoc designs on network architectures. For example, the case with $n$ treatments and $m$ associated dosage requires $n\times m$ branches for methods like DRNet \citep{schwab2020drnet}, which put a rigid requirement on the extrapolation capacity and infeasible given observational data. 

\textbf{Transformers and Attention Mechanisms} Transformers~\citep{vaswani2017attention} have demonstrated exemplary performance on a broad range of language tasks and their variants have been successfully adapted to representation learning over images~\citep{dosovitskiy2020image}, programming languages~\citep{chen2021evaluating}, and graphs~\citep{ying2021do} partly due to their flexibility and expressiveness. Their wide utility has motivated a line of work for 
%erran1: general-purpose perception \citep{jaegle2021perceiver, jaegle2022perceiver} about various modalities like images, point clouds, audios and videos. 
general-purpose neural architectures  \citep{jaegle2021perceiver, jaegle2022perceiver} that can be trained to perform tasks in various modalities such as images, point clouds, audios and videos. 
But causal inference is fundamentally 
%erran: add
different
from the above models' focus, i.e. supervised learning. And one of our goals is to explore the generalizability of attention-based models for TEE across domains with high-dimensional inputs, an important desideratum in causal representation learning \citep{scholkopf2021toward}. 
There are recent attempts to use attention mechanisms for TEE Tasks~\citep{guo2021cetransformer,xu2022learning}.
CETransformer~\citep{guo2021cetransformer} embeds covariates for different treatments as a T-learner. But it only trivially learns covariate embeddings without representing treatments. %, while the latter is shown to be more important for TEE tasks. 
In contrast, \abbr~is an S-learner, which is more suited to account for causal heterogeneity \citep{kunzel2019metalearners, curth2021on, curth2021nonparametric}.
ANU~\citep{xu2022learning} uses attention mechanisms to map the original covariate space $X$ into a latent space $Z$ with a single model. We detail the difference in Appendix~\ref{app:related}.

%% file: arxiv/sections/03_problem.tex
\label{sec:problem}
\textbf{Treatment Effect Estimation.} We consider a setting in which we are given $N$ observed samples $(\x_i,t_i,s_i, y_i)_{i=1}^N$, each containing $N$ pre-treatment covariates $\{\x_i\in\mathbb{R}^p\}_{i=1}^{N}$. The treatment variable $t_i$ in this work has various support, \eg $\{0,1\}$ for binary treatment settings, $\mathbb{R}$ for continuous treatment settings, and graphs/words for structured treatment settings. For each sample, the potential outcome ($\mu$-model) $\mu(\x,t)$ or $\mu(\x,t, s)$ is the response of the $i$-th sample to a treatment $t$, where in some cases each treatment will be associated with a dosage $s_{t_i}\in\mathbb{R}$. The propensity score ($\pi$-model) is the conditional probability of treatment assignment given the observed covariates $\pi(T=t|X=\x)$. The above two models can be parameterized as $\mu_\theta$ and $\pi_\phi$, respectively. The task is to estimate the Average Dose Response Function (ADRF): $\mu(\x, t) = \mathbb{E}[Y|X=\x, do(T=t)]$ \citep{shoichet2006interpreting}, which includes special cases in discrete treatment scenarios that can also be estimated as the average treatment effect (ATE): $ATE=\mathbb{E}[\mu(\x,1)-\mu(\x,0)]$ and its individual version ITE. 

What makes the above problem more challenging than supervised learning is that we never see the missing counterfactuals and ground truth causal effects in observational data. Therefore, we first introduce the required fundamentally important assumptions that give the strongly ignorable condition such that statistical estimands can be interpreted causally.

\begin{assump}(Ignorability/Unconfoundedness) implies no hidden confounders such that $Y(T=t) \indep T|X$. In the binary treatment case, $Y(0),Y(1) \indep T|X$.
\label{assump:unconfound}
\end{assump}

\begin{assump}(Positivity/Overlap) The treatment assignment is non-deterministic such that, i.e. $0 < \pi(t|x) < 1, \forall x \in \mathcal{X}, t \in \mathcal{T}$
\label{assump:positivity}
\end{assump}

Assumption \ref{assump:unconfound} ensures the causal effect is identifiable, 
implying that treatment is assigned independent of the potential outcome and randomly for every subject regardless of its covariates, which allows estimating ADRF using $\mu(t) := \mathbb{E}[Y|do(T=t)] = \mathbb{E}[\mathbb{E}[[Y|\x, T=t]]$ \citep{rubin1978bayesian}. A naive estimator of $\mu(\x,t)=\mathbb{E}[Y|X=\x, T=t]$ is the sample average $\mu(t)=\sum_{i=1}^{n}\hat{\mu}(\x_i,t)$. Assumption \ref{assump:positivity} states that there is a chance of seeing units in every treated group.

%% file: arxiv/sections/04_transformers.tex
\begin{figure*}
    \centering
    \includegraphics[width=.99\textwidth]{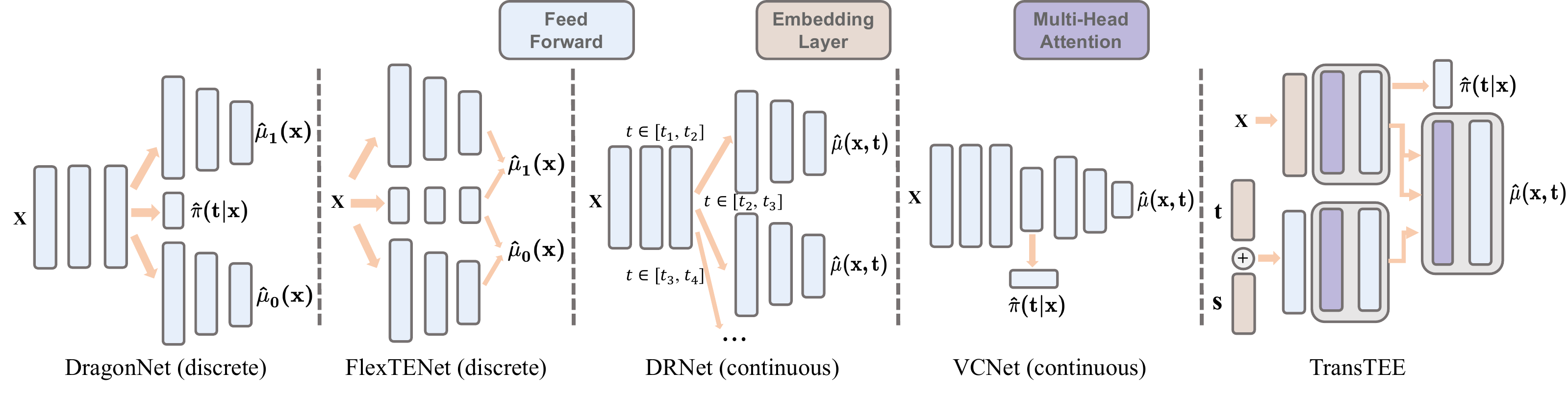}
    \caption{\textbf{A schematic comparison} of \abbr~and recent works including DragonNet\citep{shi2019dragonnet}, FlexTENet\citep{curth2021on}, DRNet\citep{schwab2020drnet} and VCNet\citep{nie2021vcnet}. \abbr~handles all the scenarios without handcrafting treatment-specific architectures 
    and any additional parameter overhead.
    }
    \label{fig:framework}
    \vspace{-0.2cm}
\end{figure*}

The systematic similarity of potential outcomes of different treatment groups is important for TEE \citep{curth2021on}. Note that $\x$ is often high-dimensional while $t$ is not, which means naively feeding $(\x,t)$ to MLPs is not favorable since the impacts of treatment tend to be lost. 
As a result, various architectures and regularizations 
have been proposed to enforce structural similarity 
and differences among treatment groups.
However, they are limited to specific use cases,
as shown in Section \ref{sec:related} and Figure \ref{fig:framework}. To remedy it, we use three simple yet effective design choices based on attention mechanisms. The resulting scalable framework \abbr~can tackle the problems of most existing treatment effect estimators 
(\eg multiple/continuous/structured treatments, 
treatments interaction, and treatments with dosage) 
without ad-hoc architectural designs, \eg multiple branches. 

\textbf{Preliminary.} The main module in \abbr~is the attention layer~\cite{vaswani2017attention}: 
given $d$-dimensional query, key, and value matrices 
$Q\in\mathbb{R}^{d\times d_k},K\in\mathbb{R}^{d\times d_k},V\in\mathbb{R}^{d\times d_v}$, 
attention mechanism computes the outputs as $\mathcal{H}(Q,K,V)=\text{softmax}(\frac{QK^T}{\sqrt{d_k}})V$.
In practice, multi-head attention is preferable to jointly attend to the information from different representation subspaces.
\begin{equation}
\begin{aligned}
\mathcal{H}_M(Q,K,V)=\text{Concat}(head_1, ...,head_h)W^O, \text{where } head_i=\mathcal{H}(QW_i^Q,KW_i^K,VW_i^V),\nonumber
\end{aligned}
\end{equation}
where $W_i^Q\in\mathbb{R}^{d\times d_k},W_i^V\in\mathbb{R}^{d\times d_k},W_i^V\in\mathbb{R}^{d\times d_v}$ and $W^O\in\mathbb{R}^{hd_v\times d}$ are learnable matrices. 

\subsection{Covariate and Treatment Embedding Layers}
\textbf{Treatment Embedding Layer.} As illustrated in Figure \ref{fig:framework} and Table.~\ref{tab:diffs}, as treatments are often of much lower dimension compared to covariates, to avoid missing the impacts of treatments, previous works (\eg DragonNet~\citep{shi2019dragonnet}, FlexTENet \citep{curth2021on}, DRNet \citep{schwab2020drnet}) assign covariates from different treatment groups to different branches, which is \textit{highly parameter inefficient}. Besides, We analyze in Proposition~\ref{prop1} (Appendix \ref{app:fail}) that, for continuous treatments/dosages, the performance is affected by both number of branches and the value interval of treatment. However, almost all previous works on continuous treatment/dosage assume the treatment or dosage is in a fixed value interval \eg $[0,1]$ and Figure~\ref{fig:ADRF} 
shows that prevalent works \textit{fail when tested under shifts of treatments}. These two observations motivate us to use two learnable linear layers to project scalar treatments and dosages to $d$-dimension vectors separately: $M_t=\text{Linear}(t), M_s=\text{Linear}(s)$, where $M_t\in\mathbb{R}^{d}$. $M_s\in\mathbb{R}^{d}$ exists just when each treatment has a dosage parameter, otherwise, only treatment embedding is needed. When multiple ($n$) treatments act simultaneously, the projected matrix will be $M_t\in\mathbb{R}^{d\times n},M_s\in\mathbb{R}^{d\times n}$ and when facing structural treatments (languages, graphs), the treatment embedding will be projected by language models and graph neural networks respectively. By using the treatment embeddings, \abbr~is shown to be (i) \textit{robust under treatment shifts}, and  (ii) \textit{{parameter-efficient}}.

\textbf{Covariates Embedding Layer.} Different from previous works that embed all covariates by one fully connected layer, where the differences between covariates tend to be lost, it is difficult to study the function of an individual covariate in a sample. \abbr~learns different embeddings for each covariate, namely $M_x=\text{Linear}(\x)$, and $M_x\in\mathbb{R}^{d\times p}$, where $p$ is the number of covariate. Covariates embedding enables us to study the effect of individual covariate on the outcome.

\subsection{Covariate and Treatment Self-Attention}
For covariates, prevalent methods represent covariates as a whole feature using MLPs, where {pair-wise covariate interactions are lost} when adjusting covariates. Therefore, we cannot study the effect of each covariate on the estimated result. 
In contrast, \abbr~processes each covariate embedding independently and model their interactions by self-attention layers. Namely,
\begin{equation}
\begin{aligned}
&\hat{M}_x^{l}=\mathcal{H}_M(M_x^{l-1},M_x^{l-1},M_x^{l-1})+M_x^{l-1}, {M}_x^{l}=\text{MLP}(\text{BN}(\hat{M}_x^{l}))+\hat{M}_x^{l}.
\end{aligned}\nonumber
\end{equation}
where $M_x^{l}$ is the output of $l$ layer and $\text{BN}$ is the BatchNorm layer. Simultaneously, the treatments and dosages embeddings are concatenated and projected to the latent dimension by a linear layer, which generates a new embedding $M_{st}\in\mathbb{R}^{d}$. Then self-attention is applied
\begin{equation}
\begin{aligned}
&M_{st}^{l}=\mathcal{H}_M(M_{st}^{l-1},M_{st}^{l-1},M_{st}^{l-1})+M_{st}^{l-1}, {M}_{st}^{l}=\text{MLP}(\text{BN}(\hat{M}_{st}^{l}))+\hat{M}_{st}^{l}.
\end{aligned}\nonumber
\end{equation}
The self-attention layer for treatments enables treatment interactions, an important desideratum for S- and T-learners. Namely, \abbr~can \textit{model the scenario where multiple treatments are applied and attain strong practical utility}, \eg multiple prescriptions in healthcare or different financial measures in economics. This is an effective remedy for existing methods which are limited to settings where various treatments are not used simultaneously.

\subsection{Treatment-Covariate Cross-Attention}
One of the fundamental challenges of {causal meta-learners is to model treatment-covariate interactions}. \abbr~realizes this by a cross-attention module, treating ${M}_{st}$ as query and ${M}_x$ as key and value
\begin{equation}
\begin{aligned}
&\hat{M}^{l}=\mathcal{H}_M(M_{st}^{l-1},M_x^{l-1},M_x^{l-1})+M^{l-1},\\
&{M}^{l}=\text{MLP}(\hat{M}^{l})+\hat{M}^{l},\hat{y}=\text{MLP}(\text{Pooling}({M}^{L})),
\end{aligned}\nonumber
\end{equation}
where ${M}^{L}$ is the output of the last cross-attention layer and $M^0=M_{st}^L$. 
The above interactions are particularly important for adjusting proper covariate or confounder sets for estimating treatment effects \citep{vanderweele2019principles}, which empirically yields \textit{suitable covariate adjustment principles (the Disjunctive Cause Criteria) \citep{de2011covariate, vanderweele2019principles} about pre-treatment covariates and confounders} as intuitively illustrated in Figure \ref{fig:teaser1} and corroborated in our experiments.

Denote $\hat{y} \coloneqq \mu_\theta(\x,t)$ and the training objective is the mean square error (MSE) of the outcome regression: 
\begin{equation}
\mathcal{L}_{\theta}(\x, y, t) = \sum_{i=1}^n\left(y_i-\mu_\theta(\x_i,t_i)\right)^2.
    \label{eq:outcome}
\end{equation}

\textbf{Remark.} We include an illustration of \abbr~by a concrete example in Appendix~\ref{sec:example}. Note that, although the embedding technique and attention mechanisms are commonly used in Computer Vision, Neural Language Processing communities, it is not well understood \textit{how to guide the design of these modules for causal inference} and \textit{why these techniques benefit TEE tasks are underexplored}. In this work, through the flexible use of embedding and attention mechanisms, we design a strong TEE architecture, we further use conceptual analysis and empirical results to show the benefit brought by the used design choices. 
Besides, when combined with the strong modeling capacity of Transformers, \textit{\abbr~can be extended to high-dimensional data flexibly and effectively} on structured data. The generalizability of the \abbr~also allows new applications like auditing language models beyond semi-synthetic settings as shown in the next section. 

\vspace{-4.6mm}

%% file: arxiv/sections/06_experiments.tex
\vspace{-1.6mm}

We elaborate on basic experimental settings, results, analysis, and empirical studies in this section.
See Appendix \ref{app:settings} for full details
of all experimental settings and detailed definitions of metrics.
See Appendix \ref{app:exp} for many more results and remarks.

\begin{table*}[t]

\caption{\textbf{Experimental results comparing NN-based methods on the IHDP datasets,} where \textbf{——} means the model is not suitable for continuous treatments. We report results based on 100 repeats, and the numbers after $\pm$ are the estimated standard deviation of the average value. For the vanilla setting with binary treatment, we report the mean absolute difference between the estimated and true ATE. For Extrapolation ($h=2$), models are trained with $t\in[0.1,2.0]$ and tested in $t\in[0,2.0]$. For Extrapolation ($h=5$), models are trained with $t\in[0.25,5.0]$ and tested in $t\in[0,5]$. } 
\label{tab:continuous1}
\vspace{-0.2cm}
\begin{center}
\begin{scriptsize}
\begin{sc}
\resizebox{\textwidth}{!}{
\begin{tabular}{cccccc}
\toprule
Methods & Vanilla (Binary) & Vanilla ($h=1$) & Extrapolation ($h=2$) &Vanilla ($h=5$) & Extrapolation ($h=5$) \\
\midrule
    TARNet& 0.3670 $\pm$ 0.61112& 2.0152 $\pm$ 1.07449& 12.967 $\pm$ 1.78108 &5.6752 $\pm$ 0.53161 & 31.523 $\pm$ 1.5013 \\
 DRNet& 0.3543 $\pm$ 0.60622& 2.1549 $\pm$ 1.04483& 11.071 $\pm$ 0.99384& 3.2779 $\pm$ 0.42797 & 31.524 $\pm$ 1.50264 \\
 FlexTENet& 0.2700 $\pm$ 0.10000& \textbf{——}& \textbf{——} & \textbf{——} & \textbf{——} \\
 VCNet& 0.2098 $\pm$ 0.18236& 0.7800 $\pm$ 0.61483& nan & nan &nan\\
  \rowcolor{Gray}
   \abbr& \textbf{0.0983 $\pm$ 0.15384}& 0.1151 $\pm$ 0.10289&0.2745 $\pm$ 0.14976 & 0.1621 $\pm$ 0.14443 & 0.2066 $\pm$ 0.23258\\
   \rowcolor{Gray}
   \abbr+MLE& 0.1721 ± 0.40061& 0.0877 $\pm$ 0.03352&0.2685 $\pm$ 0.17552 & 0.2079 $\pm$ 0.17637 &0.1476 $\pm$ 0.07123\\
   \rowcolor{Gray}
   \abbr+TR& 0.1913 $\pm$ 0.29953& 0.0781 $\pm$ 0.03243&0.2393 $\pm$ 0.08154 & \textbf{0.1143 $\pm$ 0.03224} & \textbf{0.0947 $\pm$ 0.0824}\\
   \rowcolor{Gray}
   \abbr+PTR& 0.2193 $\pm$ 0.34667& \textbf{0.0762 $\pm$ 0.07915}&\textbf{0.2352 $\pm$ 0.17095} & 0.1363 $\pm$ 0.08036 & 0.1363 $\pm$ 0.08035\\

\bottomrule
\end{tabular}
}
\end{sc}
\end{scriptsize}
\end{center}
\vspace{-0.4cm}
\end{table*}

\vspace{-1.6mm}

\subsection{Experimental Settings}

\textbf{Datasets.} 
Since the true counterfactual outcome (or ADRF)
are rarely available for real-world data, 
we use synthetic 
or semi-synthetic data for empirical evaluation. 
 for continuous treatments, we use 
one synthetic dataset
and two semi-synthetic datasets: the \textit{IHDP} and \textit{News} datasets. For treatment with continuous dosages, 
we obtain covariates from a real dataset TCGA~\citep{chang2013cancer} 
and generate treatments, where each treatment is accompanied by a dosage.
The resulting dataset is named \textit{TCGA (D)}.
Following~\citep{kaddour2021causal},
datasets with structured treatments 
include \textit{Small-World (SW)},
which contains $1,000$ uniformly sampled covariates
and $200$ randomly generated Watts–Strogatz
small-world graphs~\citep{watts1998collective} as treatments, 
and \textit{TCGA (S)}, which uses $9,659$ 
gene expression of cancer patients~\citep{chang2013cancer} for covariates 
and $10,000$ molecules from the QM9 dataset~\citep{ramakrishnan2014quantum} as treatments. For the study on language models, 
we use the \textit{Enriched Equity Evaluation Corpus (EEEC)}~\citep{feder2021causalm}. 

\textbf{Baselines.}
Baselines for \textbf{continuous and binary} treatments 
include TARnet~\citep{shalit2017tarnet}, 
Dragonnet~\citep{shi2019dragonnet}, 
DRNet~\citep{schwab2020drnet}, FlexTENet~\citep{curth2021on},
and VCNet~\citep{nie2021vcnet}. 
SCIGAN~\citep{bica2020scigan} is chosen 
as the baseline for \textbf{continuous dosages}.
Besides, we revise DRNet~\citep{schwab2020drnet},
TARNet~\citep{shalit2017tarnet}, 
and VCNet~\citep{nie2021vcnet} 
to DRNet (D), TARNet (D), VCNet (D), respectively, 
which allow multiple treatments and dosages.
Specifically, DRNet (D) has $T$ main flows,
each corresponding to a treatment 
and is divided into $B_D$ branches for continuous dosage. 
Baselines for \textbf{structured} treatments 
include Zero~\citep{kaddour2021causal}, 
GNN~\citep{kaddour2021causal},
GraphITE~\citep{harada2021graphite}, 
and SIN~\citep{kaddour2021causal}. To compare the performance of different frameworks fairly,
all of the models regress on the outcome 
with empirical samples without any regularization. 
For MLE training of the propensity score model, 
the objective is the negative log-likelihood: 
$\mathcal{L}_\phi \coloneqq -\frac{1}{n}\sum_{i=1}^{n}\log \pi_\phi(t_i|\x_i)$.

\textbf{Evaluation Metric.}
For \textbf{continuous and binary} treatments, 
we use the average mean squared error on the test set.
For \textbf{structured} treatments, 
following~\citep{kaddour2021causal}, 
we rank all treatments by their propensity $\pi(t|\mathbf{x})$ in a descending order. Top $K$ treatments are selected 
and the treatment effect of each treatment pair 
is evaluated by unweighted/weighted expected 
Precision in Estimation of Heterogeneous Effect (PEHE)~\citep{kaddour2021causal},
where the WPEHE@K accounts for the fact 
that treatment pairs that are less likely to have higher estimation errors should be given less importance. 
For \textbf{multiple treatments and dosages}, 
AMSE is calculated over all dosage and treatment pairs, 
resulting in AMSE$_{\mathcal{D}}$.

\begin{figure}[t]
    \centering
    \begin{minipage}[t]{0.96\textwidth}        
    \subfigure[$h=1$ in training and testing.]
	{\includegraphics[width=0.32\textwidth]{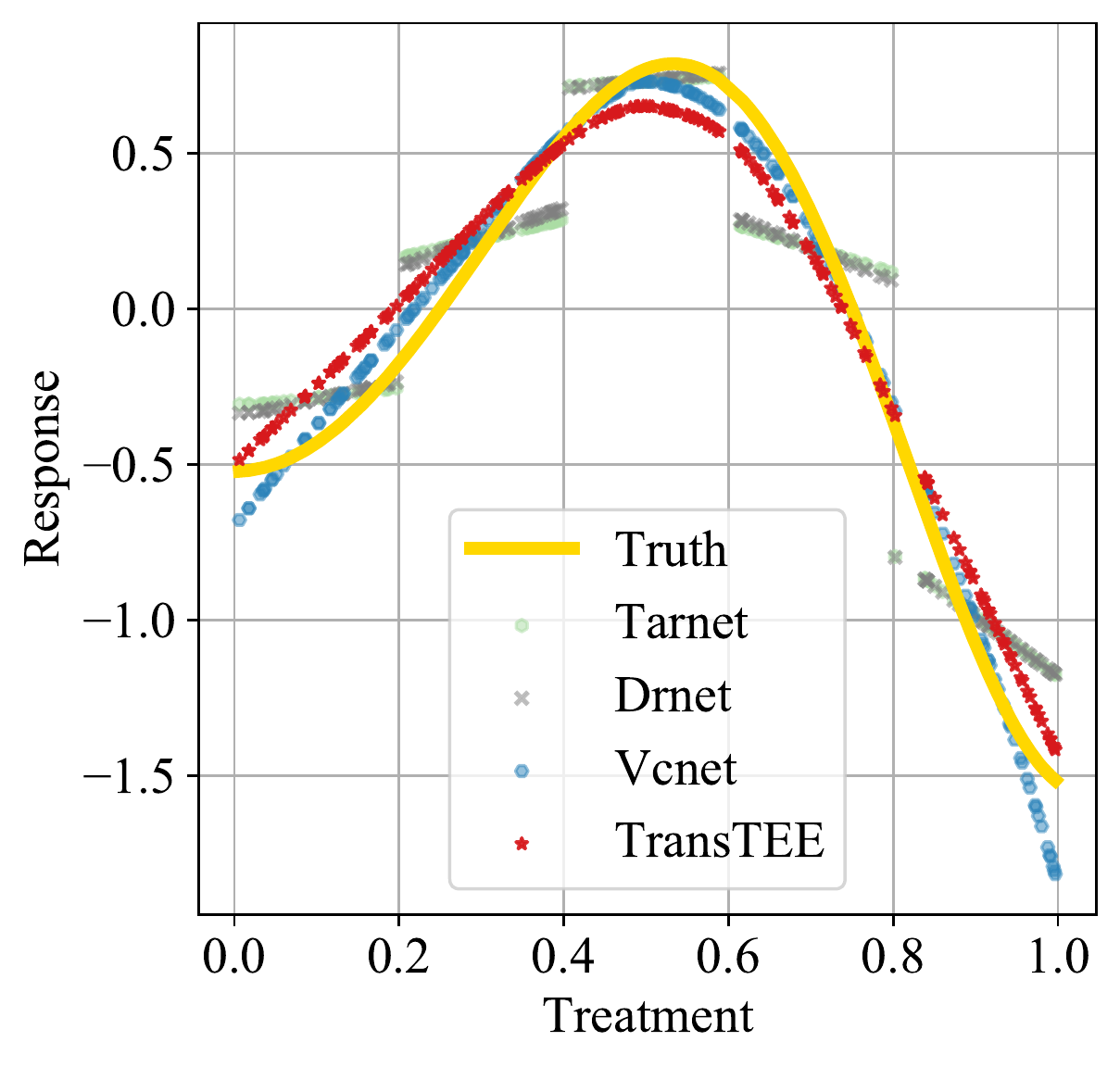}}
	\subfigure[$h=1.75$ in training and $h=2$ in testing (extrapolation).]
	{\includegraphics[width=0.32\textwidth]{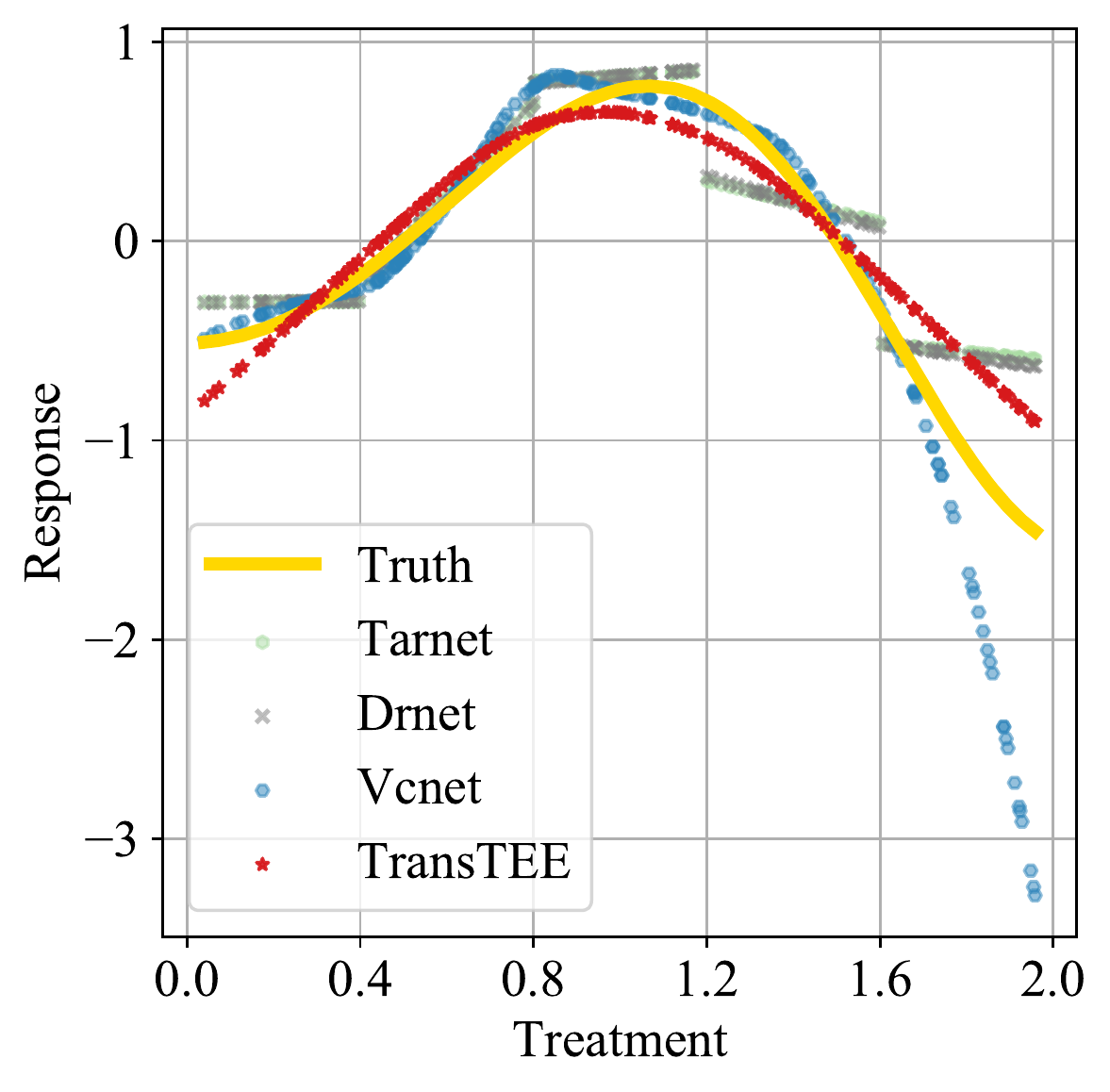}}
	\subfigure[$h=5$ in training and testing.]
	{\includegraphics[width=0.32\textwidth]{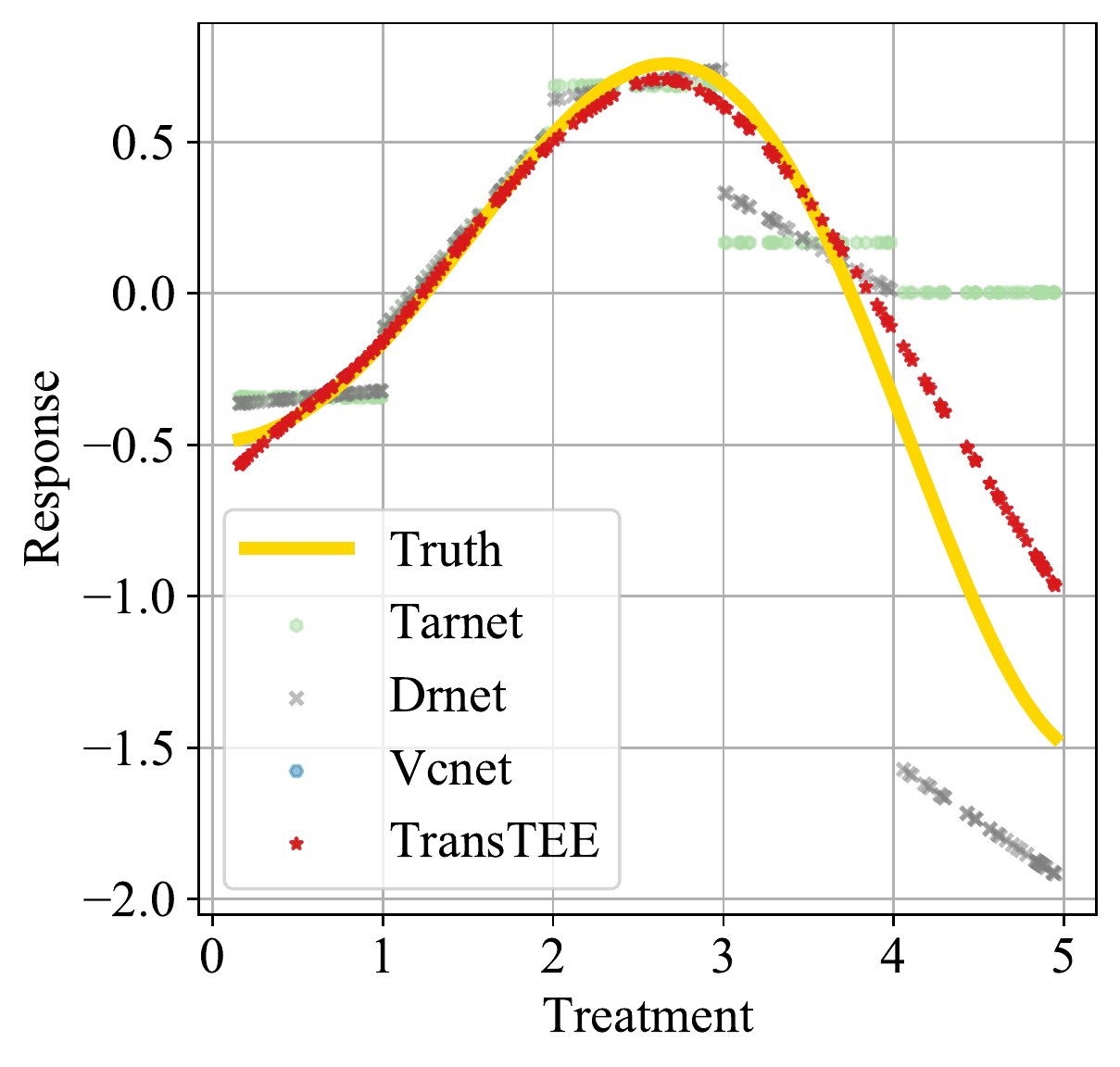}}
    \end{minipage}
    \caption{\textbf{Estimated ADRF on the synthetic dataset}, where treatments are sampled from an interval $[l,h]$, where $l=0$.}
    \label{fig:ADRF}
    \vspace{-0.2cm}
\end{figure}

\subsection{Case Study and Numerical Results}
\textbf{Case study on treatment distribution shifts}
\label{sec:case_study} 
We start by conducting a case study on treatment distribution shifts
(\figurename~\ref{fig:ADRF}),
and exploring an extrapolation setting
in which the treatment may subsequently be administered at values never seen before during training. 
Surprisingly, we find that 
while standard results rely on
constraining the values
of treatments~\cite{nie2021vcnet}
and dosages~\cite{schwab2020drnet}
to a specific range,
our methods perform 
surprisingly well when 
extrapolating beyond these ranges
as assessed on several benchmarks.
In comparison, other methods appear to be comparatively brittle in these same settings.
See Appendix \ref{app:fail} for a detailed discussion.

\textbf{Case study of propensity modeling.} 
\abbr~is conceptually simple and effective. However, when the sample size is small, it becomes important to account for selection bias \citep{alaa2018limits}. However, most existing regularizations can only be used when treatments are discrete~\citep{bica2020estimating,kallus2020deepmatch,du2021adversarial}. Thus we propose two regularization variants for continuous treatment/dosages, which are termed Treatment Regularization (TR, $\mathcal{L}_{\phi}^{TR}(\x, t) = \sum_{i=1}^n\left(t_i - \pi_\phi(\hat{t}_i|\x_i)\right)^2$) and its probabilistic version Probabilistic Treatment Regularization (PTR, $\mathcal{L}_{\phi}^{PTR} = \sum_{i=1}^n \left[\frac{\left(t_i - \pi_\phi(\mu|\x_i)\right)^2}{2\pi_\phi(\sigma^2|\x_i)} + \frac{1}{2} \log \pi_\phi(\sigma^2|\x_i) \right]$) respectively. The overall model is trained in a adversarial pattern, namely $\min_\theta \max_\phi \mathcal{L}_{\theta}(\x, y, t) - \mathcal{L}_{\phi}(\x, t)$. Specifically, a propensity score model $\pi_\phi(t|\x)$ parameterized by an MLP is learned by minimizing $\mathcal{L}_{\phi}(\x, t)$, and then the outcome estimators $\mu_{\theta}\left(\x, t\right)$ are trained by $\min_\theta \mathcal{L}_{\theta}(\x, y, t) - \mathcal{L}_{\phi}(\x, t)$. 
To overcome selection biases, bilevel optimization enforces effective treatment effect estimation while modeling the discriminative propensity features to partial out parts of covariates that cause the treatment but not the outcome and dispose of nuisance variations of covariates \citep{kaddour2021causal}. Such a recipe can account for \textit{selection bias} where $\pi(t|\x)\neq p(t)$ and leave spurious correlations out, which can also be more robust under model misspecification especially in the settings that require extrapolation on treatment (see Table \ref{tab:continuous1} and Appendix \ref{app:discussion} for concrete formalisms and discussions.).

As in Table \ref{tab:continuous1}, Appendix Table \ref{tab:dosage} and Table \ref{tab:continuous2},
with the addition of adversarial training as well as TR and PTR, 
\abbr's estimation error with continuous treatments can be further reduced. 
Overall, TR is better in the continuous case
with smaller treatment distribution shifts,
while PTR is preferable when shifts are greater. 
Both TR and PTR cannot achieve performance gains in discrete cases. 
The superiority of TR and PTR in combination
with \abbr~over comprehensive existing works, 
especially in semi-synthetic benchmarks like IHDP 
that may systematically favor some types of algorithms over others \citep{curth2021really},
also calls for more understanding 
of NNs' inductive biases in treatment effect
estimation problems of interest.
Furthermore, the visualization of covariate selection 
in TR and PTR (Figure~\ref{fig:weights} , Table~\ref{tab:weights} and Appendix \ref{app:exp}) 
supports the idea that modeling the propensity score 
effectively promotes covariate adjustment 
and partials out the effects of the covariates 
on the treatment features. We also compare the training dynamic of different regularizations in Appendix \ref{app:exp}, where TR and PTR are further shown to be able to improve the convergence of \abbr.

\textbf{Continuous treatments.}
To evaluate the efficiency with which \abbr~estimates 
the average dose-response curve (ADRF),
we compare it to other recent NN-based methods 
(Tables~\ref{tab:continuous1}). 
By comparing the results in each column, 
we observe performance boosts for \abbr. 
Further, \abbr~attains 
a much smaller error than baselines 
in cases where the treatment interval
is not restricted to $[0,1]$ (\eg $t\in[0,5]$) 
and when the training and test treatment intervals
are different (extrapolation). 
Interestingly, even vanilla \abbr~produces competitive performance compared to 
that of $\pi(t|\x)$ trained additionally using MLE,
demonstrating the ability of \abbr~to effectively model treatments and covariates. 
The estimated ADRF curves on the IHDP and News datasets 
are shown in~\figurename~\ref{fig:ADRF_ihdp} 
and \figurename~\ref{fig:ADRF_news} in Appendix. 
TARNet and DRNet produce discontinuous ADRF estimators 
and VCNet only performs well 
when $t\in[0,1]$.
However, \abbr~attains lower estimation error 
and preserves the continuity of ADRF
on different treatment intervals.

\begin{table*}[]
\caption{\textbf{Effect of Gender (top) and Race (bottom) on POMS classification with the EEEC dataset}, where ATE$_{GT}$ is the ground truth ATE based on 3 repeats with confidence intervals [CI] constructed using standard deviations.}
\label{causalm}
\footnotesize
\resizebox{\textwidth}{!}{
\begin{tabular}{@{}ccccccccc@{}}
\toprule
       & \multicolumn{4}{c}{Correlation/Representation Based Baselines} & \multicolumn{4}{c}{Treatment Effect Estimators} \\ \midrule
TC     & ATE$_{GT}$    & TReATE   & CONEXP   & INLP    & TarNet  & DRNet  & VCNet & \abbr \\ \midrule
Gender & 0.086        & 0.125    & 0.02     & 0.313   & 0.0067 & 0.0088 & 0.0085 & \textbf{0.013}               \\ 
{[}CI{]} &
  {[}0.082,0.09{]} &
  {[}0.110,0.14{]} &
  {[}0.0,0.05{]} &
  {[}0.304,0.321{]} &
  {[}0.0049, 0.0076{]} &
  {[}0.0084,0.009{]} &
  {[}0.0036, 0.0111{]} &
  {[}0.008, 0.0168{]} \\\midrule
Race   & 0.014        & 0.046    & 0.08     & 0.591   & 0.005   & 0.006  & 0.003 & \textbf{0.0174}               \\
{[}CI{]} &
  {[}0.012,0.016{]} &
  {[}0.038,0.054{]} &
  {[}0.02,0.014{]} &
  {[}0.578,0.605{]} &
  {[}0.0021, 0.0069{]} &
  {[}0.0047, 0.0081{]} &
  {[}0.0025, 0.0037{]} &
  {[}0.0113, 0.0238{]}\\\bottomrule
\end{tabular}
}
\end{table*}

\textbf{Continuous dosage.} In Table~\ref{tab:dosage}, we compare \abbr~against baselines 
on the TCGA (D) dataset with default 
treatment selection bias $2.0$ 
and dosage selection bias $2.0$. 
As the number of treatments increases, 
\abbr~and its variants (with regularization term) 
consistently outperform the baselines
by a large margin on both training and test data. 
\abbr's effectiveness is also shown in Appendix \figurename~\ref{fig:dosage_adrf},
where the estimated ADRF curve of each treatment 
considering continuous dosages is plotted.
Compared to baselines, \abbr~attains
better results over all treatments.
Stronger selection bias in the observed data 
makes estimation more difficult 
because it becomes less likely to see certain treatments or particular covariates. 
Taking into account different dosages and treatment selection biases, 
Appendix \figurename~\ref{fig:dosage_bias} shows that as biases increase, \abbr~consistently performs the best. 

\textbf{Structured treatments.} 
We compared the performance of \abbr~to baselines 
on the training and test set of both SW and TCGA datasets
with varying degrees of treatment selection bias. 
The numerical results are shown in Appendix Table~\ref{tab:results_structure}. 
The performance gain between GNN and Zero 
indicates that taking into account graph information 
significantly improves estimation. 
The results suggest that, overall, 
the performance of \abbr~is the best 
due to the strong modeling ability 
and advanced model structure to process high-dimensional treatments.
SIN is the best model among these baselines. 
However, when the bias is equal to $0.1$, 
SIN fails to achieve estimation results 
better than the Zero baseline.
To evaluate the robustness of each model 
to varying levels of selection bias, 
performance curve with $\kappa\in[0,40]$ 
for the SW dataset and $\kappa\in[0,0.5]$ 
for the TCGA dataset 
are shown in Figure~\ref{fig:kappa_structure}
and Figure~\ref{fig:kappa_structure_unweigt}
in Appendix. 
Considering both metrics,
\abbr~outperforms baselines by a large margin 
across the entire range of evaluated treatment selection biases. 
\vspace{-2.0mm}
\subsection{Analysis}
\vspace{-2.0mm}
\begin{figure*}[t]
    \centering
    \begin{minipage}[t]{\textwidth}        
    \subfigure[]
	{\includegraphics[width=0.27\textwidth]{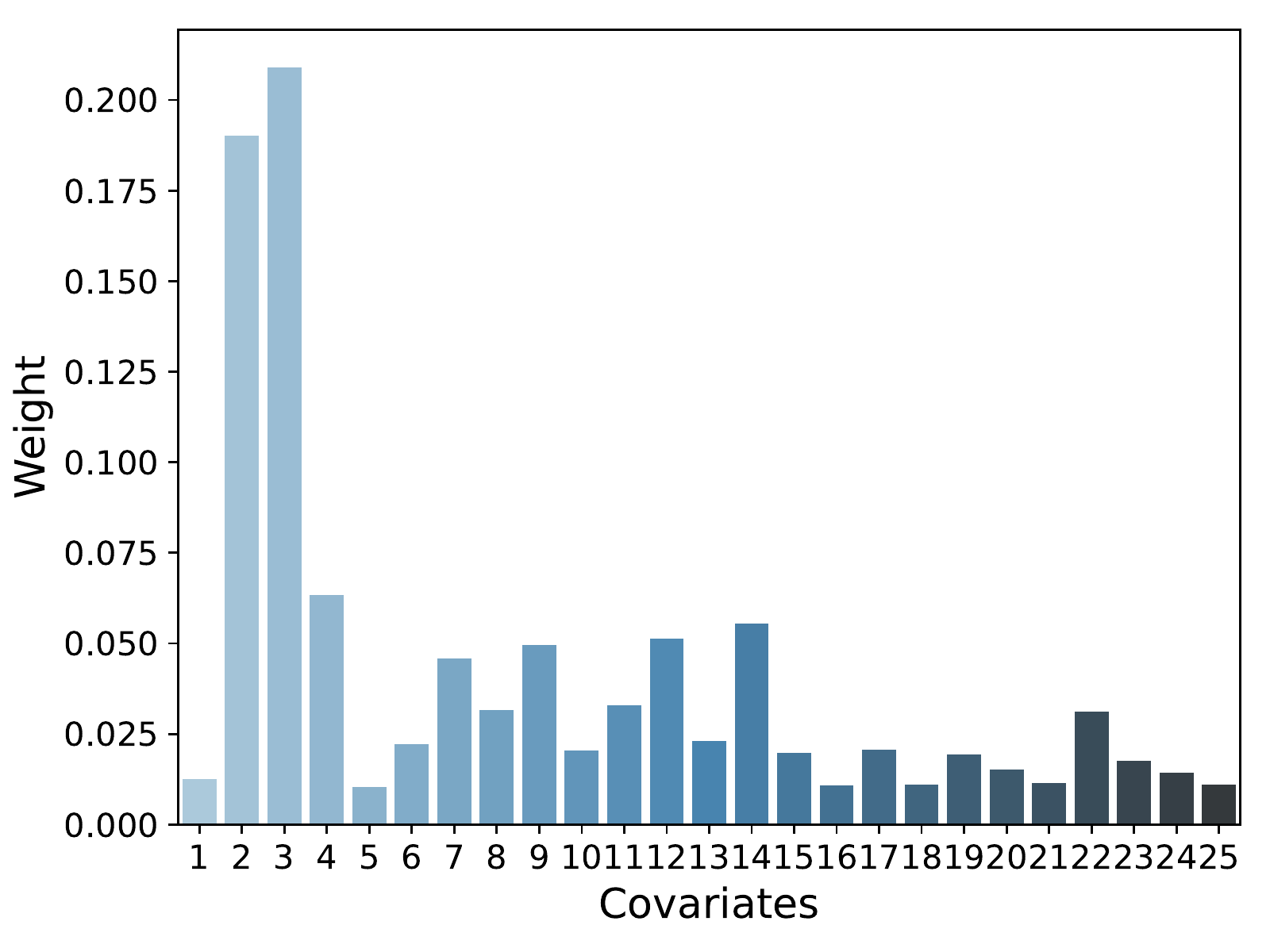}\label{fig:weights}}
	\subfigure[]
	{\includegraphics[width=0.27\textwidth]{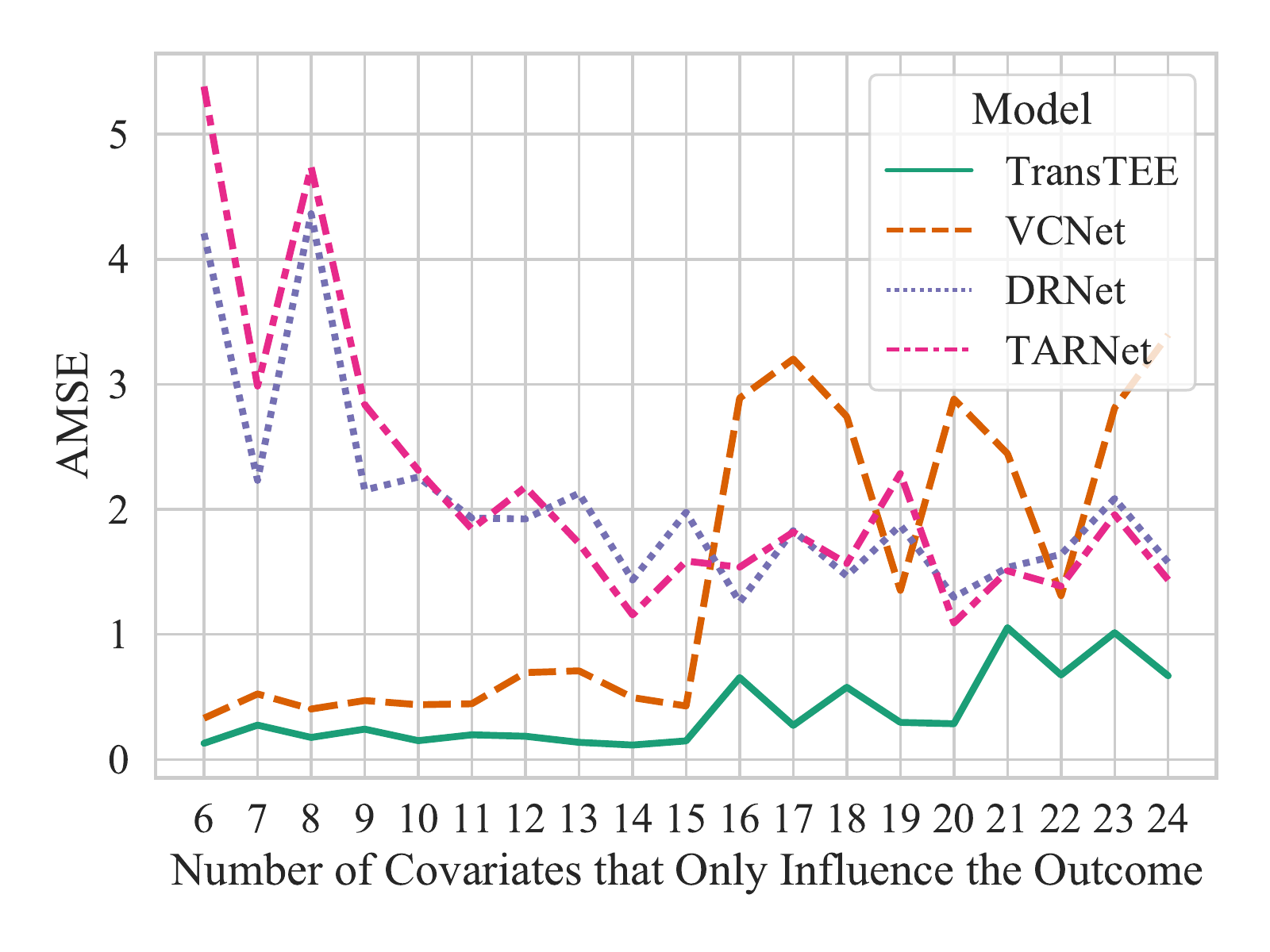}\label{fig:useful_y}}
	\subfigure[]
	{\includegraphics[width=0.45\textwidth]{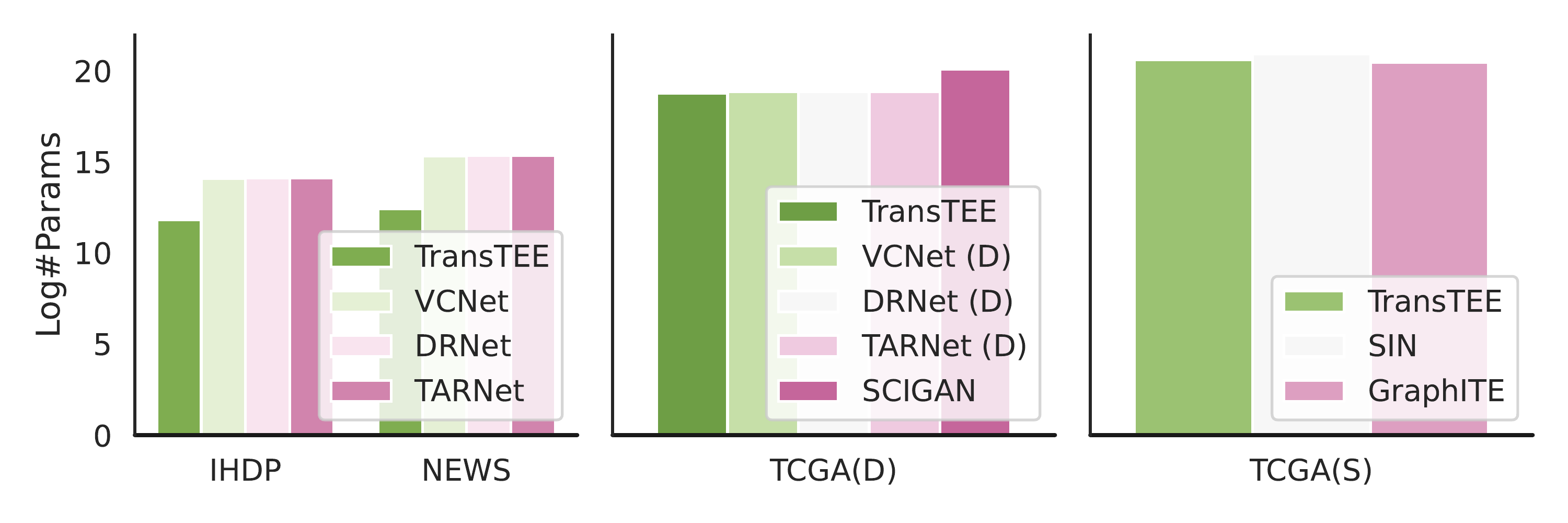}\label{fig:num_params}}
    \end{minipage}
    \vspace{-0.4cm}
    \caption{(a) The learned weights of the cross-attention module on IHDP dataset. \abbr~adjusts confounders $S_{con}=\{1,2,3,5,6\}$ properly with higher weights during the cross attention process. (b) AMSE attained by models on IHDP with different numbers of noisy covariates. (c) Number of parameters for different models on four different datasets, where the log on the y-axis is base 2.}
\end{figure*}

\label{sec:exp_analysis}
\begin{wrapfigure}{l}{0.15\textwidth}
  \begin{center}
    \vspace{-0.2in}  
    \includegraphics[width=0.15\textwidth]{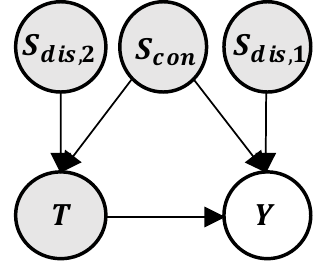}
    \caption{\footnotesize The causal graph of IHDP dataset.}
   \label{fig:adjust_scm}
   \vspace{-0.7cm}
  \end{center}
\end{wrapfigure}
\textbf{Analysis of covariate adjustment of cross-attention module.}  \abbr~embeds each covariate independently and then model the interaction of treatments and covariates for prediction by cross-attention. The resulting interpretability of the covariate adjustment process using attention weights is one clear advantage over existing work. Thus we visualize the covariate selection results (cross-attention weights) in~\figurename~\ref{fig:weights}. 
As elaborated in Appendix~\ref{sec:data_generating}, the IHDP dataset has $25$ covariates, which is divided into $3$ groups: $S_{con}=\{1,2,3,5,6\}$, $S_{dis,1}=\{4, 7\sim 15\}$, and $S_{dis,2}=\{16\sim 25\}$. $S_{con}$ influences both $T$ and $Y$, $S_{dis,1}$ influences only $Y$, and $S_{dis,1}$ influences only $T$. Covariates in $S_{dis,1}$ are named noisy covariates since they have no correlation with the treatment. Their causal relationships are illustrated in Figure~\ref{fig:adjust_scm}. 
Interestingly, confounders $S_{con}$ 
are assigned higher weights 
while noisy covariates 
(those influence the outcome but are irrelevant to the treatment)
lower $S_{dis,1}$,
which matches the principles
in \citep{vanderweele2019principles} 
and corroborate the ability
of \abbr~to estimate 
treatment effects in complex datasets by properly controlling pretreatment variables and confounders. 
Furthermore, \figurename~\ref{fig:useful_y} shows 
that \abbr~consistently outperforms baselines
across different numbers of noisy covariates.

\begin{wraptable}{r}{4.7cm}
\vspace{-0.5cm}
\caption{\textbf{Attention weights} for $S_{con}$, $S_{dis,1}$, and $S_{dis,2}$ respectively.}\label{tab:weights}
\small
\begin{tabular}{@{}cccc@{}}
\toprule
                  & $w_{con}$   & $w_1$   & $w_2$   \\ \midrule
\textbf{TransTEE} & 0.27 & 0.37 & 0.36 \\
\textbf{+TR}       & \textbf{0.59}  & \textbf{0.20} & 0.21 \\
\textbf{+PTR}      & 0.32 & 0.33 & 0.35 \\ \bottomrule
\end{tabular}
\vspace{-0.5cm}
%\end{table}
\end{wraptable}
We further conduct 10 repetitions for \abbr~and its TR and PTR counterparts as reported in Table~\ref{tab:weights} (Appendix \figurename~\ref{fig:attention_weights} visualizes their cross-attention weights). 
Denote $w_{con},w_1,w_2$ as the summation of weights assigned to $S_{con}, S_{dis,1}, S_{dis,2}$ 
respectively. We can see that, incorporated with both TR and PTR regularization, \abbr~assigns more weights to confounding covariates ($S_{con}$) and fewer weights on noisy covariates, which further verifies the compatibility of \abbr~with propensity score modeling since both TR and PTR improve confounding control.
Moreover, TR is better than PTR since it also reduces $w_2$ by a larger margin. This observation gives a suggestion that we should systematically investigate TR and PTR in addition to comparing their numerical performance, especially in settings where 
the assumption of unconfoundedness is violated \citep{ding2017instrumental} and controlling instrumental variables will incur biases in TEE.

\setlength{\columnsep}{8pt}

\textbf{Amount of model parameters comparison.} 
The experiment is to corroborate the conceptual comparison in Table \ref{tab:diffs}.
We find that the proposed \abbr~has consistently fewer parameters 
than baselines on all the settings 
as shown in Figure \ref{fig:num_params}.
Besides, increasing the number of treatments allows a more accurate approximation 
for continuous treatments/dosages, 
most of these baselines need to increase branches, 
which incurs manual efforts and parameter redundancy. 
However, \abbr~is much more efficient. 

\vspace{-2.0mm}
\subsection{Empirical Study on Pre-trained Language Models}\label{sec:lm}
\vspace{-2.0mm}
To evaluate the real-world utility of \abbr, in this subsection, we demonstrate an initial attempt for auditing and debugging large pre-trained language models, an important use case in NLP that is beyond semi-synthetic settings and under-explored in the causal inference literature. Specifically, we use \abbr~to estimate the treatment effects for detecting the effects of domain-specific factors of variation (such as the change of subject's attributes in a sentence) on the predictions of pre-trained language models. We experiment with BERT \citep{kenton2018bert}
(\eg racial and gender-related nouns) over natural language on the (real) EEEC dataset. 
We use both the correlation/representation-based baselines 
introduced in~\citep{feder2021causalm} 
and implement treatment effect estimators
(\eg TARnet, 
DRNet,
VCNet,
and the proposed \abbr).

Interestingly, results in Table~\ref{causalm} 
show that \abbr~effectively estimates 
the treatment effects of domain-specific variation perturbations 
even without substantive downstream fine-tuning on specialized datasets. 
\abbr~outperforms baselines adapted from MLP. 
Moreover, we showcase the top-k samples 
with the maximal/minimal ITE
and analysis in Appendix~\ref{sec:samples_causalm}. 
The results show that \abbr~has the potential 
to serve as surrogate estimators for practical use cases in predicting model predictions \citep{ribeiro2016should}. 
For example, those identified samples 
can provide actionable insights, such as functions as contrast sets 
for analyzing and understanding LMs \citep{gardner2020evaluating, abraham2022cebab} 
and \abbr~can estimate ATE 
to enforce invariant or fairness constraints 
for LMs \citep{veitch2021counterfactual} 
in a lightweight and efficient manner, 
which we leave for future work.

%% file: arxiv/sections/07_conclusions.tex
\label{sec:conclusion}

In this work, we show attention mechanisms 
can be effective and versatile design choices for TEE tasks.
Extensive experiments well verify 
the effectiveness and utility of the proposed \abbr, which also imply that more challenging and unified evaluation alternatives of TEE are needed. Moreover, we hope that our findings can lay the groundwork for future work in developing advanced machine learning techniques like pre-training in large-scale observational data in estimating treatment effects, where \abbr~can serve as an effective backbone. Similar to almost all causal inference methods on observational data, one potential limitation of \abbr~is the reliance on the ignorability assumption. Therefore, an important future direction is extending \abbr~to settings
with more complex causal graphs 
and generate identifiable causal functionals that are 
tractable for optimization \citep{jung2020learning}
supported by identification theory. 
Since adjusting covariates without 
accounting for the causal graph 
might yield inaccurate or biased estimates
of the causal effect \citep{pearl2009causality}, 
how to integrate \abbr~with domain knowledge \citep{imbens2015causal} for alleviating its potential negative societal impacts in consequential decision making will also be important.

%% file: arxiv/sections/appendix.tex
\onecolumn

\section{Extended Related Work}
\label{app:related}

\textbf{Propensity Score.} Most related works fundamentally rely on strongly ignorable conditions. 
Still even under ignorability,
treatments may be selectively 
assigned according to propensities
that depend on the covariates. 
To overcome the impact of such confounding, 
many statistical methods \citep{austin2011introduction} like covariate adjustment \citep{austin2011introduction}, matching \citep{rubin1996matching, abadie2016matching}, stratification \citep{frangakis2002principal}, reweighting \citep{hirano2003efficient}, g-computation \citep{imbens2015causal}, 
have been proposed.
More recent approaches include propensity dropout \citep{alaa2017deep}, and multi-task Gaussian process \citep{alaa2017bayesian}. Explicitly modeling the propensity score, which reflects the underlying policy for assigning treatments to subjects, has also shown to be effective in reasoning about the unobserved counterfactual outcomes and accounting for confounding. Based upon it, double robust estimators and targeted regularization are proposed to guarantee the consistency of estimated treatment effects under misspecification of either the outcome or propensity score model \citep{kang2007demystifying, funk2011doubly}. There are also works using adversarial training for balanced representations~\citep{bica2020estimating,kallus2020deepmatch,du2021adversarial}.
However, most traditional approaches are restricted to binary treatments and the capacity of NNs for such problems has not been fully leveraged. 

\textbf{Domain Adaptation} There are some close connections between causal inference and domain adaptation, in particular, out-of-distribution robustness. Intuitively, traditional domain adversarial training learns representations that are indistinguishable from the domain classifier by minimizing the worst-domain empirical error \citep{ganin2016domain, zhao2018adversarial, wang2021toward, zhang2022principled}. The algorithmic insights can be handily translated to the TEE domain \citep{shalit2017tarnet, johansson2020generalization, feder2021causalm}. Here we also have the desideratum that covariate representations should be balanced such that the selection bias is minimized and the effect is maximally determined by the treatment. Algorithmically, when the treatment is continuous, we connect our method to continuously indexed domain adaptation \citep{wang2020continuously}. Our formulation and algorithm also serve to build connections to a diverse set of statistical thinking on causal inference and domain adaptation, of which much can be gained by mutual exchange of ideas \citep{johansson2020generalization}. Explicitly modeling the propensity score also seeks to connect causal inference with transfer learning to inspire domain adaptation methodology and holds the potential to handle a wider range of problems like hidden stratification in domain generalization, which we leave for future work. 

\textbf{Comparision between \abbr~and ANU~\citep{xu2022learning}.}   (i) The model structure is different. ANU performs cross-attention between $z_x$, and $z_t$, and no self-attention is applied. However, TransTEE performs self-attention on $z_x,z_t$ respectively and then cross-attention is performed between $z_x,z_t$. When facing high-dimensional data, such as texts, images, and graphs, without multiple self-attention layers on $z_x,z_t$ separately, the representations will be weak. That is why in machine translation, object detection, and segmentation tasks, the representations of images/texts will be firstly processed by multiple self-attention layers and then perform cross-attention with queries. We will verify this point in the following experiments.  (ii) ANU cannot be applied to multi-treatment settings, which have been extensively studied recently~\citep{kaddour2021causal,bica2020scigan,parbhoo2021ncore}. The comparison experiments are in Section~\ref{exp:ahu}.

\section{An Illustrative Example}
\label{sec:example}

\begin{wrapfigure}{r}{0.4\textwidth}
    \begin{center}
    %\centering
    \vspace{-10pt}
    \includegraphics[width=0.4\textwidth]{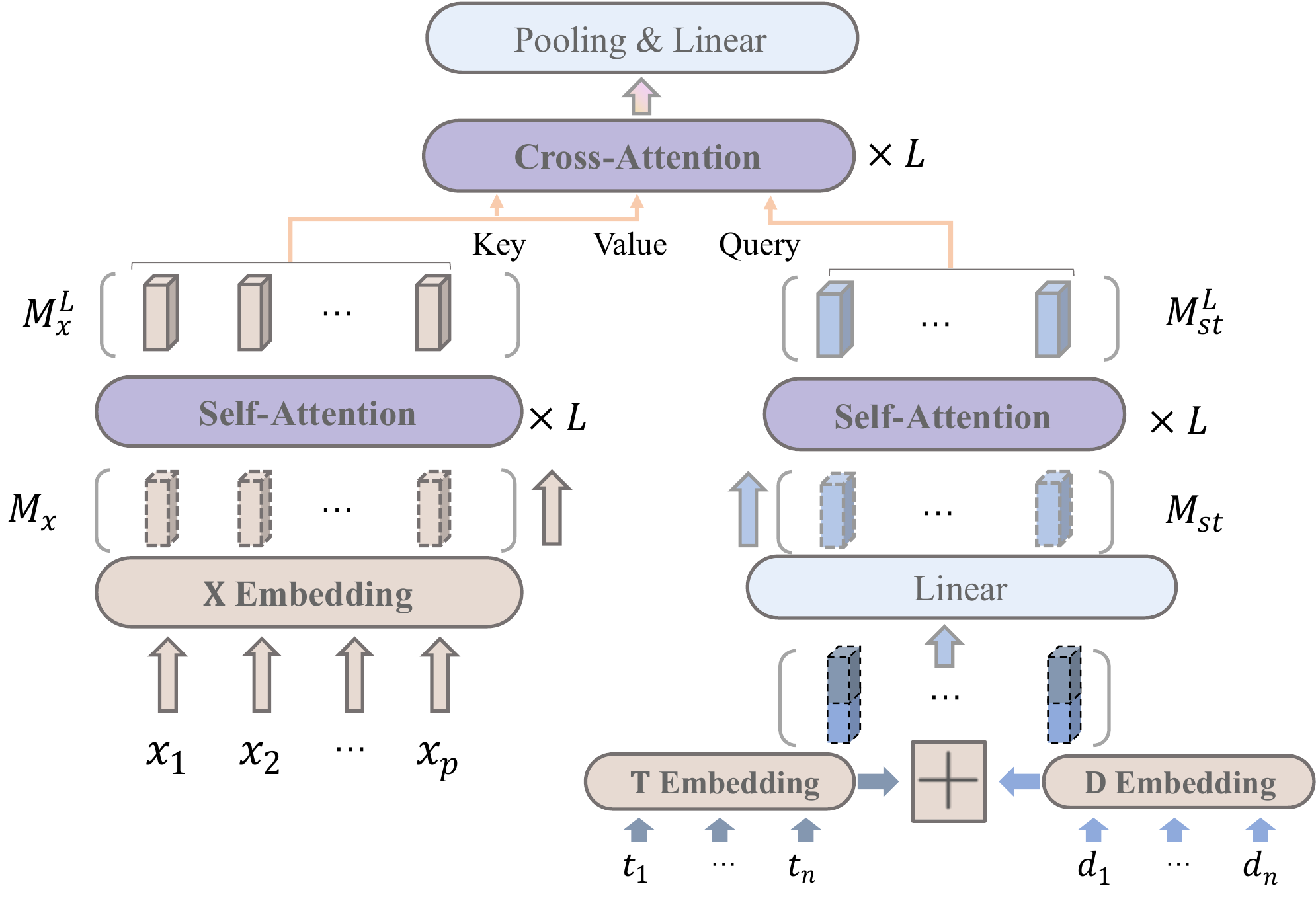}
    \end{center}
    \caption{An Illustrative Example about the workflow of \abbr.}
    \label{fig:illustration}
\end{wrapfigure}
To better understand the workflow with the above designs, we present a simple illustration here. Consider a use case in medicine effect estimation, where $\x$ contains $p$ patient information, \eg \textit{Age, Sex, Blood Pressure (BP), and Previous infection condition (Prev)} with a corresponding causal graph (Figure~\ref{fig:teaser1}). $n$ medicines (treatments) are applied simultaneously and each medicine has a corresponding dosage. As shown in \figurename~\ref{fig:illustration}, each covariate, treatment, and dosage will first be embedded to $d$-dimension representation by a specific \textbf{learnable embedding layer}. Each treatment embedding will be concatenated with its dosage embedding and the concatenated feature will be projected by a linear layer to produce $d$ dimensional vectors. 
\textbf{Self-attention modules} optimizes these embeddings by aggregating contextual information. Specifically, attribute \textit{Prev} is more related to \textit{age} than \textit{sex}, hence the attention weight of \textit{Prev} feature to \textit{age} feature is larger and the update of \textit{Prev} feature will be more dependent on the \textit{age} feature. Similarly, the interaction of multi-medicines is also attained by the self-attention module. The last \textbf{Cross-attention module} enables treatment-covariate interactions, which is shown in Figure \ref{fig:framework} that, each medicine will assign a higher weight to relevant covariates especially confounders (\textit{BP}) than irrelevant ones. Finally, we pool the resulting embedding and use one linear layer to predict the outcome.

\section{Details and Discussions about Propensity Score Modelling}
\label{app:discussion}
\input{arxiv/sections/05_propensity}

\section{Analysis of the Failure Cases over Treatment Distribution Shifts}\label{sec:distribution_shift}

As shown in~\figurename~\ref{fig:ADRF} (a,c), with the shifts of the treatment interval, the estimation performance of DRNet and TARNet decline significantly. VCNet achieves $\infty$ estimation error when $h=5$ partly because its hand-craft projection matrix can only process values near $[0,1]$. Another problem brought by this assumption is the extrapolation dilemma, which can be seen in~\figurename~\ref{fig:ADRF}(b). When training on $t\in[0,1.75]$, these discrete approximation methods cannot transfer to new distribution $t\in(1.75,2.0]$. These unseen treatments are rounded down to the nearest neighbors $t^\prime$ in $T$ and be seemed the same as $t^\prime$. We conduct ablation about the treatment embedding as in Table~\ref{tab:syn_emb} in Appendix. Such a simple fix (VCNet+Embeddings) removes the demand on a fixed interval constraint to treatments and attains superior performance on both interpolation and extrapolation settings. The result clearly shows the pitfalls of hand-crafted feature mapping for TEE. We highlight that it is neglected by most existing works \citep{schwab2020drnet, nie2021vcnet, shi2019dragonnet, guo2021cetransformer}. Extrapolation is still a challenging open problem. We can see that no existing work does well when training and test treatment intervals have big gaps. However, the empirical evidence validates the improved effectiveness of \abbr~ that uses learnable embeddings to map continuous treatments to hidden representations.

Below we show the assumption that the value of treatments or dosages are in a fixed interval $[l,h]$ is sub-optimal and thus these methods get poor extrapolation results. For simplicity, we only consider a data sample that has only one continuous treatment $t$ and the result is similar for continuous dosage.  
\label{app:fail}
\begin{prop}
Given a data sample $(\mathbf{x},t,y)$, where $\mathbf{x}\in\mathbb{R}^d,t\in[l,h],y\in \mathbb{R}$. Assume $\mu$ is a L-Lipschitz function over $(\mathbf{x},t)\in\mathbb{R}^{d+1}$, namely $|\mu(\mathbf{u})-\mu(\mathbf{v})|\leq L\|\mathbf{u}-\mathbf{v}\|$. Partitioning $[l,h]$ uniformly into $\delta$ sub-interval, and then get $T=\left[l+\frac{h-l}{\delta}*0,l+\frac{h-l}{\delta}*1,...,l+\frac{h-l}{\delta}*\delta\right]$. Previous studies most rounding down a treatment $t$ to its nearest value in $T$ (either $l+\left\lfloor\frac{t\delta}{h-l}\right\rfloor\frac{h-l}{\delta}$ or $l+\left\lceil\frac{t\delta}{h-l}\right\rceil\frac{h-l}{\delta}$) and use $|T|$ branches to approximate the entire continuum $[l,h]$. The approximation error can be bounded by
\begin{equation}
\begin{aligned}
& \max \left\{\mu\left(\mathbf{x},\left\lfloor\frac{t\delta}{h-l}\right\rfloor\frac{h-l}{\delta}\right)-\mu(\mathbf{x},t),\mu\left(\mathbf{x},\left\lceil\frac{t\delta}{h-l}\right\rceil\frac{h-l}{\delta}\right)-\mu(\mathbf{x},t)\right\}\\
&\leq \max \left\{L\left(\left|\left\lfloor\frac{t\delta}{h-l}\right\rfloor\frac{h-l}{\delta}-t\right|\right), L\left(\left|\left\lceil\frac{t\delta}{h-l}\right\rceil\frac{h-l}{\delta}-t\right|\right)\right\}\\
&\leq L\frac{h-l}{\delta}
\end{aligned}
\label{eq:prox}
\end{equation}
\label{prop1}
\end{prop}
The bound is affected by both the number of branches $\delta$ and treatment interval $[l,h]$. However, as far as we know, most previous works ignore the impacts of the treatment interval $[l,h]$ and adopt a simple but much stronger assumption that treatments are all in the interval $[0,1]$~\cite{nie2021vcnet} or a fixed interval~\cite{schwab2020drnet}. These observations well manifest the motivation of our general framework for TEE without the need for treatment-specific architectural designs.

\begin{table*}[h]
\caption{\textbf{Experimental results comparing NN-based methods on simulated datasets.} Numbers reported are AMSE of test data based on 100 repeats, and numbers after $\pm$ are the estimated standard deviation of the average value. For Extrapolation ($h=2$), models are trained with $t\in[0,1.75]$ and tested in $t\in[0,2]$. For Extrapolation ($h=5$), models are trained with $t\in[0,4]$ and tested in $t\in[0,5]$}
\label{tab:syn_emb}
\vspace{-0.1cm}
\begin{center}
\begin{small}
\begin{sc}
\resizebox{\textwidth}{!}{
\begin{tabular}{ccccc}
\toprule 
Methods & Vanilla & Vanilla ($h=5$) &Extrapolation ($h=2$) & Extrapolation ($h=5$) \\
\midrule
    TARNet~\citep{shalit2017tarnet}& 0.045 $\pm$ 0.0009& 0.3864 $\pm$ 0.04335& 0.0984 $\pm$ 0.02315 & 0.3647 $\pm$ 0.03626 \\
 DRNet~\citep{schwab2020drnet}& 0.042 $\pm$ 0.0009& 0.3871 $\pm$ 0.03851& 0.0885 $\pm$ 0.00094 & 0.3647 $\pm$ 0.03625\\
 VCNet\citep{nie2021vcnet}& 0.018 $\pm$ 0.0010& nan& 0.0669 $\pm$ 0.05227 & nan \\
 \rowcolor{Gray}
   VCNet+Embeddings& 0.013 $\pm$ 0.00465& 0.0167 $\pm$ 0.01150& 0.0118 $\pm$ 0.00482 &0.0178 $\pm$ 0.00887 \\
\bottomrule
\end{tabular}}
\end{sc}
\end{small}
\end{center}
\end{table*}
\vspace{-0.5cm}
\section{Additional Experimental Setups}
\label{app:settings}
All the assets (\ie datasets and the codes for baselines) we use include a MIT license containing a copyright notice and this permission notice shall be included in all copies or substantial portions of the software. We conduct all the experiments on a machine with i7-8700K CPU, 32G RAM, and four Nvidia GeForce RTX2080Ti (10GB) GPU cards.

\subsection{Detail Evaluation Metrics.}

\begin{equation}
\text{AMSE}_\mathcal{T}=\frac{1}{N}\sum_{i=1}^N\int_{\mathcal{T}}\left[\hat{f}(\mathbf{x}_i, t)-f(\mathbf{x}_i,t)\right]\pi(t)dt
\label{equ:amse}
\end{equation}
\begin{equation}
\begin{aligned}
&\text{UPEHE@K}=\frac{1}{N}\sum_{i=1}^N\bigg[\frac{1}{C^2_K}\sum_{t,t'}\left[\hat{f}(\mathbf{x}_i,t,t')-f(\mathbf{x}_n,t,t')\right]^2\bigg]\\
&\text{ WPEHE@K}=\frac{1}{N}\sum_{i=1}^N\bigg[\frac{1}{C^2_K}\sum_{t,t'}\left[\hat{f}(\mathbf{x}_i,t,t')-f(\mathbf{x}_i,t,t')\right]^2  p(t|\x)p(t'|\x)\bigg], 
\end{aligned}
\label{equ:pehe}
\end{equation}
\begin{equation}
\text{AMSE}_\mathcal{D}=\frac{1}{NT}\sum\limits_{i=1}^N\sum\limits_{t=1}^T\int_{\mathcal{D}}\left[\hat{f}(\mathbf{x}_i,t,s)-f(\mathbf{x}_n,t,s)\right]\pi(s)dt
\label{equ:amse_dosage}
\end{equation}
\subsection{Network Structure and Parameter Setting}
Table.~\ref{tab:arc} and Table.~\ref{tab:paras} show the detail of \abbr~architecture and hyper-parameters. For all the synthetic and semi-synthetic datasets, we tune parameters based on 20 additional runs.  In each run, we simulate data, randomly split it into training and testing, and use AMSE on testing data for evaluation. For fair comparisons, in all experiments, the model size of TransTEE is less than or similar to the baselines. 
\begin{table}[h]
\caption{\textbf{Architecture details of \abbr}, where $p$ is the number of covariates.}\label{tab:arc}
\centering

\begin{tabular}{@{}ccc@{}}
\toprule
Module           & Covariates   &  Treatment \\ \midrule
Embedding Layer  & $\left[\text{Linear}\right]$ & $\left[\text{Linear}\right]$ \\
Output Size      & $\text{Bsz}\times p\times \# \text{Emb}$                                                                                                                              & $bsz\times 1\times \#$ Emb   \\\hline
Self-Attention   & $\left[\begin{array}{c} \text{Multi-head Att}\\ \text{BatchNorm} \\ \text{Linear} \\ \text{BatchNorm} \end{array}\right] \times \# \text{Layers}$ & $\left[\begin{array}{c} \text{Multi-head Att}\\ \text{BatchNorm} \\ \text{Linear} \\ \text{BatchNorm} \end{array}\right] \times \# \text{Layers}$ \\
Output Size      & $\text{Bsz}\times p\times \# \text{Emb}$ & $\text{Bsz}\times 1\times \# \text{Emb}$ \\\hline

Cross-Attention  & \multicolumn{2}{c}{$\left[\begin{array}{c} \text{Multi-head Att}\\ \text{BatchNorm} \\ \text{Linear} \\ \text{BatchNorm} \end{array}\right] \times \# \text{Layers}$}\\
Output Size      & \multicolumn{2}{c}{$\text{Bsz}\times 1\times \# \text{Emb}$}   \\\hline
Projection Layer & \multicolumn{2}{c}{$\left[\text{Linear}\right]$}          \\
Output Size      & \multicolumn{2}{c}{$\text{Bsz}\times 1$}  \\ \bottomrule
\end{tabular}

\end{table}
\begin{table}[h]
\caption{\textbf{Hyper-parameters on different datasets}. Bsz indicates the batch size, $\#$ Emb indicates the embedding dimension, Lr. S indicates the scheduler of the learning rate (Cos is the cosine annealing Learning rate).}\label{tab:paras}
\centering
\begin{tabular}{@{}ccccccc@{}}
\toprule
Dataset & Bsz & \# Emb & \# Layers & \# Heads & Lr     & Lr. S \\ \midrule
Simu    & 500        & 10               & 1         & 2        & 0.01   & Cos          \\
IHDP    & 128        & 10               & 1         & 2        & 0.0005 & Cos          \\
News    & 256        & 10               & 1         & 2        & 0.01   & Cos          \\
SW      & 500        & 16               & 1         & 2        & 0.01   & None         \\
TCGA    & 1000       & 48               & 3         & 4        & 0.01   & None         \\ \bottomrule
\end{tabular}

\end{table}

\subsection{Simulation details.}\label{sec:data_generating}

\textbf{Synthetic Dataset}~\citep{nie2021vcnet}. The synthetic dataset contains 500 training points and 200 testing points. Data is generated as follows: $x_j\sim \text{Unif}[0,1]$, where $x_j$ is the $j$-th dimension of $x\in\mathbb{R}^6$, and
\begin{equation}
\begin{aligned}
&\tilde{t}|x=\frac{10\sin\left(\max(x_1,x_2,x_3)\right)+\max(x_3,x_4,x_5)^3}{1+(x_1+x_5)^2}+\sin(0.5x_3)\left(1+\exp(x_4-0.5x_3)\right)+\\
&\quad\quad\quad\quad\quad x_3^2+2\sin(x_4)+2x_5-6.5+\mathcal{N}(0,0.25)\\
&y|x,t=\cos(2\pi(t-0.5))\left(t^2+\frac{4\max(x_1,x_6)^3}{1+2x_3^2}\right)+\mathcal{N}(0,0.25)
\end{aligned}\nonumber
\end{equation}

where $t=(1+\exp(-\tilde{t}))^{-1}$. 

for treatment in $[0,h]$, we revised it to $t=(1+\exp{-\tilde{t}})^{-1}*h$,
 
\textbf{IHDP}~\citep{hill2011bayesian} is a semi-synthetic dataset containing 25 covariates, 747 observations and binary treatments. For treatments in $[0,1]$, we follow VCNet~\citep{nie2021vcnet} and generate treatments and responses by: 
$$\tilde{t}|x=\frac{2x_1}{1+x_2}+\frac{2\max(x_3,x_5,x_6)}{0.2+\min(x_3,x_5,x_6)}+2\tanh\left(5\frac{\sum_{i\in S_{dis,2}}(x_i-c_2)}{|S_{dis,2}|}-4+\mathcal{N}(0,0.25)\right)$$
$$y|x,t=\frac{\sin(3\pi t)}{1.2-t}\left(\tanh\left(5\frac{\sum_{i\in S_{dis,1}}(x_i-c_1)}{|S_{dis,1}|}\right)+\frac{\exp(0.2(x_1-x_6))}{0.5+5\min(x_2,x_3,x_5)}\right)+\mathcal{N}(0,0.25),$$
where $t=(1+\exp(-\tilde{t}))^{-1}$, $S_{con}=\{1,2,3,5,6\}$ is the index set of continuous features, $S_{dis,1}=\{4,7,8,9,10,11,12,13,14,15\}$, $S_{dis,2}=\{16,17,18,19,20,21,22,23,24,25\}$ and $S_{dis,1}\bigcup S_{dis,2}=[25]-S_{con}$. Here $c_1=\mathbb{E}\left[\frac{\sum_{i\in S_{dis,1}} x_i}{|S_{dis,1}|}\right]$,$c_2=\mathbb{E}\left[\frac{\sum_{i\in S_{dis,2}} x_i}{|S_{dis,2}|}\right]$. To allow comparison on various treatment intervals $t\in[0,h]$, treatments and responses are generated by:
$$t=(1+\exp(-\tilde{t}))^{-1}*h$$
$$y|x,t=\frac{\sin(3\pi {\color{orange}t/h})}{1.2-{\color{orange}t/h}}\left(\tanh\left(5\frac{\sum_{i\in S_{dis,1}}(x_i-c_1)}{|S_{dis,1}|}\right)+\frac{\exp(0.2(x_1-x_6))}{0.5+5\min(x_2,x_3,x_5)}\right)+\mathcal{N}(0,0.25),$$
where the \textcolor{orange}{orange} part is the only different compared to the generalization of vanilla IHDP dataset ($h=1$). Note that $S_{dis,1}$ only impacts outcome that serves to be noisy covariates; $S_{dis,2}$ contains pre-treatment covariates that only impact treatments, which also serve to be instrumental variables. This allows us to observe the improvement using \abbr~when noisy covariates exist. Following~\citep{hill2011bayesian} covariates are standardized with mean $0$ and standard deviation $1$.

\textbf{News.} The News dataset consists of 3000 randomly sampled news items from the NY Times corpus~\citep{newman2008bag} and was originally introduced as a benchmark in the binary treatment setting. We generate the treatment and outcome in a similar way as~\citep{nie2021vcnet} but with a dynamic range or treatment intervals $[0,h]$. We first generate $v_1^\prime,v_2^\prime,v_3^\prime\sim\mathcal{N}(0,1)$ and then set $v_i=v_i^\prime/\|v_i^\prime\|_2; i\in\{1,2,3\}$. Given $x$, we generate $t$ from $\text{Beta}\left(2,\left|\frac{v_3^\top x}{2v_2^\top x}\right|\right)*h$.And we generate the outcome by
$$y^\prime|x,t=\exp\left(\frac{v_2^\top x}{v_3^\top x}-0.3\right)$$
$$y|x,t=2(\max(-2,\min(2,y^\prime))+20v_1^\top x)*\left(4(t-0.5)^2+\sin\left(\frac{\pi}{2}t\right)\right)+\mathcal{N}(0,0.5)$$

\textbf{TCGA (D)}~\citep{bica2020scigan} We obtain covariates $x$ from a real dataset \textit{The Cancer Genomic Atlas (TCGA)} and consider $3$ treatments, where each treatment is accompanied by one dosage and a set of parameters, $v_1^t,v_2^t,v_3^t$. For each run, we randomly sample a vector, $u_i^t\sim\mathcal{N}(0,1)$ and then set $v_i^t=u_i^t/\|u_i^t\|$ where $\|\cdot\|$ is Euclidean norm. The shape of the response curve for each treatment, $f_t(x,s)$ is given in Table~\ref{tab:outcome}. We add $\epsilon\sim\mathcal{N}(0,0.2)$ noise to the outcomes. 
Interventions are assigned by sampling a dosage, $d_t$, for each treatment from a beta distribution, $d_t|x\sim \text{Beta}(\alpha,\beta_t)$. $\alpha\geq1$ controls the dosage selection bias ($\alpha=1$ gives the uniform distribution). $\beta_t=\frac{\alpha-1}{s_t^*}+2-\alpha$, where $s_t^*$ is the optimal dosage\footnote{For symmetry, if $s_t^*=0$, we sample  $s_t^*$ from
$1-$Beta$(\alpha,\beta_t)$ 
where $\beta_t$ is set as though $s_t^*=1$.} for treatment $t$. 
We then assign a treatment according to $t_f|x\sim\text{Categorical(Softmax(}\kappa f(x,s_t)))$ 
where increasing $\kappa$ increases selection bias, and $\kappa=0$ leads to random assignments. The factual intervention is given by $(t_f,s_{t_f})$. Unless otherwise specified, we set $\kappa=2$ and $\alpha=2$.

\begin{figure*}[t]
    \centering
    \begin{minipage}[t]{\textwidth}        
    \subfigure[Performance with different dosage selection bias.]
	{\includegraphics[width=0.5\textwidth]{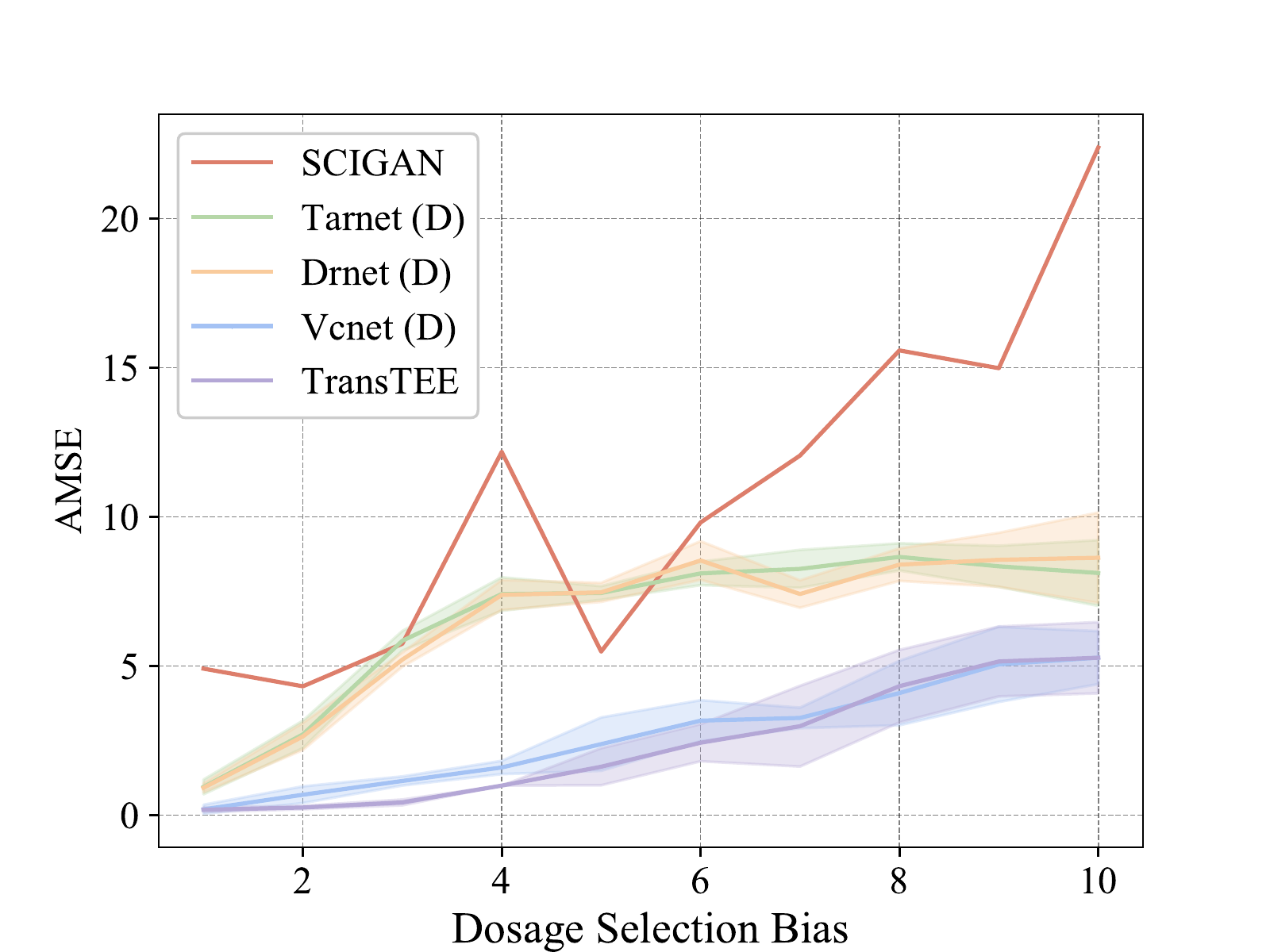}}
	\subfigure[Performance with different treatment selection bias.]
	{\includegraphics[width=0.5\textwidth]{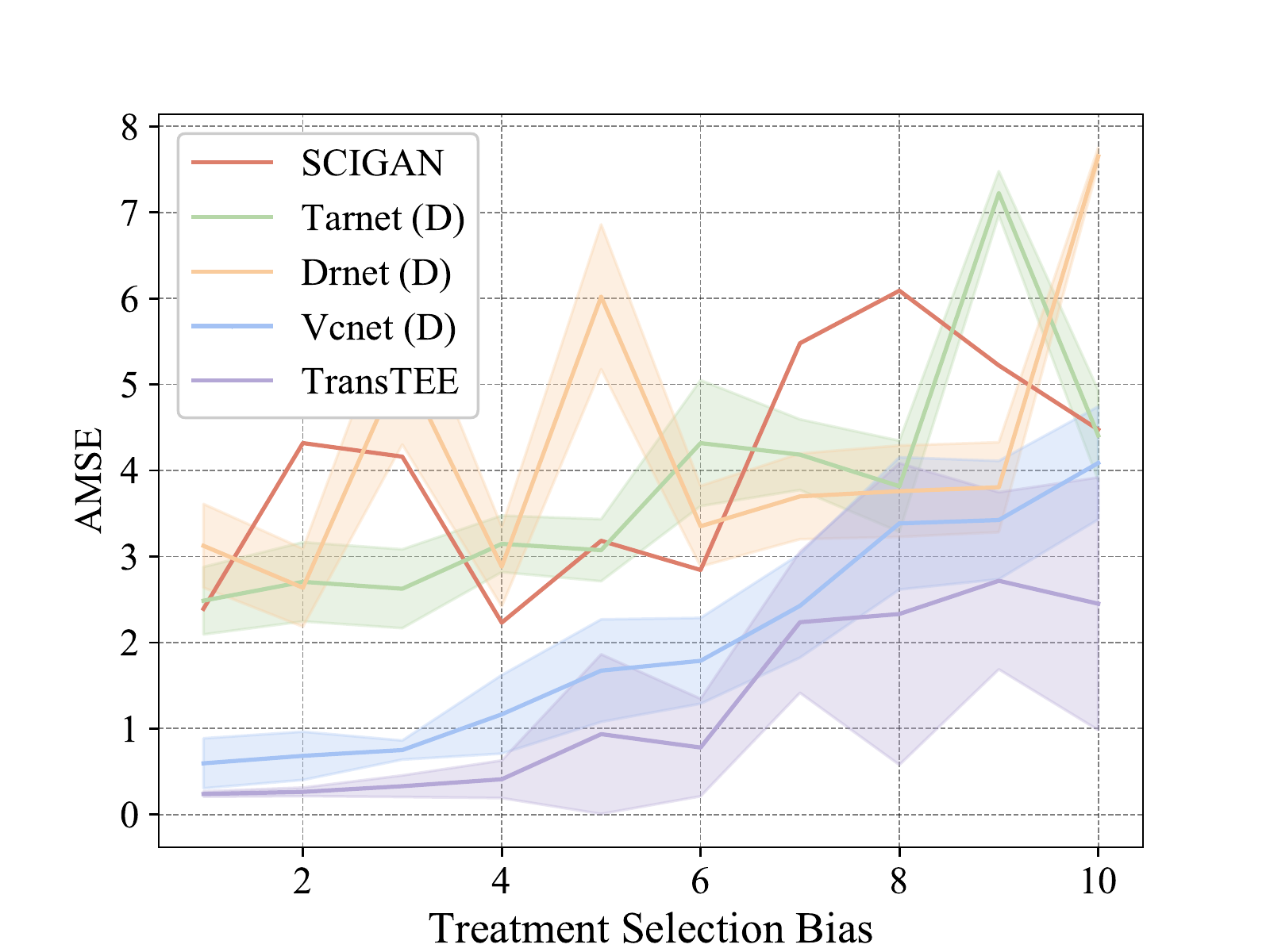}}
    \end{minipage}
    \caption{Performance of five methods on TCGA (D) dataset with varying bias levels. }
    \label{fig:dosage_bias}
\end{figure*}

\begin{figure*}[h]
\centering
\begin{minipage}[t]{\textwidth}        
\subfigure[Estimated ADRF for $t_1$.]
{\includegraphics[width=0.32\textwidth]{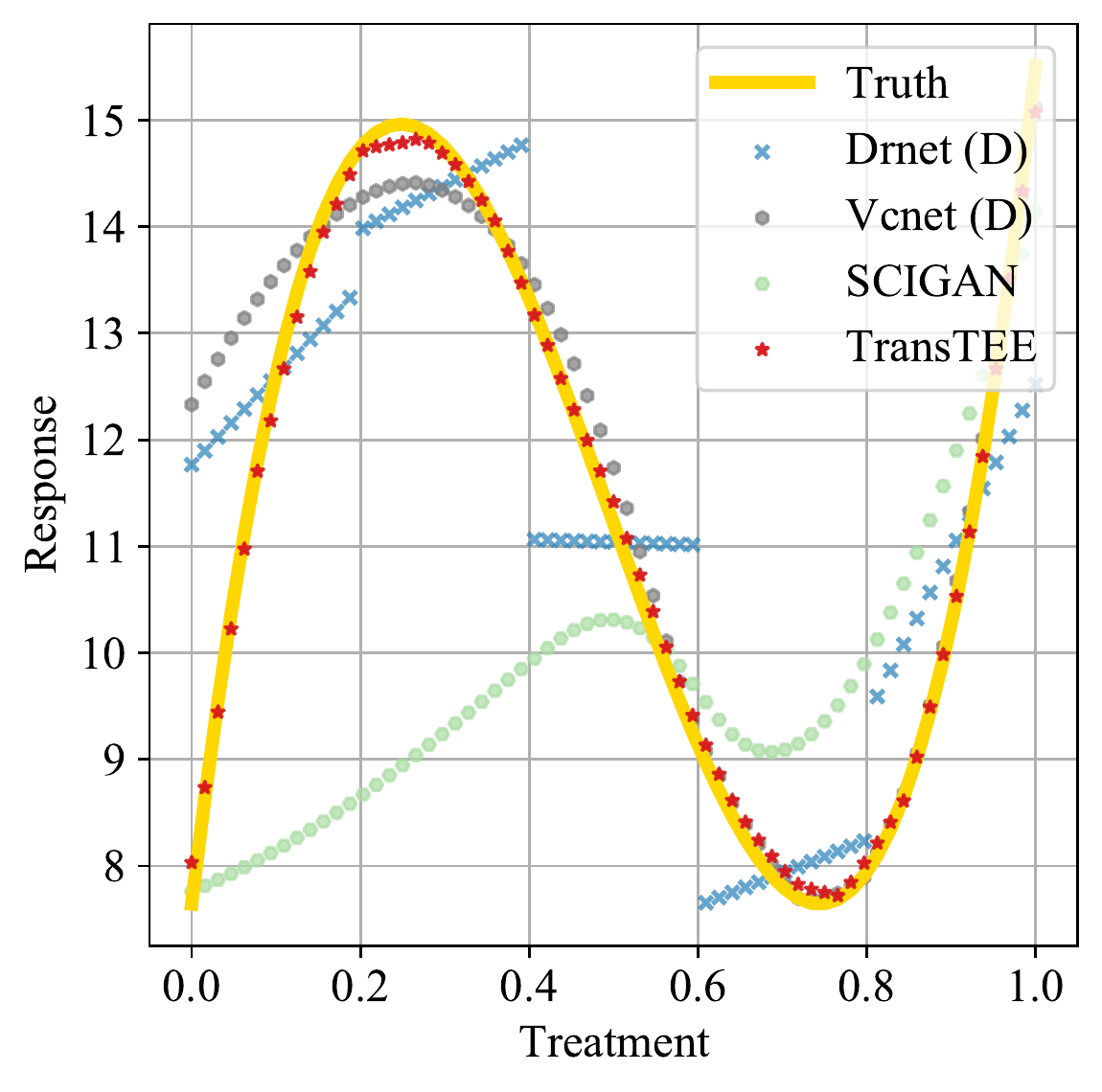}}
\subfigure[Estimated ADRF for $t_2$.]
{\includegraphics[width=0.32\textwidth]{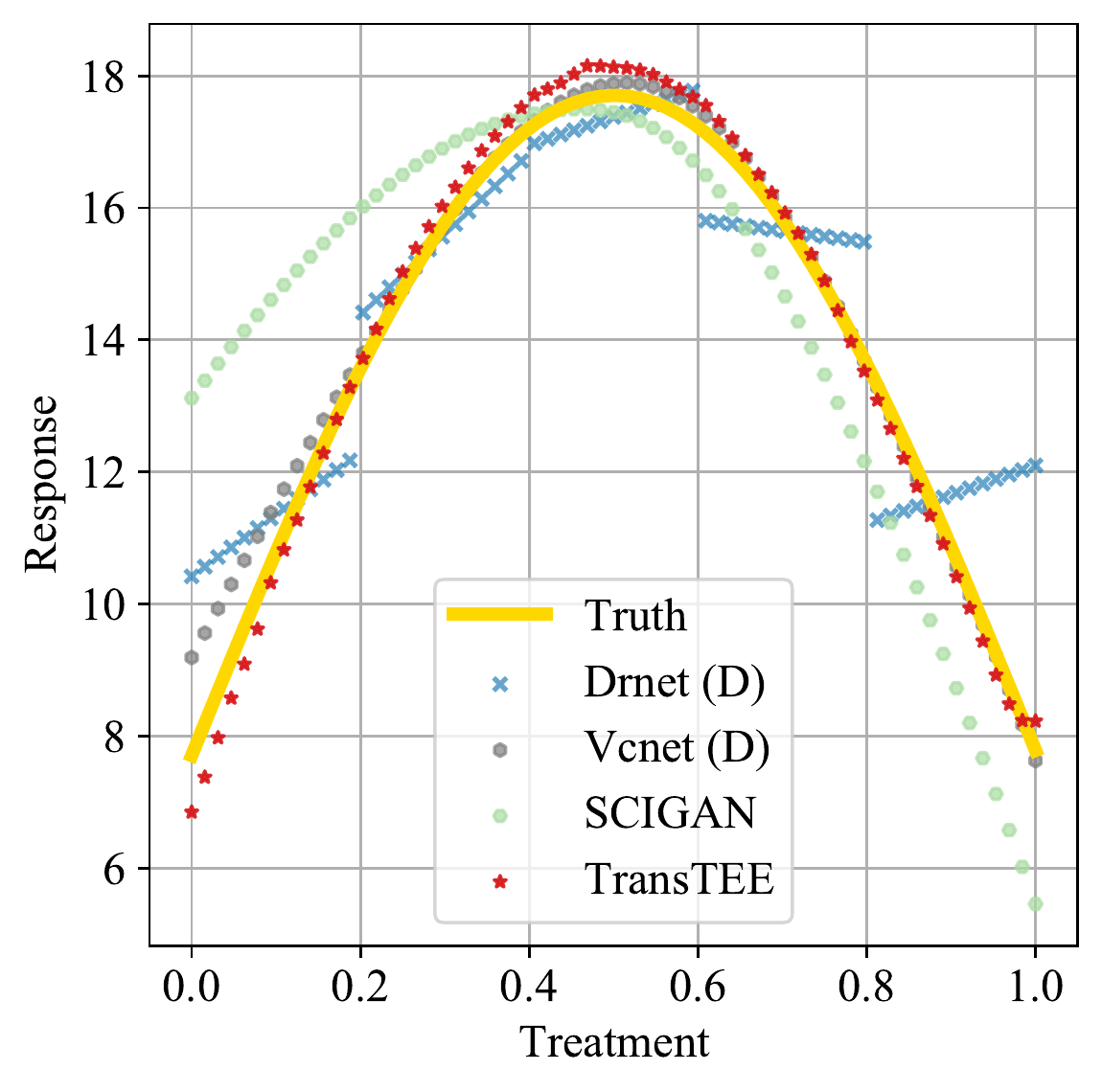}}
\subfigure[Estimated ADRF for $t_3$.]
{\includegraphics[width=0.32\textwidth]{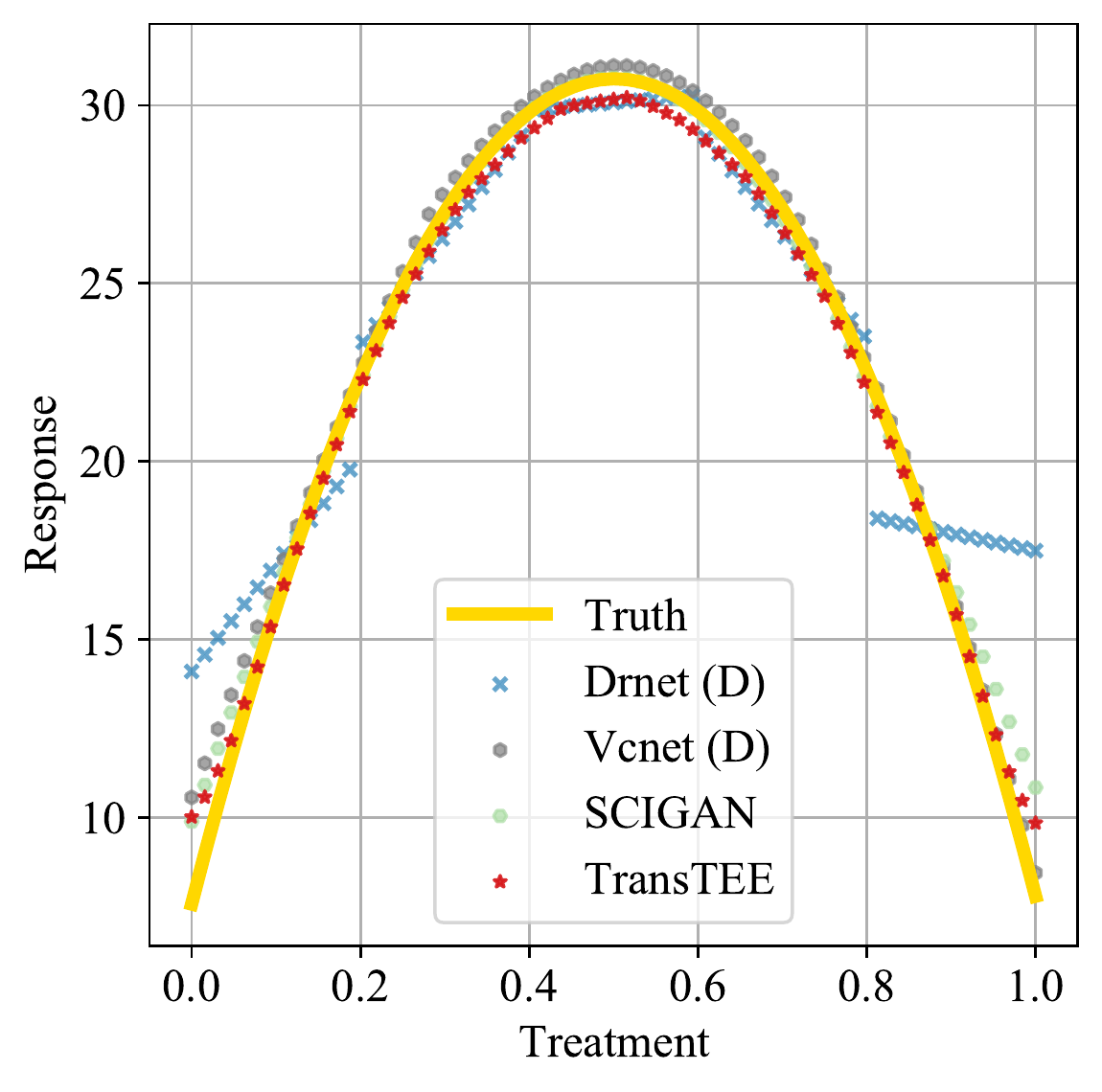}}
\end{minipage}
\caption{\textbf{Estimated ADRF} on the test set from a typical run of DRNet (D), TARNet (D), VCNet (D), and SCIGAN. All of these methods are well optimized. \abbr~can well estimate the dosage-response curve for all treatments.}\label{fig:dosage_adrf}
\end{figure*}

\begin{table*}[]
\caption{\textbf{Dose response curves used to generate semi-synthetic outcomes for patient features $x$}. In the experiments, we set $C= 10$. $v_1^t,v_2^t,v_3^t$ are the parameters associated with each treatment $t$.}\label{tab:outcome}
\centering
\begin{tabular}{@{}ccc@{}}
\toprule
Treatment & Dose-Response & Optimal dosage \\ \midrule
1         &      $f_1(x,s)=C\left((v_1^1)^\top x+12(v_3^1)^\top xs-12(v_3^1)^\top xs^2 \right)$        & $s_1^*=\frac{(v_2^1)^\top x}{2(v_3^1)^\top x}$               \\
2         &    $f_2(x,s)=C\left((v_1^2)^\top x+\sin\left(\pi(\frac{v_2^{2\top}x}{v_3^{2\top}x}s)\right)\right)$           &     $s_2^*=\frac{(v_3^2)^\top x}{2(v_2^2)^\top x}$           \\
3         &   $f_3(x,s)=C\left((v_1^3)^\top x+12s(s-b)^2, \text{where } b=0.75\frac{(v_2^3)^\top x}{(v_3^3)^\top x}\right)$  &               $\frac{b}{3}\;\text{ if }b\geq 0.75\text{ else } 1$\\ \bottomrule
\end{tabular}
\end{table*}

For structural treatments, we first define the \textbf{Baseline effect}~\citep{bica2020scigan}. For each run of the experiment, we randomly sample a vector $u_0\sim\text{Unif}[0,1]$, and set $v_0=u_0/\|u_o\|$, where $\|\cdot\|$ is the Euclidean norm. The baseline effect is defined as
$$\mu_0(x)=v_0^\top x$$

\textbf{Small-World}~\citep{kaddour2021causal}. $20$-dimensional multivariate covariates are uniformly sampled according to $x_i\sim \text{Unif}[-1,1]$. There are $1,000$ units in the in-sample dataset and $500$ in the out-sample one.
\textit{Graph interventions} For each graph intervention, a number of nodes between $10$ and $120$ are uniformly sampled, the number of neighbors for each node is between $3$ and $8$, and the probability of rewiring each edge is between $0.1$ and $1$. Watts–Strogatz small-world graphs are repeatedly generated until a connected one is get. Each vertex has one feature, i.e. its degree of centrality. A graph’s node connectivity is denoted as $\nu(\mathcal{G})$ and its average shortest path length as $\ell(\mathcal{G})$. 
Similar for the baseline effect, two randomly sampled vectors $v_\nu,v_\ell$ are generated. Then, given an assigned graph treatment $\mathcal{G}$ and a covariate vector $x$, the \textit{outcome} is generated by
$$y=100\mu_0(x)+0.2\nu(\mathcal{G})^2\cdot v_\nu^\top x+\ell(\mathcal{G})\cdot\nu_\ell^\top x+\epsilon, \epsilon\sim\mathcal{N}(0,1)$$

\textbf{TCGA (S)}~\citep{kaddour2021causal} We use $9,659$ gene expression measurements of cancer patients for covariates. The in-sample and datasets consist of $5,000$ units and the out-sample one of $4,659$ units, respectively. Each unit is a covariate vector $x\in\mathbb{R}^{4000}$ and these units are split randomly into in- and out-sample datasets in each run randomly. For each covariate vector $x$, its $8$-dimensional PCA components $x^{\text{PCA}}\in\mathbb{R}^8$ is computed. \textit{Graph interventions} We randomly sample $10,000$ molecules from the Quantum Machine 9 (QM9) dataset~\citep{ramakrishnan2014quantum} (with $133$k molecules in total) in each run. We create a relational graph, where each node corresponds to an atom and consists of $78$ atom features. We label each edge corresponding to the chemical bond types, \eg single, double, triple, and aromatic bonds. We collect $8$ molecule properties $mu, alpha, homo, lumo, gap, r2, zpve, u0$ in a vector $z\in\mathbb{R}^8$, which is denoted as the assigned molecule treatment.  Finally, we generate \textit{outcomes} by
$$y=10\mu_0(x)+0.01z^\top x^{\text{PCA}}+\epsilon,\epsilon\sim\mathcal{N}(0,1)$$

\textbf{Enriched Equity Evaluation Corpus (EEEC)}~\citep{feder2021causalm} consists of $33,738$ English sentences and the label of each sentence is the mood state it conveys. The task is also known as Profile of Mood States (POMS). Each sentence in the dataset is created using one of $42$ templates, with placeholders for a person’s name and the emotion, \eg\textit{``$<$Person$>$ made me feel $<$emotional state word$>$.''}. A list of common names that are tagged as male or female, and as African-American or European will be used to fill the placeholder (\textit{$<$Person$>$}). One of four possible mood states: \textit{Anger}, \textit{Sadness}, \textit{Fear} and \textit{Joy} is used to fill the emotion placeholder. Hence, EEEC has two kinds of counterfactual examples, which are \textit{Gender} and  \textit{Race}. For the \textit{Gender} case, it changes the name and the \textit{Gender} pronouns in the example and switches them, such that for the original example: \textit{"It was totally unexpected, but {Roger} made me feel pessimistic."} it will have the counterfactual example:\textit{``It was totally unexpected, but {Amanda} made me feel pessimistic.''} For the \textit{Race} concept, it creates counterfactuals such that for the original example \textit{``Josh made me feel uneasiness for the first time ever in my life.''}, the counterfactual example is: \textit{``Darnell made me feel uneasiness for the first time ever in my life.''}. For each counterfactual example, the person’s name is taken at random from the pre-existing list corresponding to its type.

\begin{figure*}[t]
    \centering
    \subfigure[Ablation study of PTR.]{
\begin{minipage}[t]{0.45\linewidth}
\centering
\includegraphics[width=0.7\textwidth]{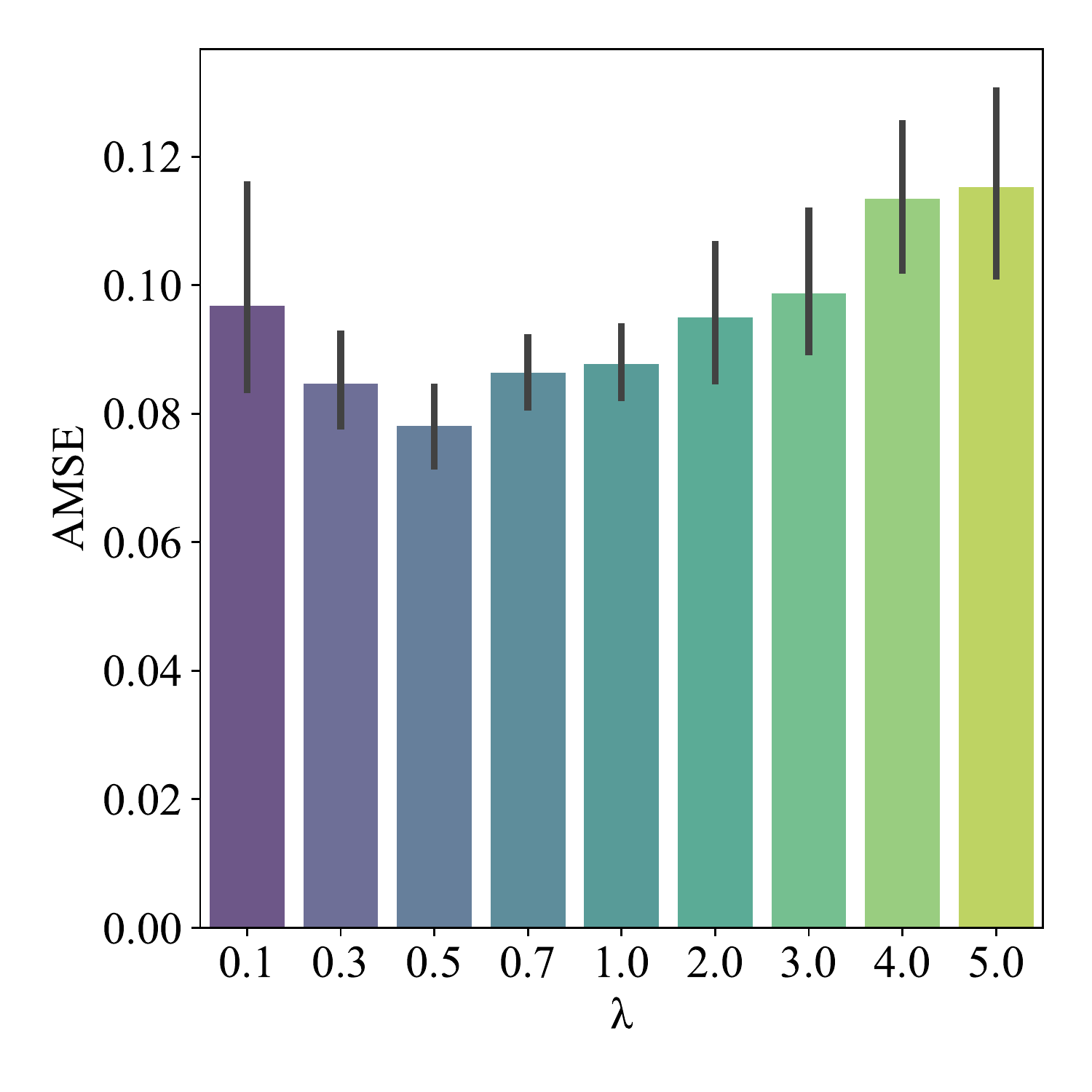}
%\caption{fig1}
\end{minipage}
}
\subfigure[Ablation study of TR.]{
\begin{minipage}[t]{0.45\linewidth}
\centering
\includegraphics[width=0.7\textwidth]{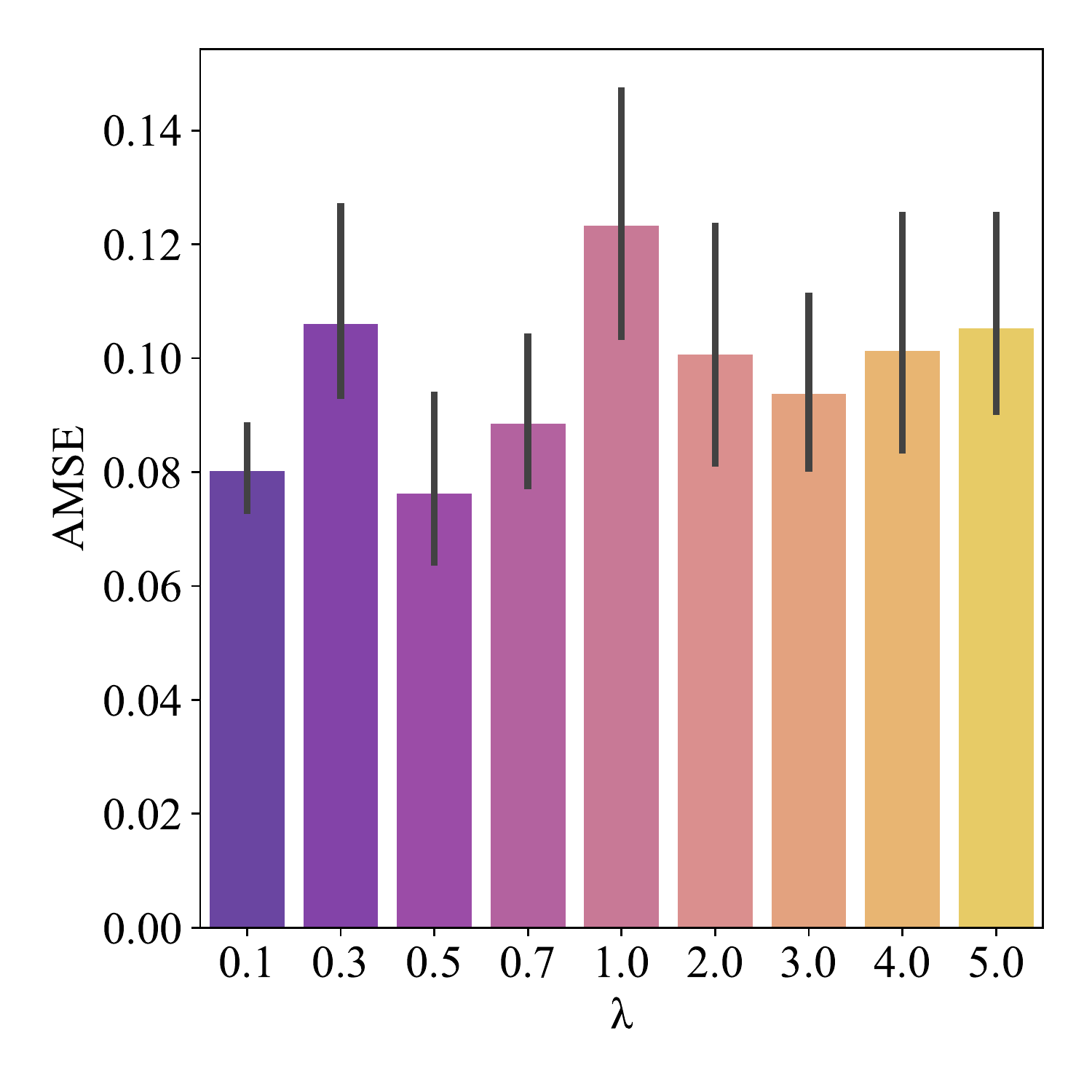}
%\caption{fig1}
\end{minipage}%
}%

\caption{Ablation study of the balanced weight for treatment regularization on the IHDP dataset.}\label{fig:ablation_lambda}
\end{figure*}

\section{Additional Experimental Results}\label{app:exp}

\subsection{Comparision between \abbr~and ANU~\citep{xu2022learning}}\label{exp:ahu}
We implement ANU and evaluate it in the same settings and show that is inferior compared to the proposed TransTEE as follows. Specifically, we compare the attentive neural uplift model (ANU)~\citep{xu2022learning} with ours in the following two settings. (1) IHDP dataset in Table~\ref{ahu:ihdp} in the main manuscript. We adjust the layers of ANU such that the total parameters of ANU and TransTEE are similar. The result is shown in the following table. With the usage of treatment embeddings, ANU is shown to be more robust than VCNet and DRNet when a treatment shift occurs. However, in both the binary treatment setting and continuous treatment settings, TransTEE performs better than ANU.

\begin{table}[]
\centering
\caption{Comparision between \abbr~and ANU~\citep{xu2022learning} on the IHDP dataset.}\label{ahu:ihdp}
\begin{tabular}{@{}cccc@{}}
\toprule
Methods  & \textbf{Vanilla (Binary)} & \textbf{Vanilla (h = 1)} & \textbf{Extrapolation (h = 2)} \\ \midrule
DRNet    & 0.3543 ± 0.6062           & 2.1549 ± 1.04483         & 11.071 ± 0.9938                \\
VCNet    & 0.2098 ± 0.18236          & 0.7800 ± 0.6148          & NAN                            \\
ANU~\citep{xu2022learning}  & 0.1482 ± 0.17362          & 0.2147± 0.32451          & 0.4244 ± 0.19832               \\
TransTEE & 0.0983 ± 0.15384          & 0.1151 ± 0.1028          & 0.2745 ± 0.1497                \\ \bottomrule
\end{tabular}%
\end{table}

(2) We further evaluate the real-world utility of ANU~\citep{xu2022learning} and the experimental setting is detailed in Section~\ref{sec:lm} in the main paper. Covariates here are long sentences. Thanks to the use of self-attention modules, TransTEE can achieve better estimation results compared to baselines (Table~\ref{ahu:eecs}). For AHU, no self-attention layer is applied, and the final estimation is inaccurate, which verifies the superiority of the proposed framework.

\begin{table}[]
\centering
\caption{Comparision between \abbr~and ANU~\citep{xu2022learning} on the IHDP dataset.}\label{ahu:eecs}
\resizebox{\columnwidth}{!}{%
\begin{tabular}{cccccc|cccc}
\toprule
       & \multicolumn{5}{c}{\textbf{Correlation/Representation Based Baselines}} & \multicolumn{4}{c}{\textbf{Treatment Effect Estimators}} \\\hline
TC     & ATE$_{GT}$      & TReATE      & CONEXP     & INLP      & CausalBERT     & TarNet     & DRNet      & ANU     & TransTEE     \\
Gender & 0.086           & 0.125       & 0.02       & 0.313     & 0.179          & 0.0067     & 0.0088     & 0.184           & 0.013        \\
Race   & 0.014           & 0.046       & 0.08       & 0.591     & 0.213          & 0.005      & 0.006      & 0.093           & 0.0174     \\\bottomrule 
\end{tabular}%
}
\end{table}

\subsection{Additional Numerical Results and Ablation Studies}

\begin{figure*}[t]
\centering
\begin{minipage}[ht]{.98\textwidth}        
\subfigure[\abbr.]
{\includegraphics[width=0.33\textwidth]{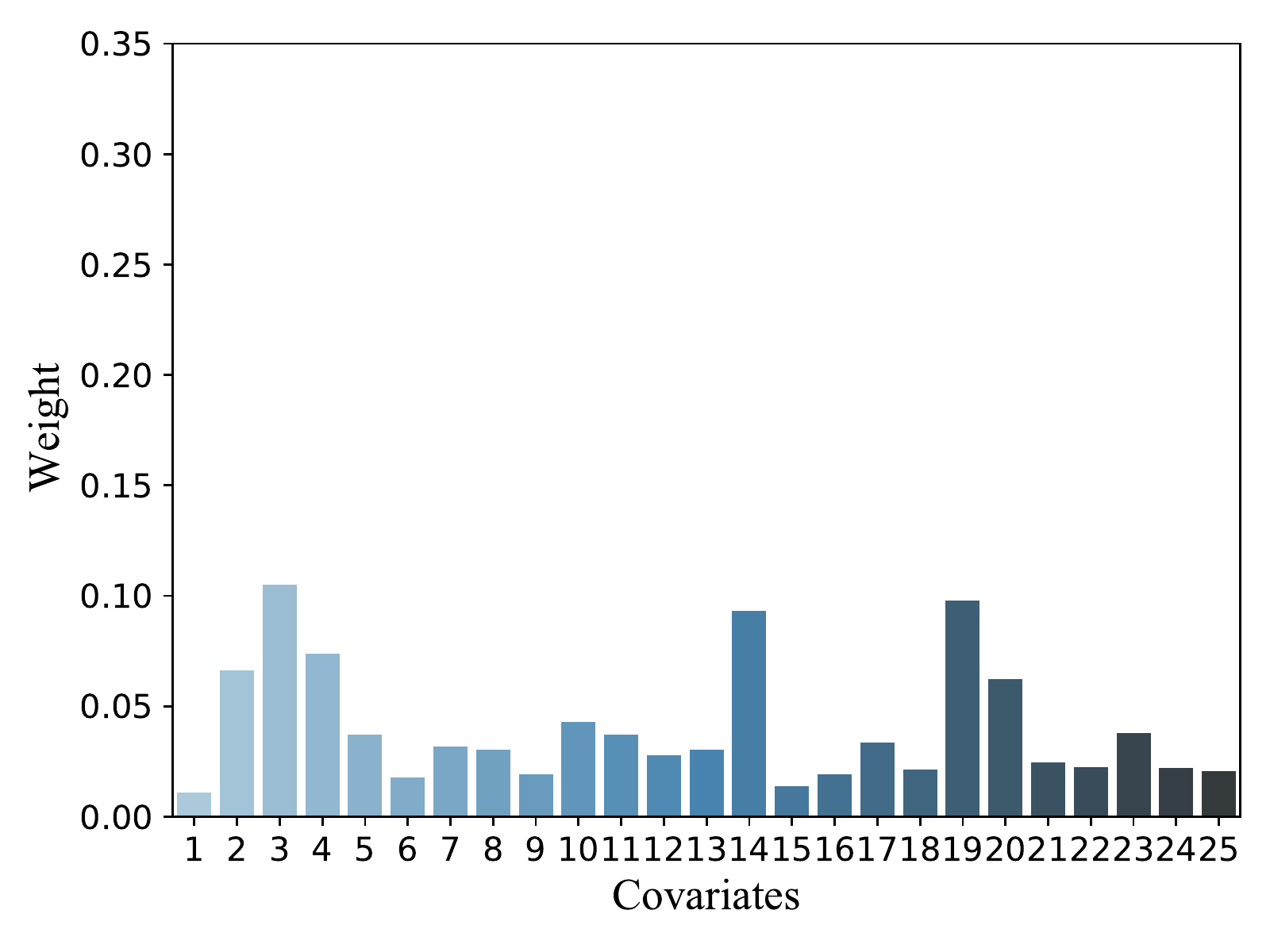}}
\subfigure[\abbr+TR.]
{\includegraphics[width=0.33\textwidth]{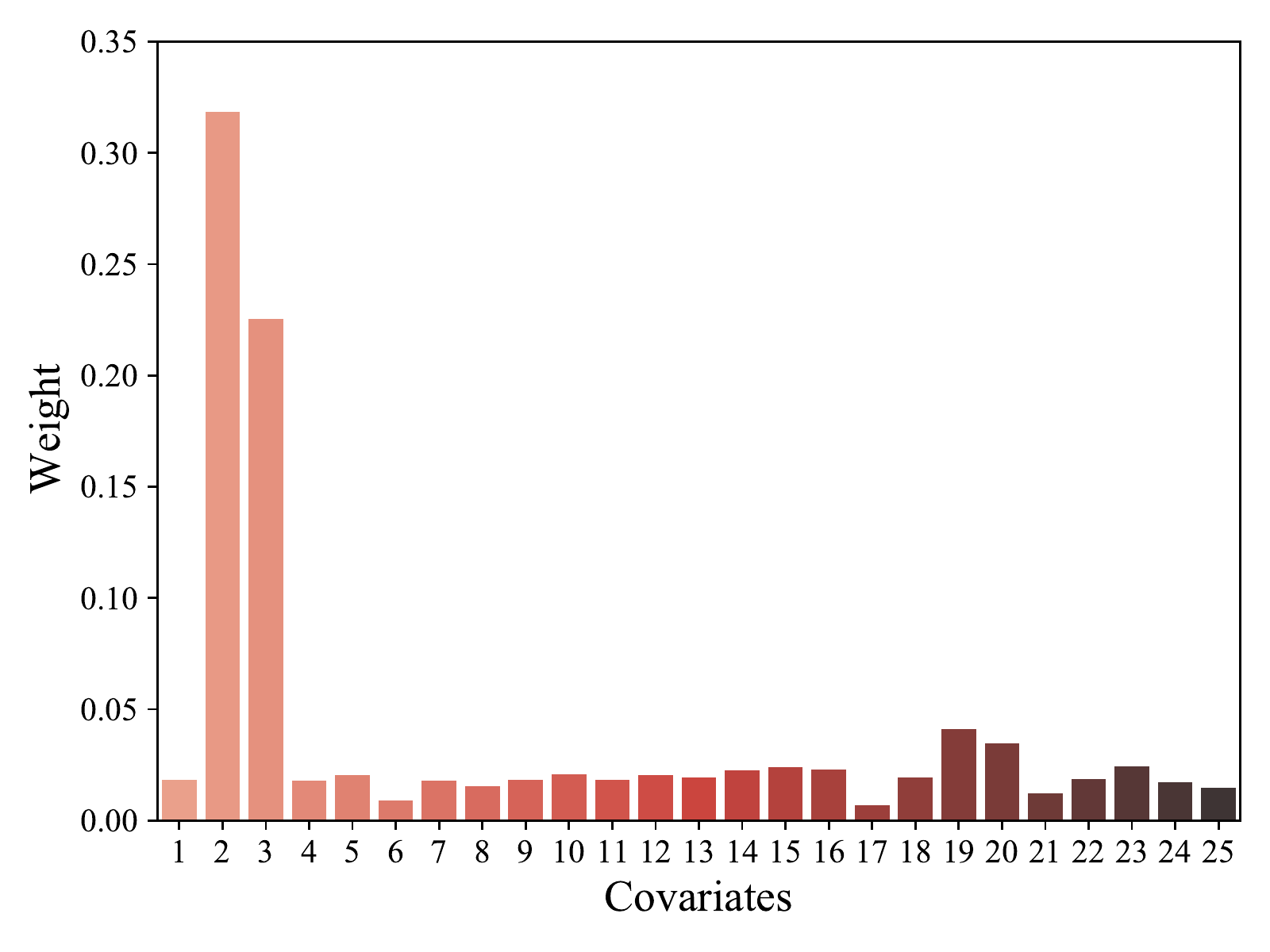}}
\subfigure[\abbr+PTR]
{\includegraphics[width=0.33\textwidth]{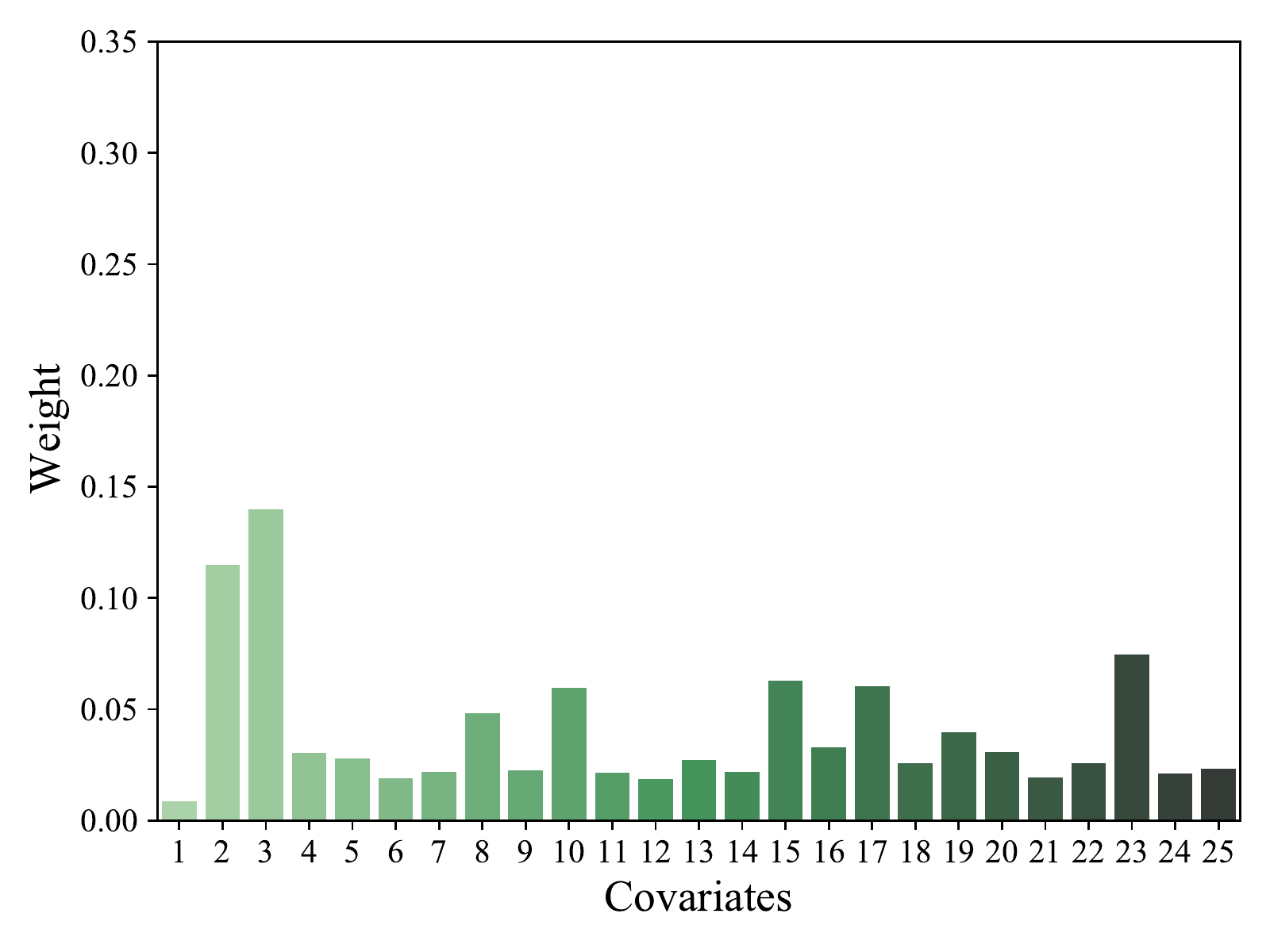}}
\end{minipage}
\caption{The distribution of learned weights for the cross-attention module on the IHDP dataset of different models. }\label{fig:attention_weights}
\end{figure*}

\begin{figure*}[h]
    \centering
    \begin{minipage}[t]{1\textwidth}        
    \subfigure[$h=1$ during training and testing.]
	{\includegraphics[width=0.24\textwidth]{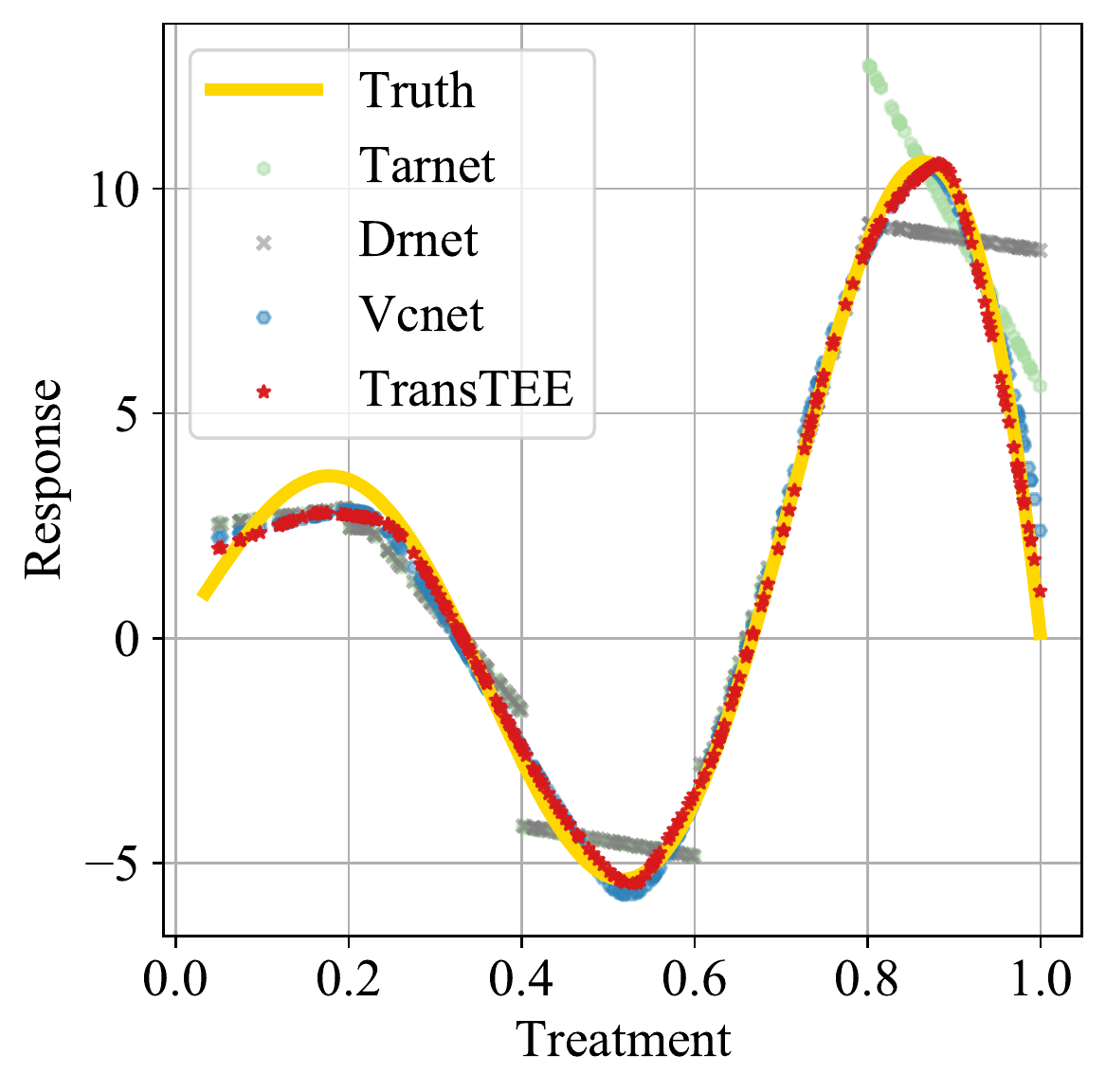}}
	\subfigure[$h=2,l=0.1$ during traning and $h=2,l=0$ during testing (extrapolation).]
	{\includegraphics[width=0.24\textwidth]{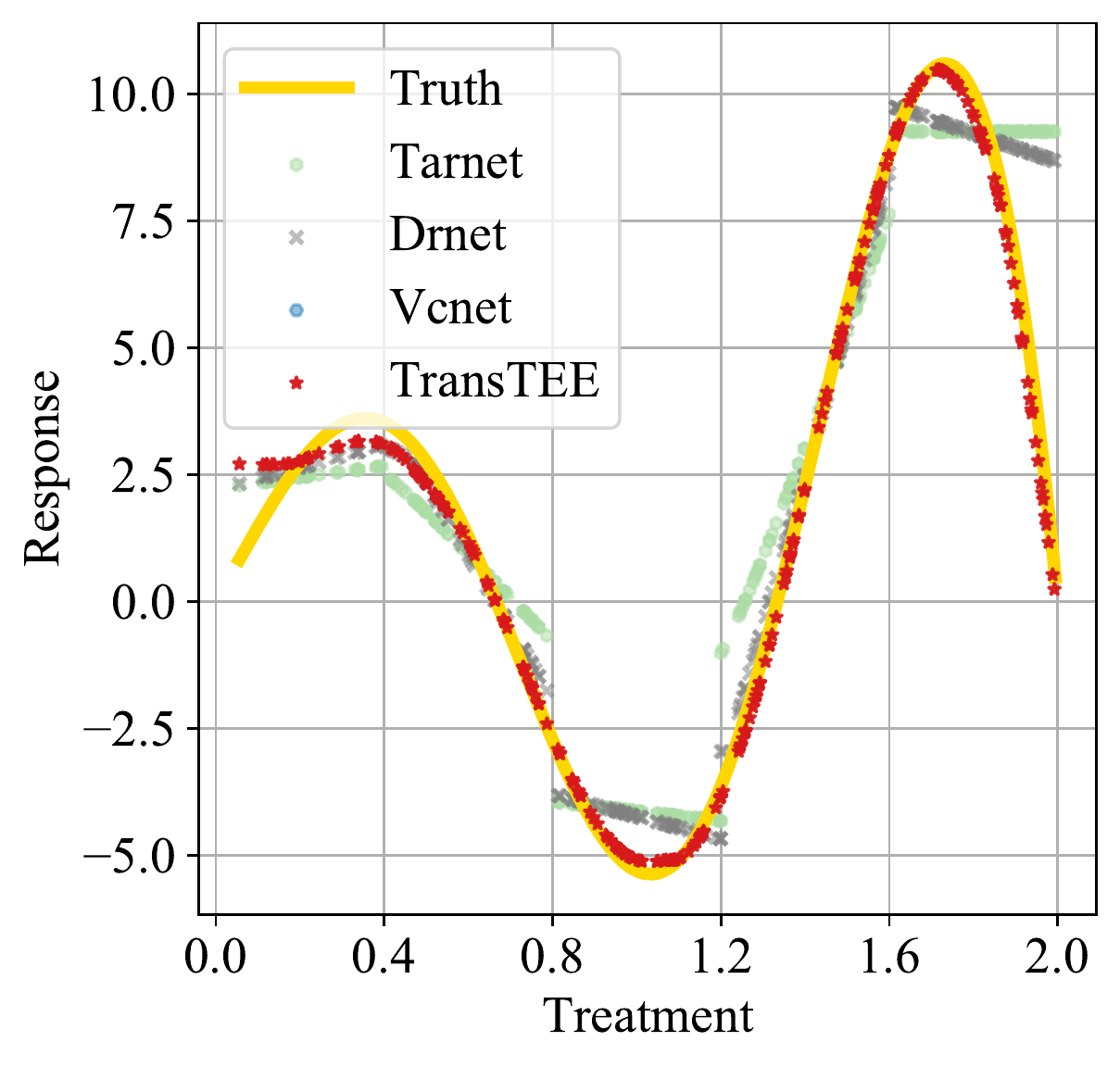}}
	\subfigure[$h=5$ during training and testing.]
	{\includegraphics[width=0.24\textwidth]{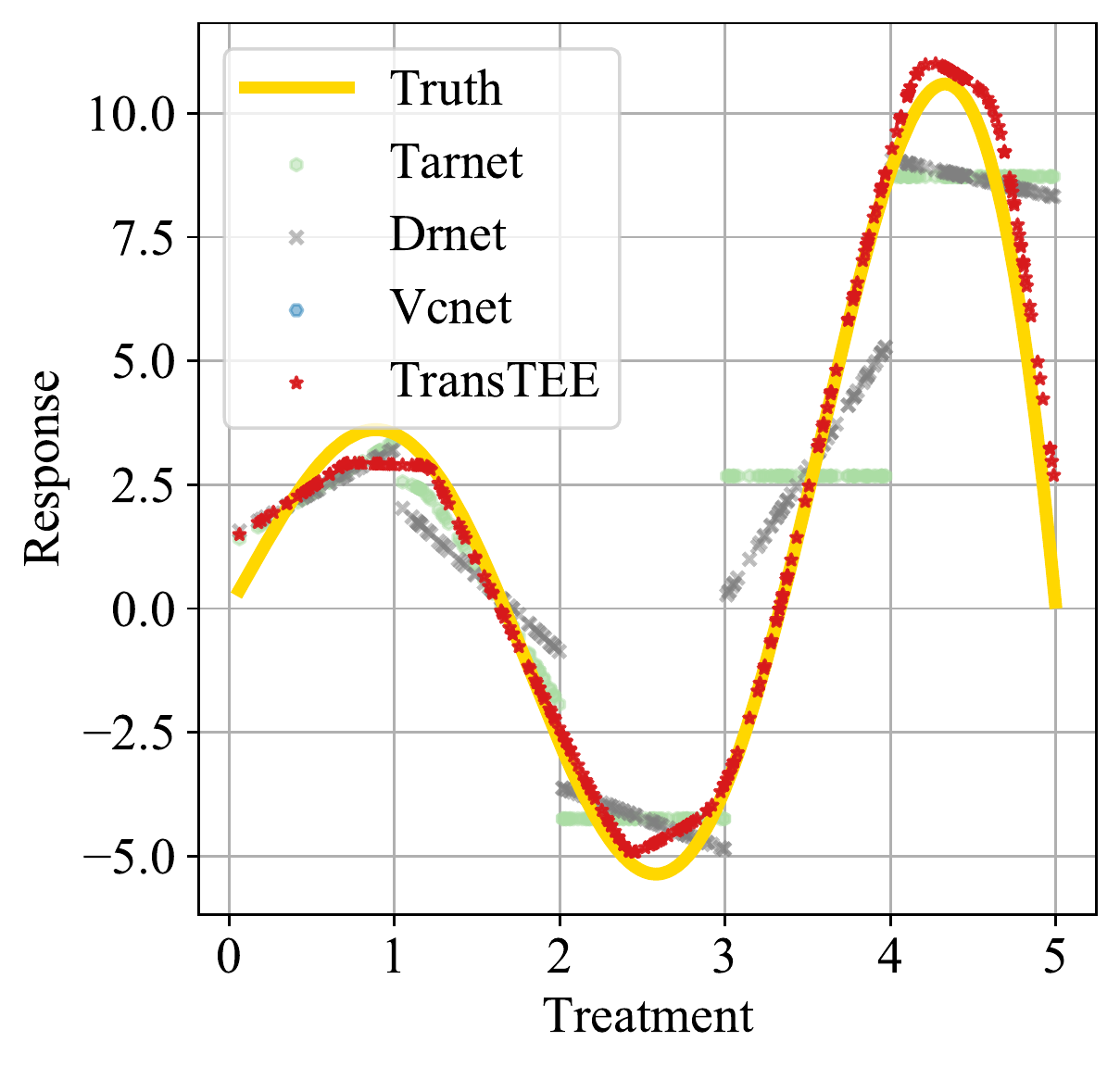}}
	\subfigure[$h=5,l=0.25$ during training and $h=5,l=0$ during testing (extrapolation).]
	{\includegraphics[width=0.24\textwidth]{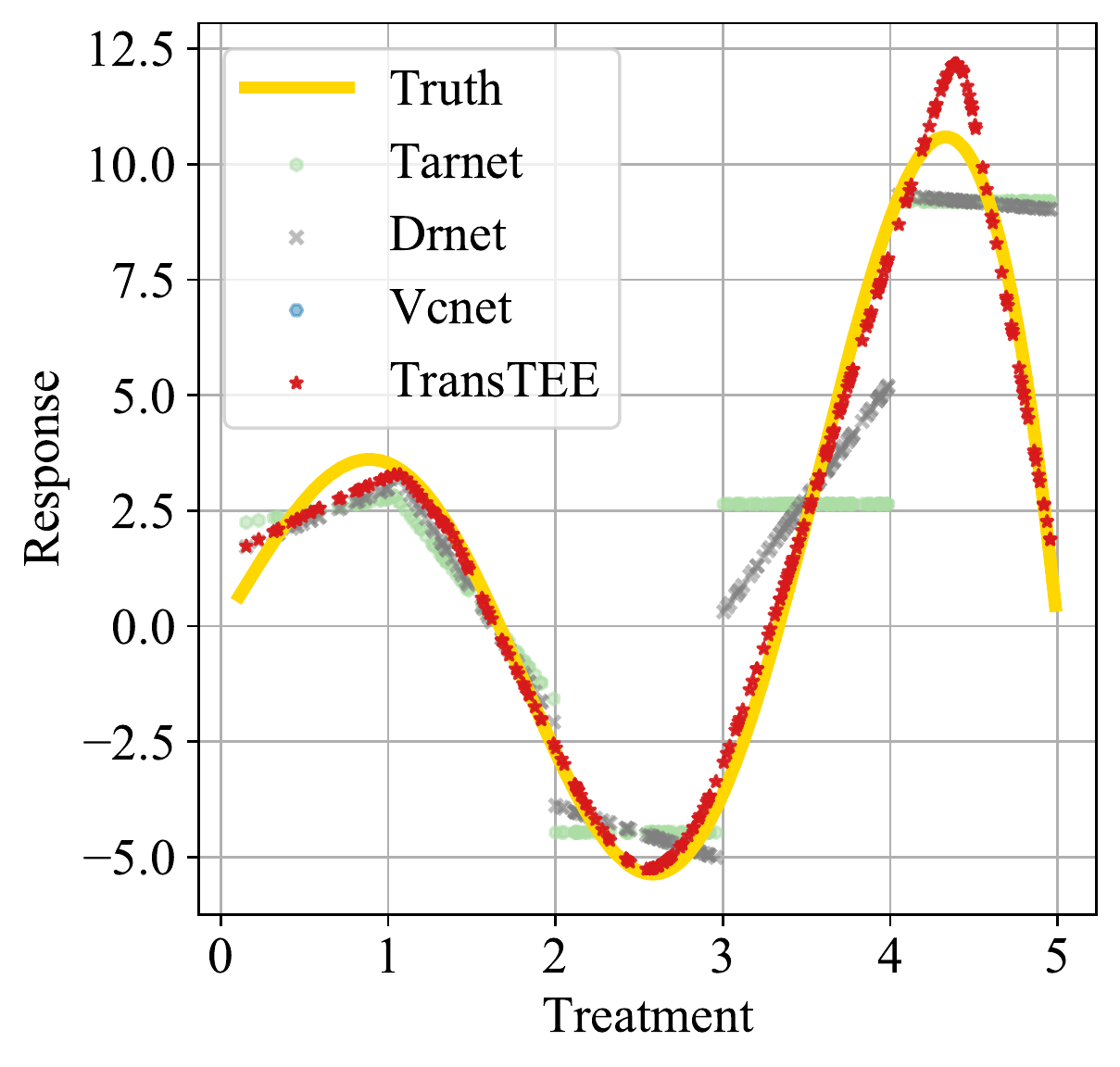}}
    \end{minipage}
    \caption{\textbf{Estimated ADRF} on test set from a typical run of TarNet~\citep{shalit2017tarnet}, DRNet~\citep{schwab2020drnet},  VCNet~\citep{nie2021vcnet} and ours on IHDP dataset. All of these methods are well optimized. (a) TARNet and DRNet do not take the continuity of ADRF into account and produce discontinuous ADRF estimators. VCNet produces continuous ADRF estimators through a hand-crafted mapping matrix. The proposed \abbr~embed treatments into continuous embeddings by neural network and attains superior results. (b,d) When training with $0.1\leq t\leq2.0$ and $0.25\leq t\leq5.0$. TARNet and DRNet cannot extrapolate to distributions with $0<t\leq2.0$ and $0\leq t\leq5.0$. (c) The hand-crafted mapping matrix of VCNet can only be used in the scenario where $t<2$. Otherwise, VCNet cannot converge and incur an infinite loss. At the same time, as $h$ is enhanced, TARNet and DRNet with the same number of branches perform worse. \abbr~needs not to know $h$ in advance and extrapolates well.}
    \label{fig:ADRF_ihdp}
\end{figure*}

\begin{figure*}[t]
\centering
\begin{minipage}[h]{.98\textwidth}        
\subfigure[Outcome regression error.]
{\includegraphics[width=0.33\textwidth]{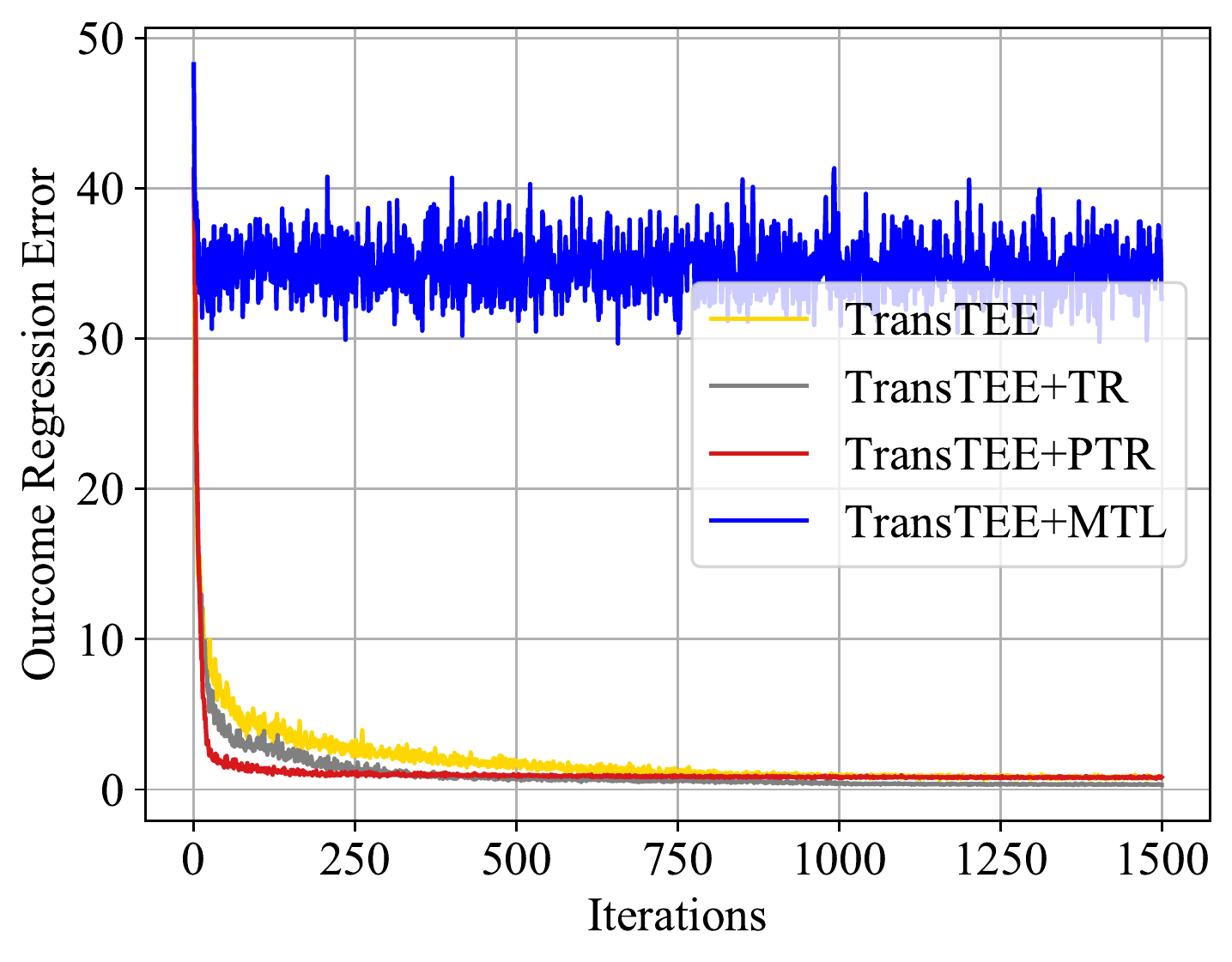}}
\subfigure[Treatment regression error.]
{\includegraphics[width=0.33\textwidth]{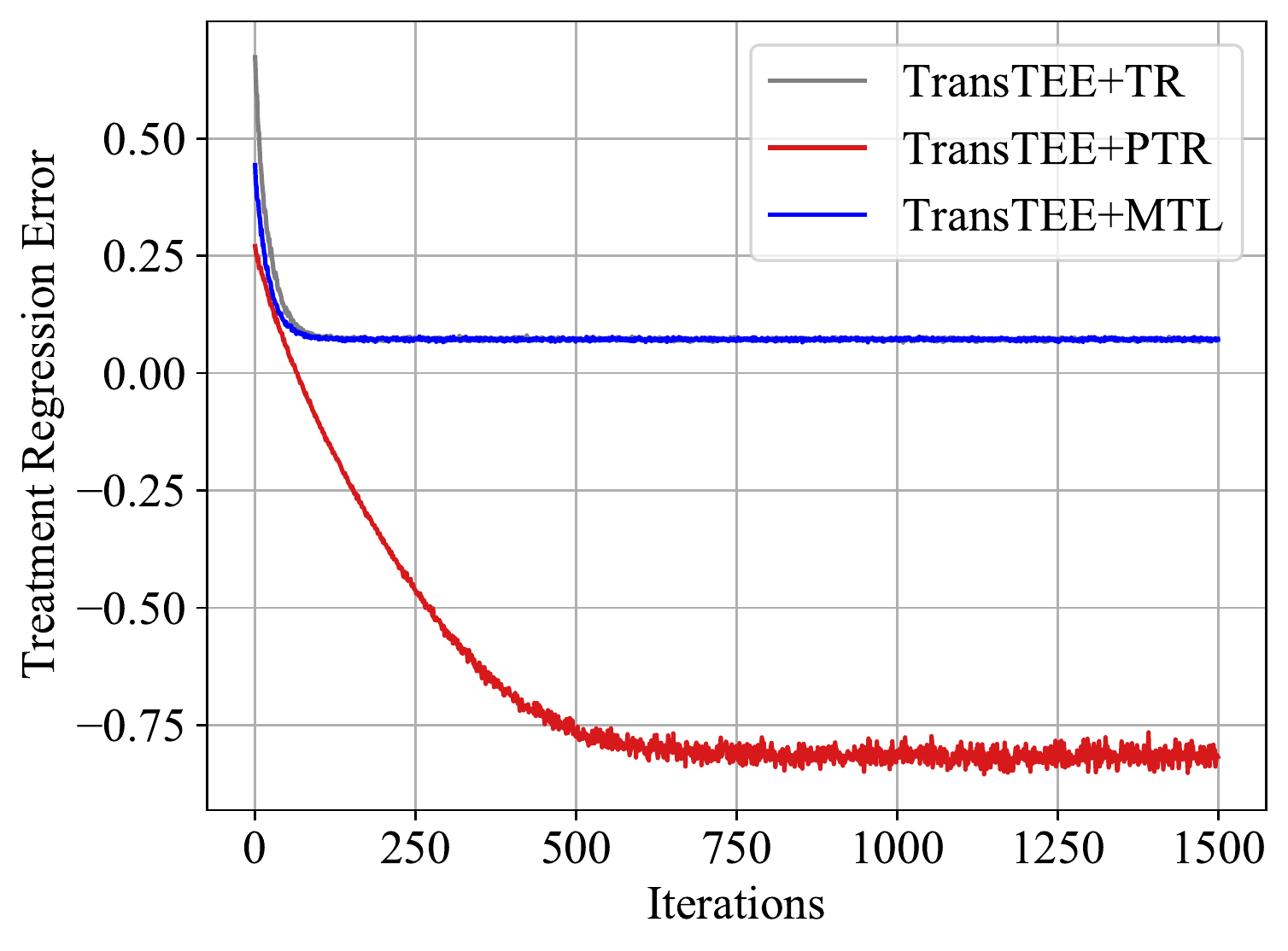}}
\subfigure[MSE in the test set.]
{\includegraphics[width=0.33\textwidth]{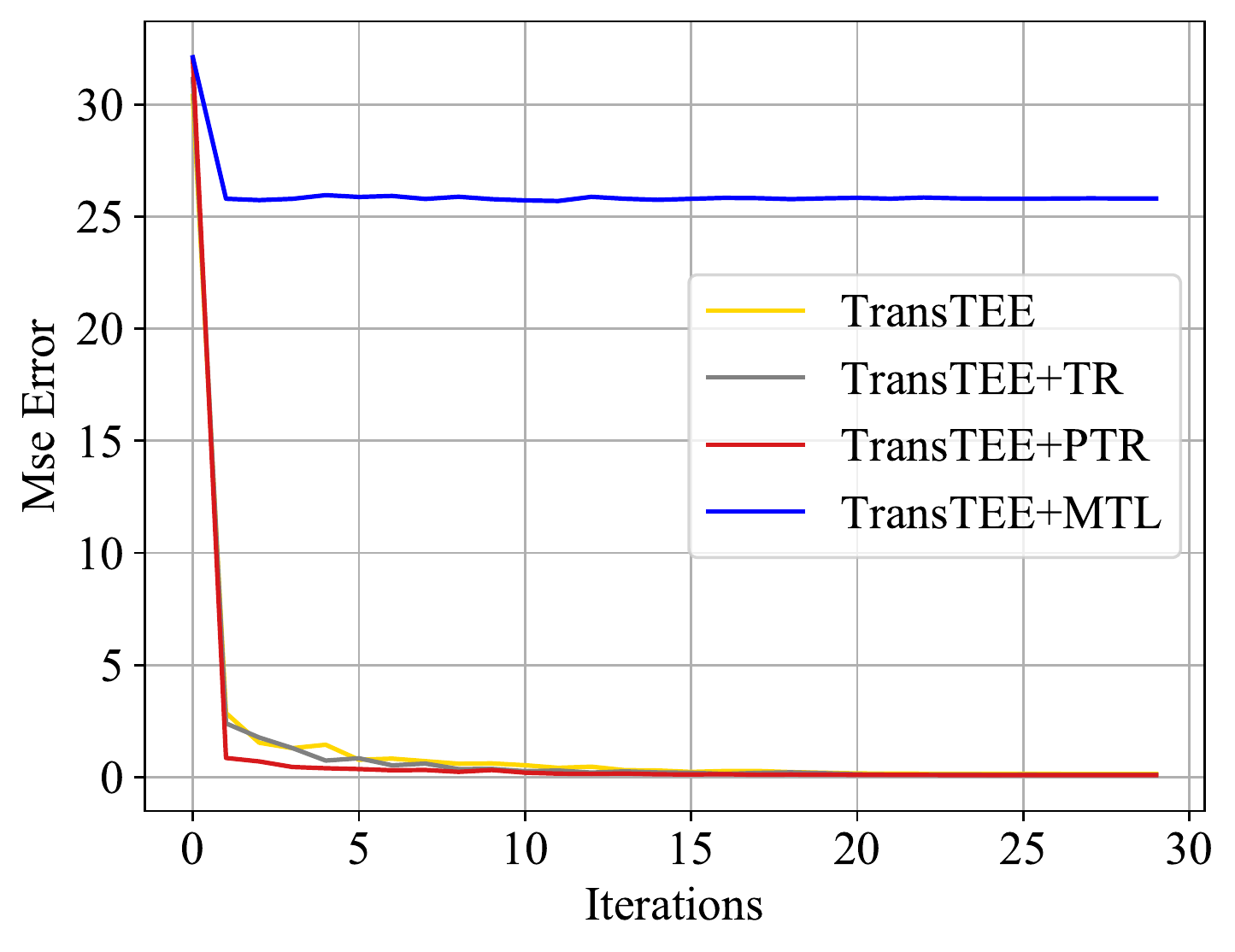}}
\end{minipage}
\caption{\textbf{Training dynamics of \abbr}~on IHDP dataset with various regularization terms, where the total training iteration is $1,500$ and (c) is evaluated on the test set per $50$ training iterations.}\label{fig:training_dynamic}
\end{figure*}

\begin{figure*}[t]
    \centering
    \begin{minipage}[t]{1\textwidth}        
    \subfigure[$h=1$ during training and testing.]
	{\includegraphics[width=0.24\textwidth]{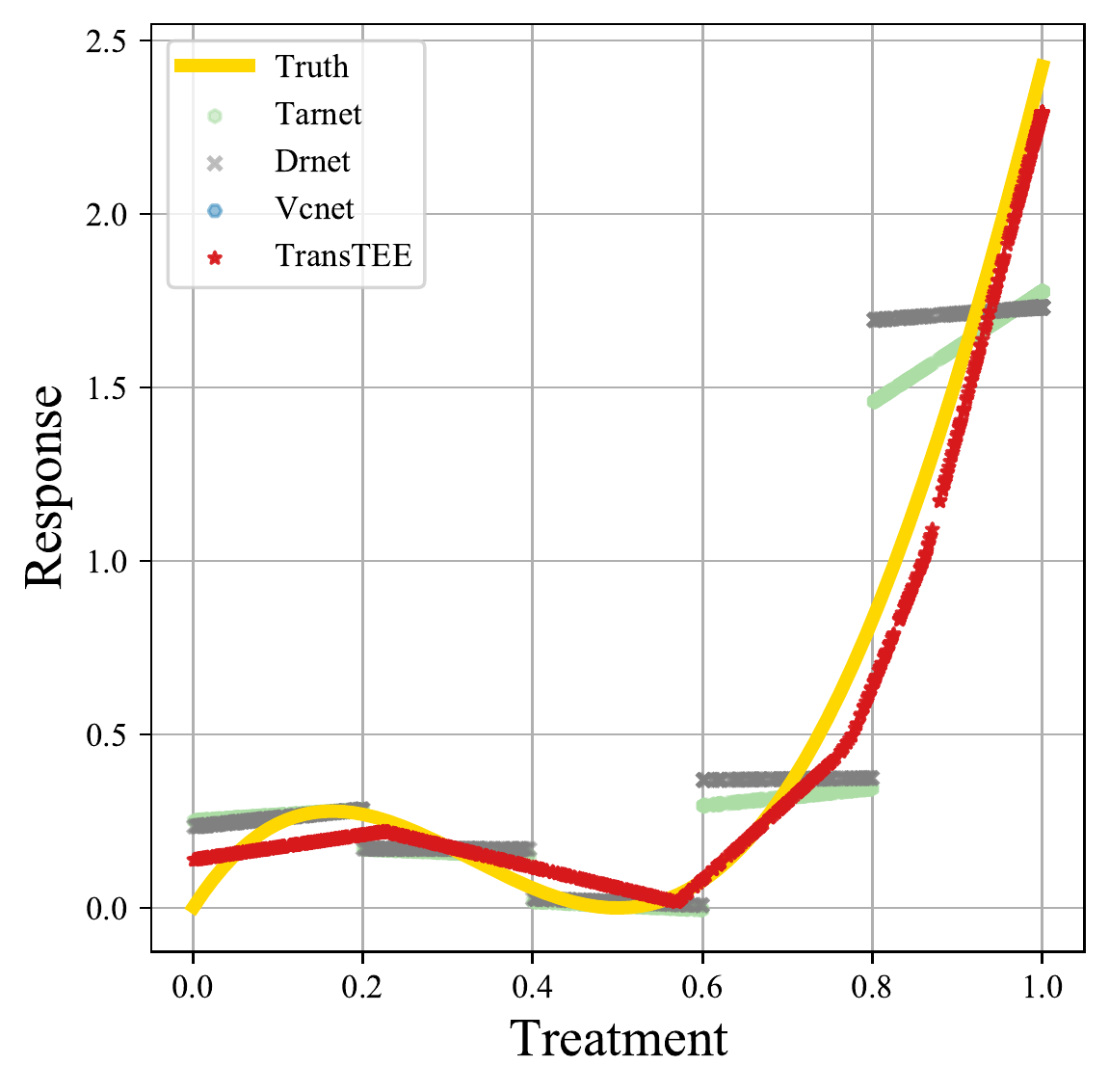}}
	\subfigure[$h=1.9,l=0$ during traning and $h=2,l=0$ during testing (extrapolation).]
	{\includegraphics[width=0.24\textwidth]{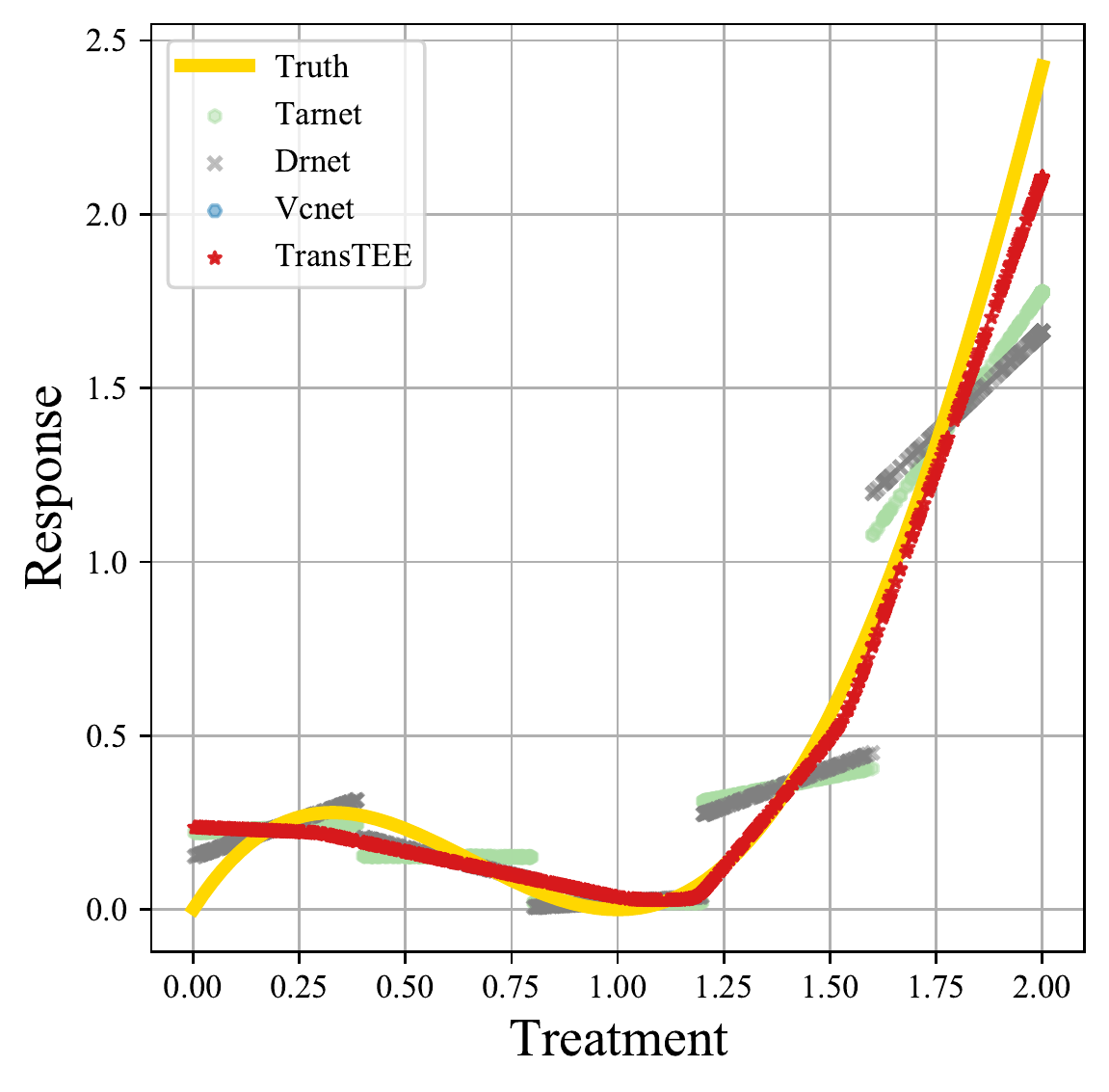}}
	\subfigure[$h=5$ during training and testing.]
	{\includegraphics[width=0.24\textwidth]{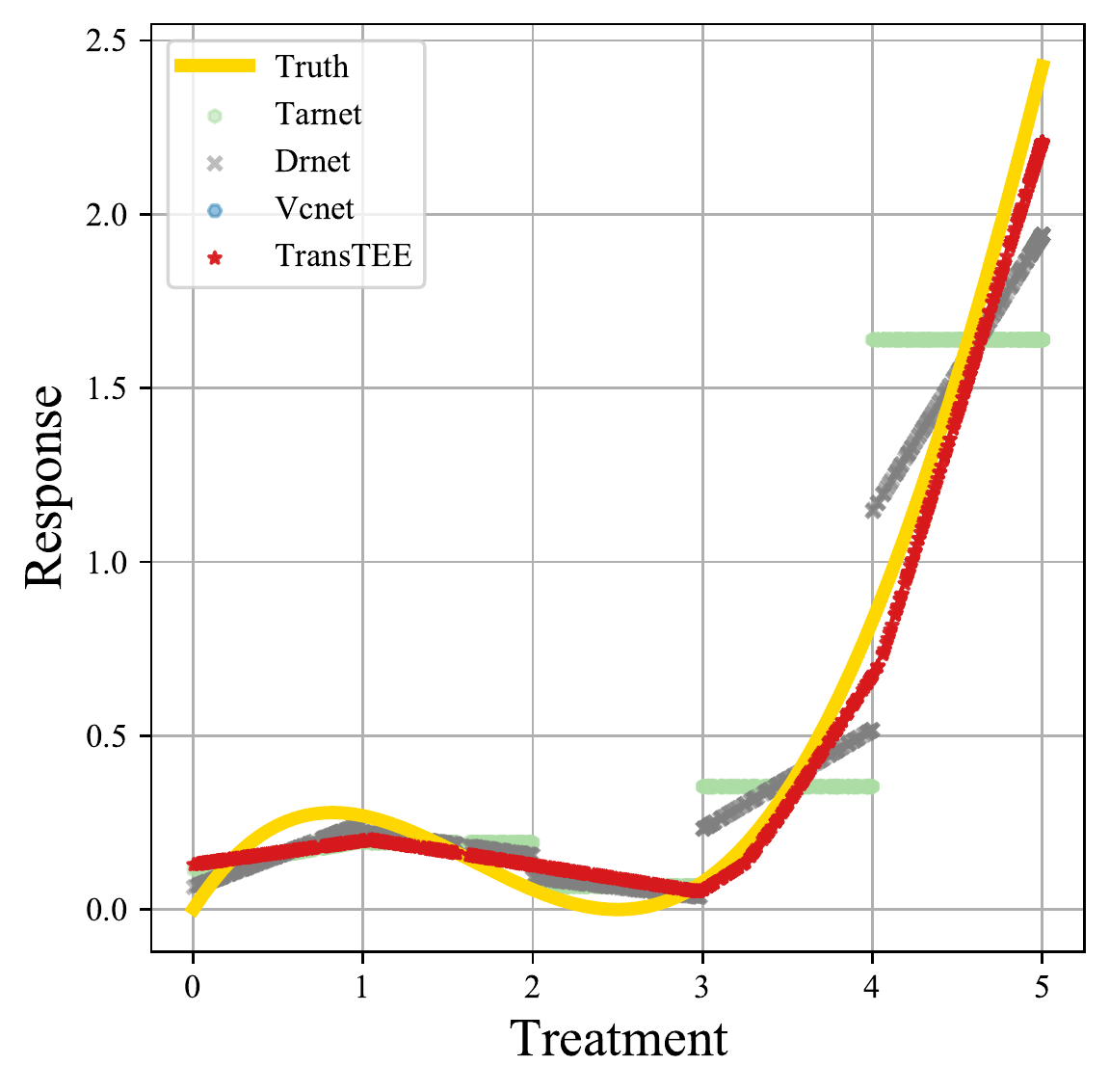}}
	\subfigure[$h=4,l=0$ during training and $h=5,l=0$ during testing (extrapolation).]
	{\includegraphics[width=0.24\textwidth]{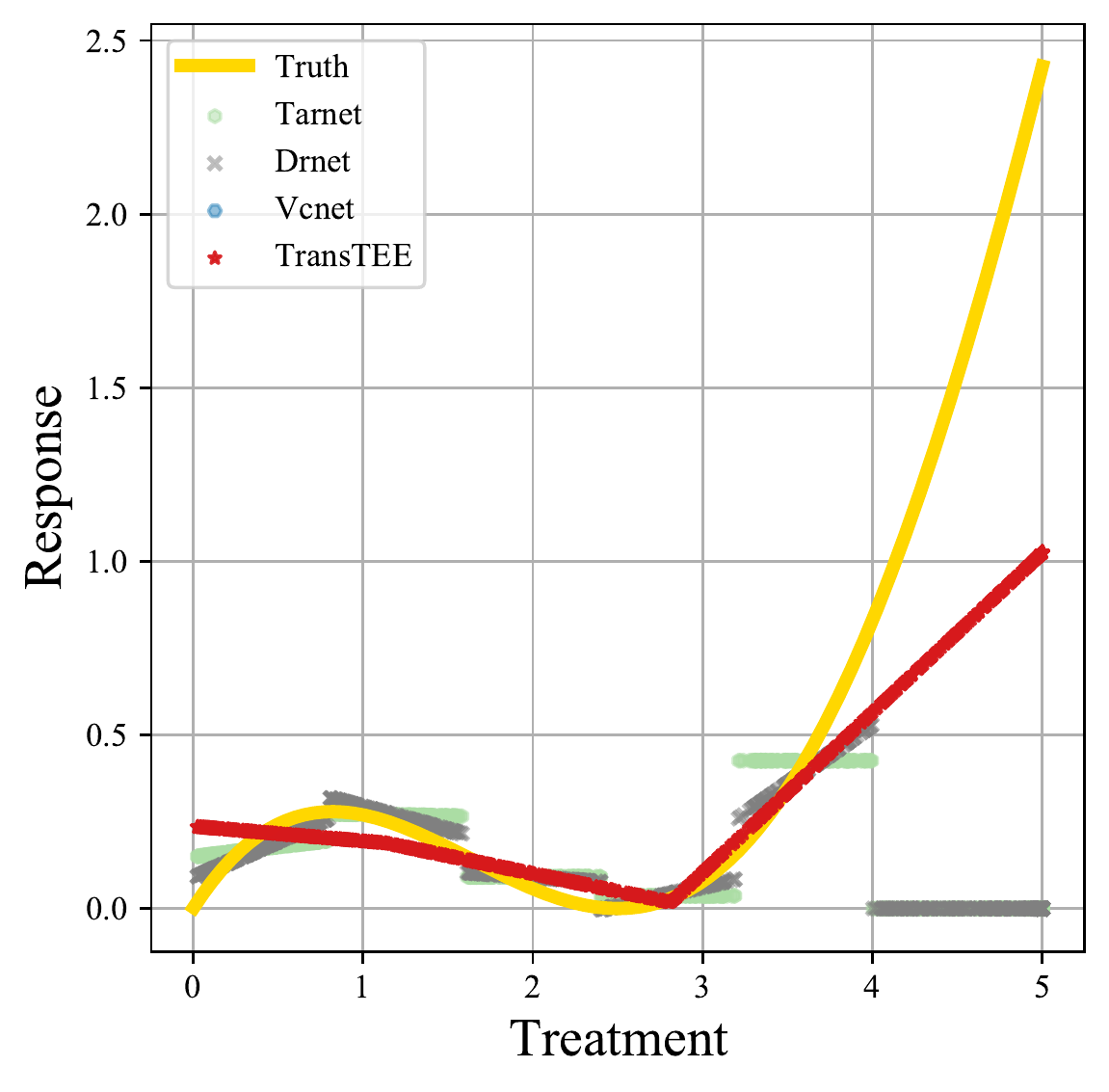}}
    \end{minipage}

    \caption{\textbf{Estimated ADRF} on the test set from a typical run of TarNet~\citep{shalit2017tarnet}, DRNet~\citep{schwab2020drnet},  VCNet~\citep{nie2021vcnet} and ours on News dataset.  All of these methods are well optimized. Suppose $t\in[l,h]$. (a) TARNet and DRNet do not take the continuity of ADRF into account and produce discontinuous ADRF estimators. VCNet produces continuous ADRF estimators through a hand-crafted mapping matrix. The proposed \abbr~embed treatments into continuous embeddings by neural network and attains superior results. (b,d) When training with $0\leq t\leq1.9$ and $0\leq t\leq4.0$. TARNet and DRNet cannot extrapolate to distributions with $0<t\leq2.0$ and $0\leq t\leq5.0$. (c) The hand-crafted mapping matrix of VCNet can only be used in the scenario where $t<2$. Otherwise, VCNet cannot converge and incur an infinite loss. At the same time, as $h$ is enhanced, TARNet and DRNet with the same number of branches perform worse. \abbr~needs not know $h$ in advance and extrapolates well.}
    \label{fig:ADRF_news}
\end{figure*}

\begin{table*}[t]
\caption{\textbf{Experimental results comparing neural network based methods on the News datasets.} Numbers reported are based on 20 repeats, and numbers after $\pm$ are the estimated standard deviation of the average value. For Extrapolation ($h=2$), models are trained with $t\in[0,1.9]$ and tested in $t\in[0,2]$. For For Extrapolation ($h=5$), models are trained with $t\in[0,4.5]$ and tested in $t\in[0,5]$}
\label{tab:continuous2}
\begin{center}
\begin{small}
\begin{sc}
\resizebox{\textwidth}{!}{ 
\begin{tabular}{ccccc}
\toprule
Methods & Vanilla & Vanilla ($h=5$) &Extrapolation ($h=2$) & Extrapolation ($h=5$) \\
\midrule
    TARNet& 0.082 $\pm$ 0.019& 0.956 $\pm$ 0.041& 0.716 $\pm$ 0.038 & 0.847 $\pm$ 0.053 \\
 DRNet& 0.083 $\pm$ 0.032&0.956 $\pm$ 0.041& 0.703 $\pm$ 0.038 & 0.834 $\pm$ 0.053\\
 VCNet& 0.013 $\pm$ 0.005& nan& nan &nan\\
 \rowcolor{Gray}
   \abbr& \textbf{0.010 $\pm$ 0.004}& 0.017 $\pm$ 0.008& 0.024 $\pm$ 0.017 &0.029 $\pm$ 0.019  \\
   \rowcolor{Gray}
   \abbr+TR& 0.011 $\pm$ 0.003& 0.016 $\pm$ 0.008& \textbf{0.019 $\pm$ 0.008} &\textbf{0.028 $\pm$ 0.002}  \\
   \rowcolor{Gray}
   \abbr+PTR& 0.011 $\pm$ 0.004& \textbf{0.014 $\pm$ 0.007}& 0.022 $\pm$ 0.008 &0.029 $\pm$ 0.016  \\
%   \abbr~v2& 0.017 $\pm$ 0.00743 & 0.0157 $\pm$ 0.00654 & 0.0284 $\pm$ 0.01543 & TODO \\

\bottomrule 
\end{tabular} }
\end{sc}
\end{small}
\end{center}
\vskip -0.1in
\end{table*}

\begin{figure*}[b]
    \centering
    \vskip -0.1in
    \begin{minipage}[t]{1.0\textwidth}        
    \subfigure[SW In-Sample.]
	{\includegraphics[width=0.5\textwidth]{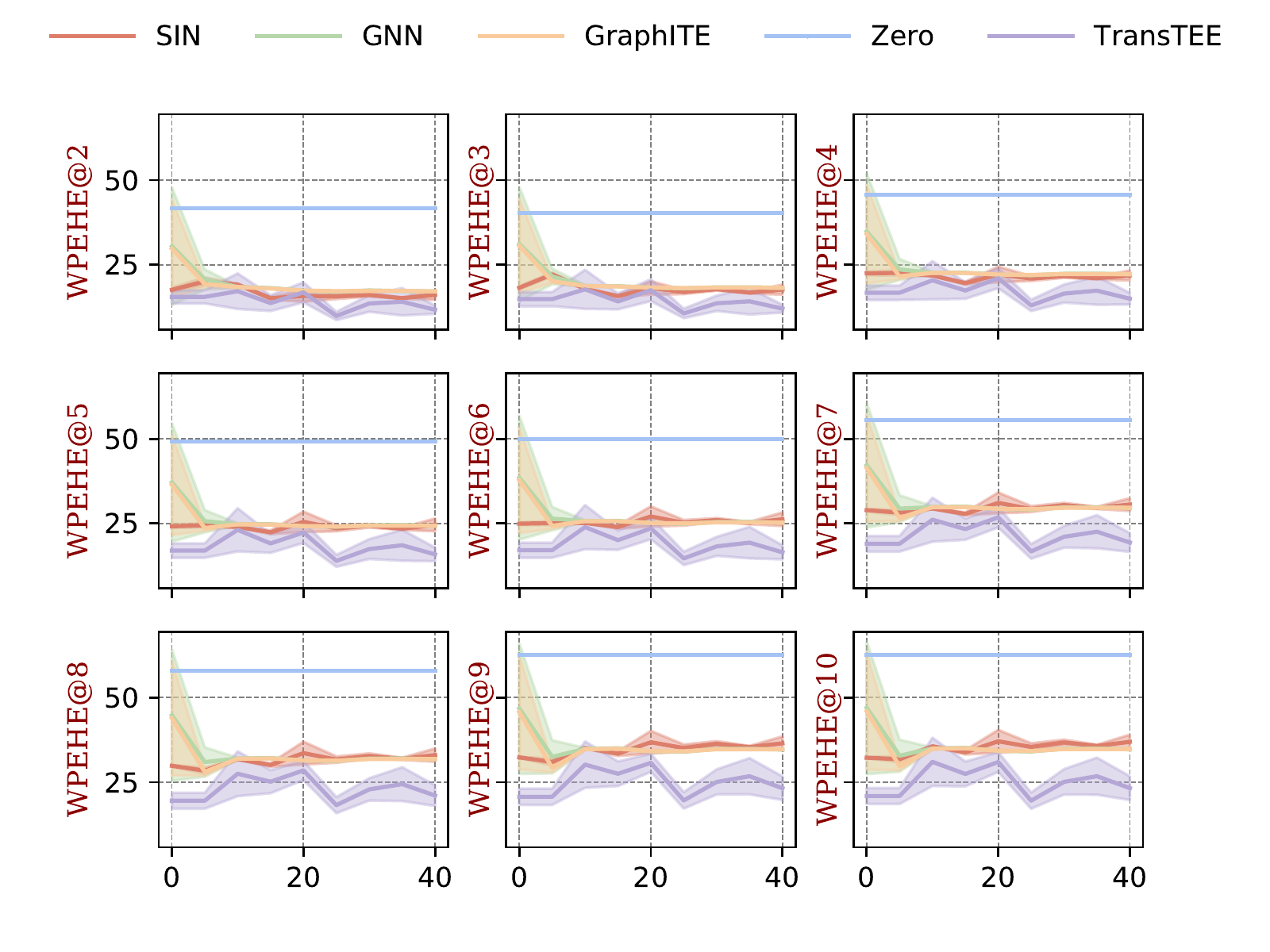}}
	\subfigure[SW Out-Sample]
	{\includegraphics[width=0.5\textwidth]{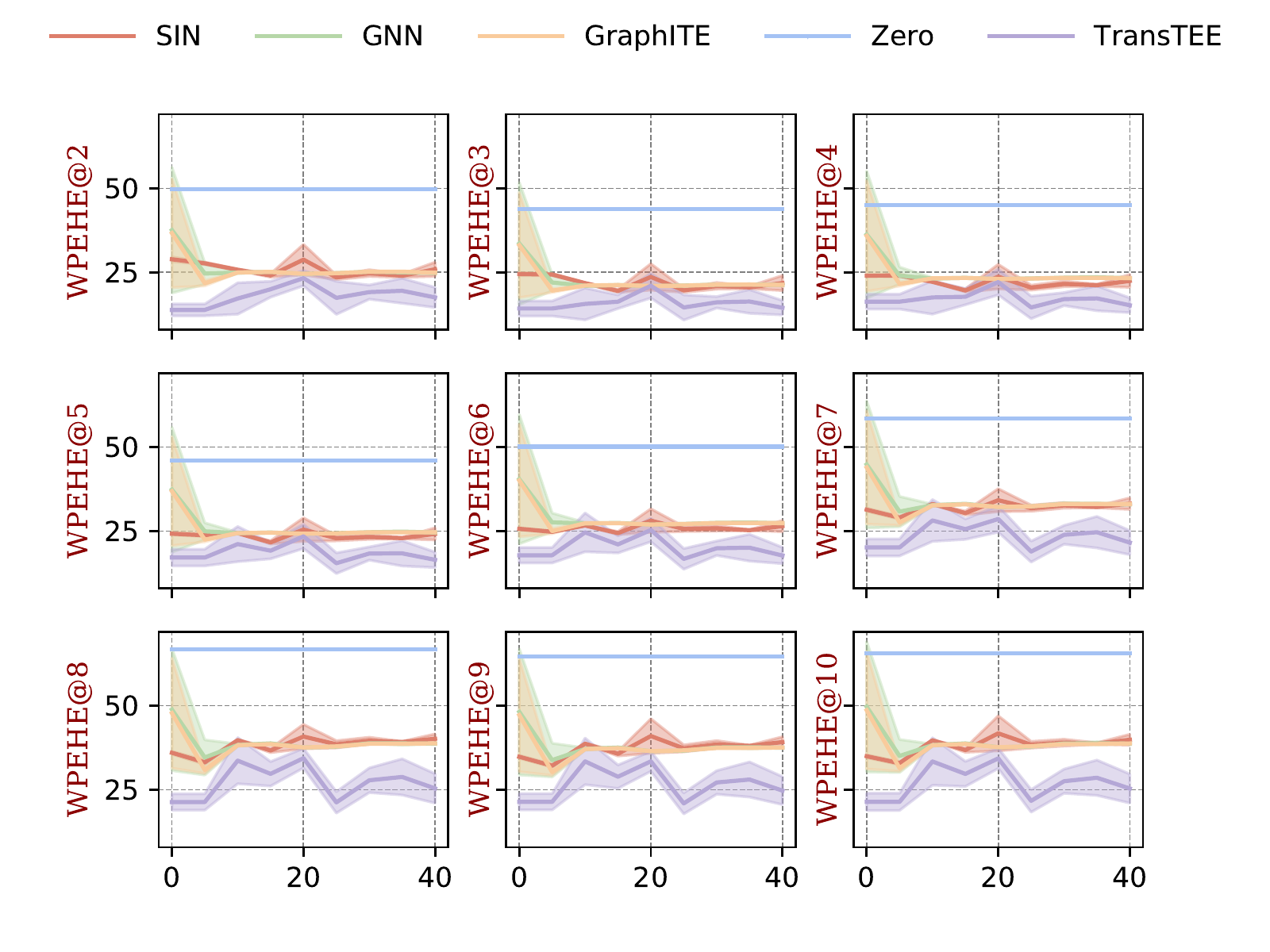}}
	\subfigure[TCGA In-Sample]
	{\includegraphics[width=0.5\textwidth]{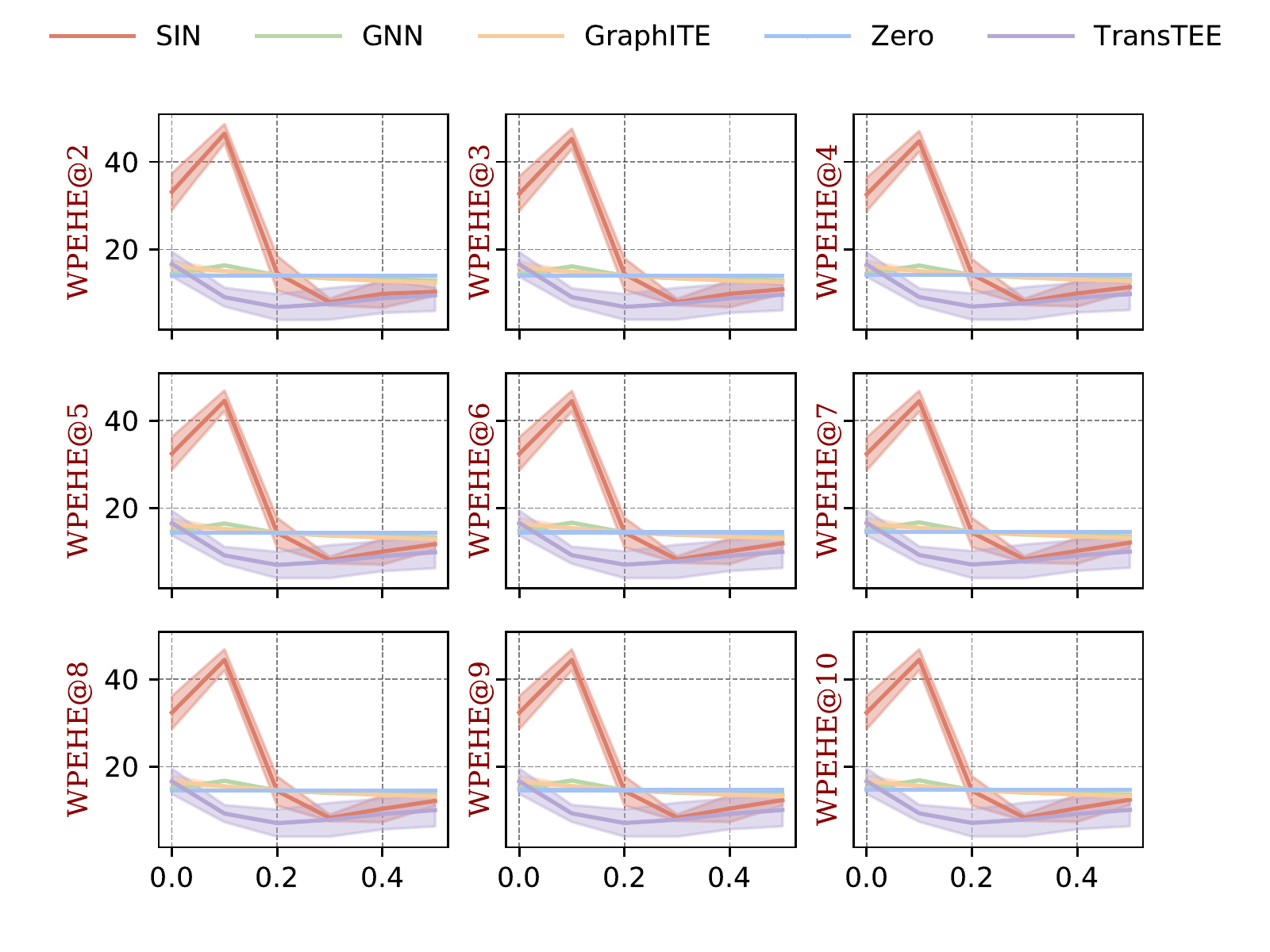}}
	\subfigure[TCGA Out-Sample]
	{\includegraphics[width=0.5\textwidth]{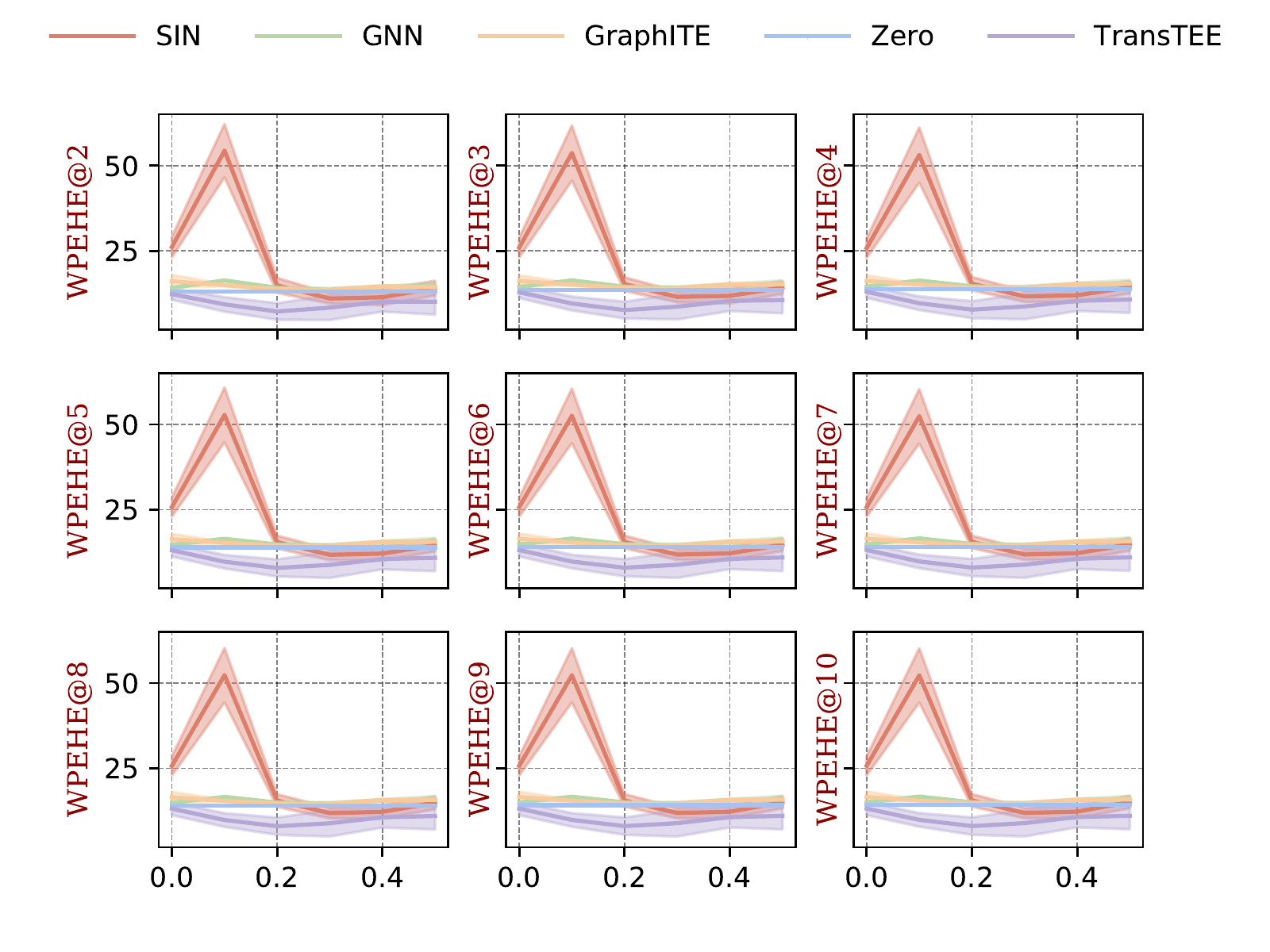}}
    \end{minipage}

    \caption{{\color{brown}{WPEHE@K}} over increasing bias strength $\kappa$ and varying $K\in\{2,...,10\}$ on the SW and the TCGA dataset.}
    \label{fig:kappa_structure}
\end{figure*}

\begin{figure*}[h]
    \centering
    \vskip -0.1in
    \begin{minipage}[t]{1.0\textwidth}        
    \subfigure[SW In-Sample.]
	{\includegraphics[width=0.5\textwidth]{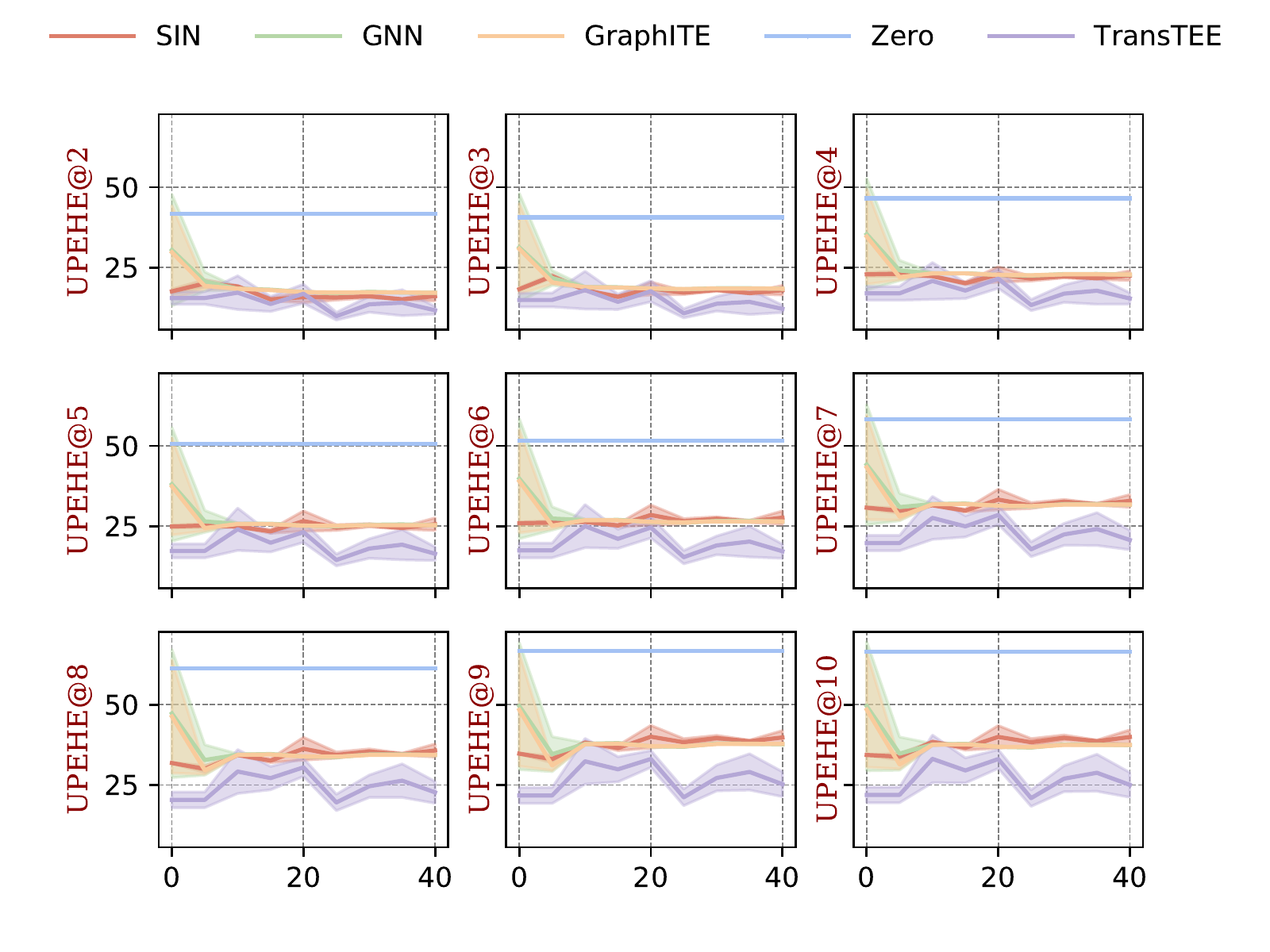}}
	\subfigure[SW Out-Sample]
	{\includegraphics[width=0.5\textwidth]{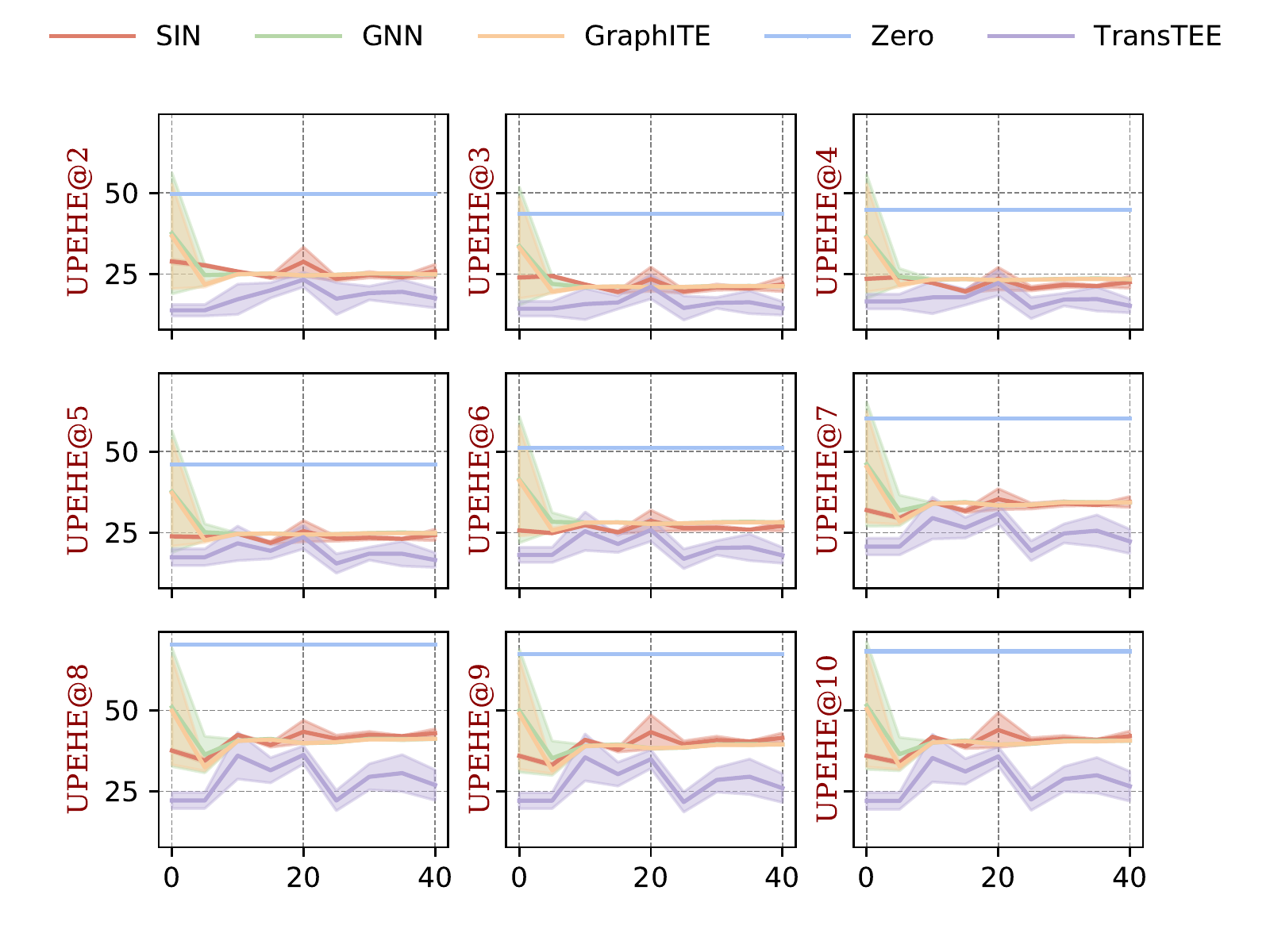}}
	\subfigure[TCGA In-Sample]
	{\includegraphics[width=0.5\textwidth]{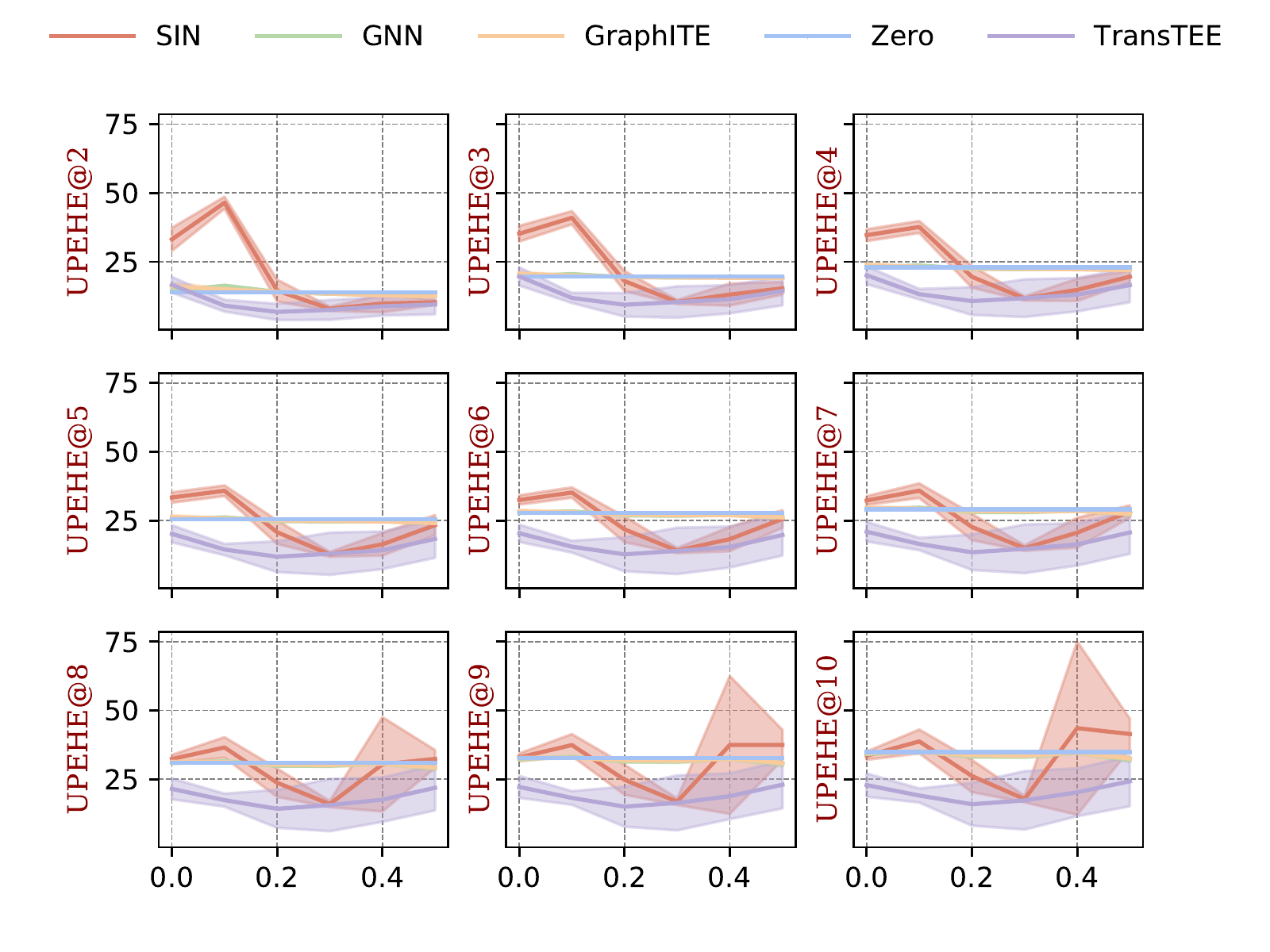}}
	\subfigure[TCGA Out-Sample]
	{\includegraphics[width=0.5\textwidth]{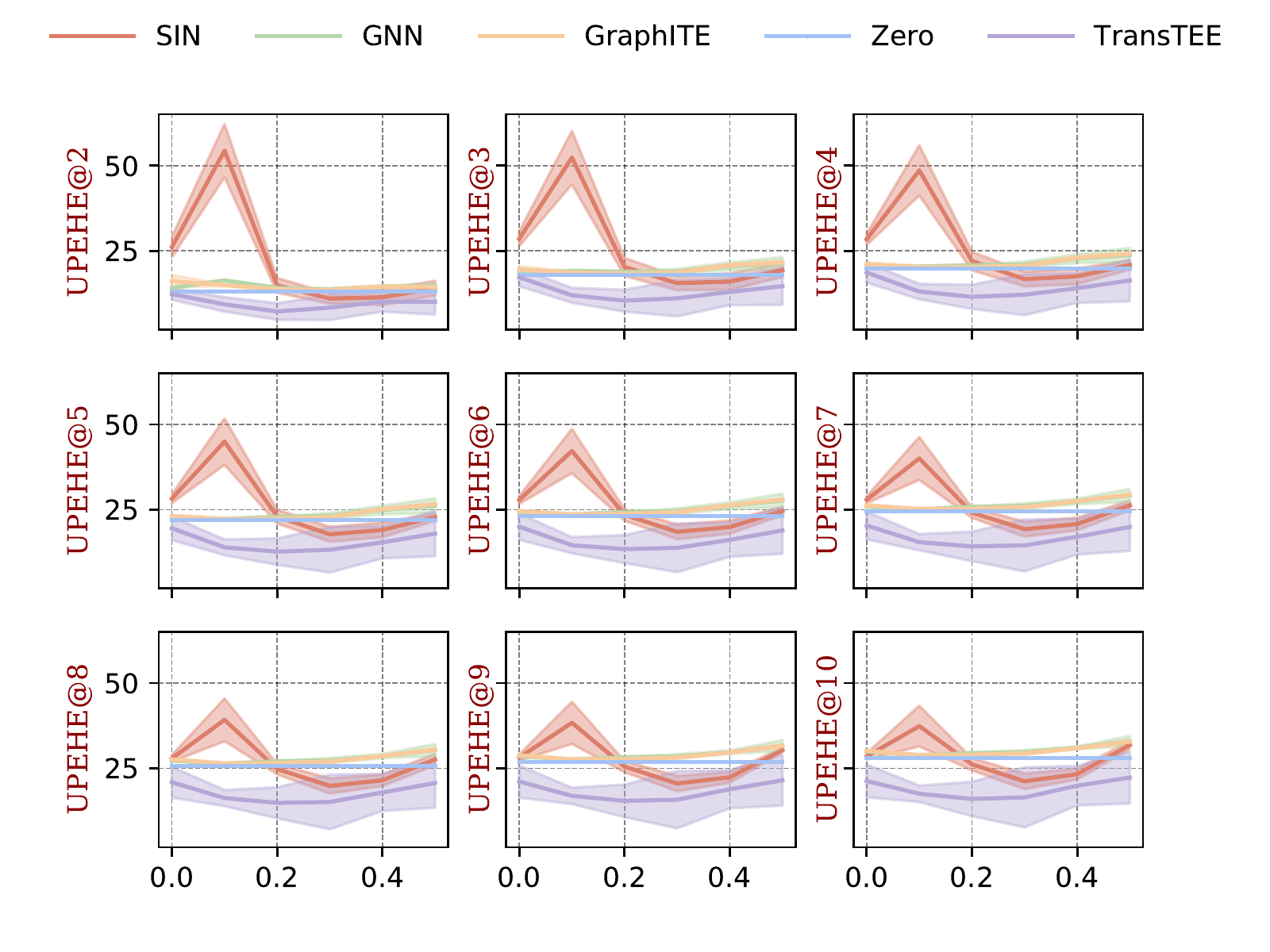}}
    \end{minipage}

    \caption{{\color{brown}{UPEHE@K}} over increasing bias strength $\kappa$ and varying $K\in\{2,...,10\}$ on the SW and the TCGA dataset.}
    \label{fig:kappa_structure_unweigt}
\end{figure*}

\textbf{Choice of the balancing weight for treatment regularization.} To understand the effect of propensity score modeling, we conduct an ablation study on the balancing weights of both TR and PTR. \figurename~\ref{fig:ablation_lambda} presents the results of the experiments on the IHDP dataset. The main observation is that both TR and PTR with a proper regularization strength consistently improve estimation compared to \abbr~without regularization. The best performers are achieved when $\lambda$ is $0.5$ for both two methods, which shows that the best balancing parameter ($0.5$ in our experiments.) for these two regularization terms should be searched carefully. Besides, training both the treatment predictor and the feature encoder simultaneously in a zero-sum game is difficult and sometimes unstable (shown in \figurename~\ref{fig:ablation_lambda} right)

\textbf{Robustness to noisy covariates.}  We manipulate $S_{dis,1},S_{dis,2}$ to generate datasets with different noisy covariates, \eg when the \textit{number of covariates that only influence the outcome} is $6$,  $S_{dis,1}=\{4\}$, and $S_{dis,2}=\{7,8,9,10,11,12,13,14,15,16,17,18,19,20,21,22,23,24,25\}$, when the \textit{number of covariates that influence the outcome} is $24$,  $S_{dis,1}=\{4,7,8,9,10,11,12,13,14,15,16,17,18,19,20,21,22,23,24,\}$, and $S_{dis,2}=\{25\}$. Figure \figurename~\ref{fig:useful_y} shows that, as the number of covariates that only influence the outcome increases, both TARNet and DRNet become better estimators, however, VCNet performs worse and even inferior to TARNet and DRNet when the number is large than $16$. In contrast, the estimation error incurred by the proposed \abbr~is always low and superior to baselines by a large margin.

\textbf{Comparison of MLE or adversarial propensity score modeling on the propensity score.} Seeing results in Table \ref{tab:continuous1}, additionally combine \abbr~with maximum likelihood training of $\pi(t|\x)$ does provide some performance gains. However, an adversarially trained $\pi$-model can be significantly better, especially for extrapolation settings. The results well manifest the effectiveness of TR and PTR in reducing selection bias and improving estimation performance. In fact, approaches like TMLE are not robust if the initial estimator is poor \cite{shi2019dragonnet}.

\textbf{Training dynamics comparison of different regularization terms.} Here we compare four regularization terms, which are \abbr~with no regularization, \abbr+TR, \abbr+PTR, and \abbr+MTL. \abbr+MTL is a simple \textbf{M}ulti-\textbf{T}ask \textbf{L}earning strategy, which uses $\mathcal{L}_{\theta}(\x, y, t)+\mathcal{L}_{\phi}^{TR}(\x, t)$ during training without an adversarial game. As shown in \figurename~\ref{fig:training_dynamic}, without adversarial training, \abbr+MTL quickly attains low treatment estimation error but further oscillates and converge with a high error, and both the outcome regression error and MSE in the test set remain high. In contrast, TR and PTR make \abbr~converge faster and attain lower test MSE. Overall, PTR consistently works the best and its low treatment regression error shows that $\pi_\phi(t|\x)$ estimates an accurate propensity score.

\subsection{Showcase of sentences and counterfactual counterparts with the maximal/minimal ATEs.}\label{sec:samples_causalm}
Table~\ref{samples_gender} showcases the top-$10$ samples with the maximal/ minimal ATEs. Interestingly, we can see most sentences with a large ATE have similar patterns, that is ``\textit{$<$ clause $>$, but/and $<$ Person $>$ made me feel $<$ Adj $>$}''. Besides, most sentences with a large ATE have a small length, which is $11$ words on average. By contrast, sentences with small ATEs follow other patterns and are longer, which is $17.6$ on average. Consider the effect of \textit{Race}, Table~\ref{samples_race} showcases the top-$10$ samples. Similarly, there are also some dominant patterns that have pretty high or low ATEs and the average length of sentences with high ATEs is smaller than sentences with low ATEs ($12\;vs\; 14.7$). Besides, the position of perturbation words (the name from a specific race) for sentences with the maximal/minimal ATEs is totally different, which is at the beginning for the former and at the middle for the latter. Namely, \abbr~helps us mitigate spurious correlations that exist in model prediction, \eg length of sentences, the position of perturbation words, certain sentence patterns and is useful in mitigating undesirable bias ingrained in the data. Besides, a well-optimized \abbr~is able to estimate the effect of every sentence and is of great benefit for model interpretation and analysis especially under high inference latency. 

\begin{table*}
\caption{\textbf{Error of CATE estimation for all methods, measured by {\color{brown}{WPEHE@2-10}}.} Results are averaged over 5 trials, ± denotes std error. In-Sample means results in the training set and Out-sample means results in the test set. (The baseline results are reproduced using the official code of \citep{kaddour2021causal} in a consistent experimental environment, which can be slightly different than the results reported in \citep{kaddour2021causal}) }\label{tab:results_structure}
\resizebox{\textwidth}{!}{
\begin{tabular}{@{}ccccccccc@{}}
\toprule
\multirow{2}{*}{\textbf{Method}} & \multicolumn{2}{c}{\textbf{SW}}                        & \multicolumn{2}{c}{\textbf{TCGA (Bias=0.1)}}         & \multicolumn{2}{c}{\textbf{TCGA (Bias=0.3)}}         & \multicolumn{2}{c}{\textbf{TCGA (Bias=0.5)}}           \\ \cmidrule(l){2-9} 
                        & In-sample             & Out-sample            & In-sample            & Out-sample           & In-sample            & Out-sample           & In-sample             & Out-sample            \\ \cmidrule(r){1-9}
\multicolumn{9}{c}{\color{brown}{WPEHE@2}}                                                                                                                                                                                         \\
Zero                    & 41.72 ± 0.00          & 49.69 ± 0.00          & 13.93 ± 0.00         & 13.13 ± 0.00         & 13.93 ± 0.00         & 13.13 ± 0.00         & 13.93 ± 0.00          & 13.61 ± 0.00          \\
GNN                     & 17.38 ± 0.01          & 24.53 ± 0.01          & 10.90 ± 7.71         & 10.91 ± 7.71         & 13.58 ± 0.18         & 13.22 ± 0.18         & 12.86 ± 0.38          & 14.62 ± 0.91          \\
GraphITE                & 17.37 ± 0.01          & 24.56 ± 0.02          & 15.04 ± 0.20         & 14.96 ± 0.30         & 13.49 ± 0.23         & 13.70 ± 0.52         & 12.41 ± 0.02          & 14.38 ± 0.30          \\
SIN                     & 15.79 ± 1.72          & 28.78 ± 4.54          & 46.47 ± 2.19         & 54.41 ± 7.81         & 7.93 ± 0.79          & 11.04 ± 1.52         & 10.31 ± 0.93          & 14.09 ± 2.14          \\
\rowcolor{Gray}
TransTEE                & \textbf{14.74 ± 0.09} & \textbf{21.78 ± 1.07} & \textbf{9.07 ± 2.15} & \textbf{9.33 ± 2.13} & \textbf{7.54 ± 3.60} & \textbf{8.37 ± 3.64} & \textbf{9.52 ± 3.59}  & \textbf{10.10 ± 3.79} \\\hline
\multicolumn{9}{c}{\color{brown}{WPEHE@3}}                                                                                                                                                                                         \\
Zero                    & 40.75 ± 0.00          & 43.76 ± 0.00          & 13.93 ± 0.00         & 13.61 ± 0.00         & 13.93 ± 0.00         & 13.61 ± 0.00         & 13.61 ± 0.00          & 14.14 ± 0.00          \\
GNN                     & 18.26 ± 0.00          & 20.91 ± 0.01          & 10.75 ± 7.60         & 10.91 ± 7.72         & 13.63 ± 0.18         & 13.58 ± 0.19         & 12.92 ± 0.33          & 15.29 ± 1.04          \\
GraphITE                & 18.27 ± 0.01          & 20.95 ± 0.02          & 14.88 ± 0.19         & 15.12 ± 0.29         & 13.49 ± 0.22         & 14.19 ± 0.43         & 12.56 ± 0.01          & 15.18 ± 0.31          \\
SIN                     & 18.15 ± 1.97          & 23.62 ± 3.93          & 45.29 ± 2.33         & 53.72 ± 8.09         & 7.94 ± 0.75          & 11.53 ± 1.59         & 10.89 ± 1.07          & 14.27 ± 1.92          \\
\rowcolor{Gray}
TransTEE                & \textbf{15.30 ± 1.12} & \textbf{18.73 ± 2.09} & \textbf{9.07 ± 2.02} & \textbf{9.58 ± 2.04} & \textbf{7.58 ± 3.62} & \textbf{8.65 ± 3.75} & \textbf{9.64 ± 3.56}  & \textbf{10.59 ± 3.88} \\\hline
\multicolumn{9}{c}{\color{brown}{WPEHE@4}}                                                                                                                                                                                         \\
Zero                    & 45.74 ± 0.00          & 44.95 ± 0.00          & 14.14 ± 0.00         & 13.75 ± 0.00         & 14.14 ± 0.00         & 13.75 ± 0.00         & 13.75 ± 0.00          & 14.31 ± 0.00          \\
GNN                     & 22.09 ± 0.01          & 23.01 ± 0.01          & 10.87 ± 7.69         & 10.88 ± 7.69         & 13.87 ± 0.18         & 13.71 ± 0.19         & 13.13 ± 0.34          & 15.47 ± 1.05          \\
GraphITE                & 22.12 ± 0.00          & 23.03 ± 0.02          & 15.05 ± 0.18         & 15.14 ± 0.28         & 13.64 ± 0.20         & 14.30 ± 0.35         & 12.77 ± 0.02          & 15.38 ± 0.30          \\
SIN                     & 22.14 ± 2.30          & 23.70 ± 3.67          & 44.72 ± 2.35         & 53.12 ± 8.09         & 7.99 ± 0.73          & 11.66 ± 1.59         & 11.38 ± 1.04          & 14.37 ± 1.83          \\
\rowcolor{Gray}
TransTEE                & \textbf{18.99 ± 0.83} & \textbf{19.65 ± 1.97} & \textbf{9.09 ± 1.97} & \textbf{9.66 ± 2.01} & \textbf{7.67 ± 3.70} & \textbf{8.71 ± 3.78} & \textbf{9.78 ± 3.63}  & \textbf{10.74 ± 3.91} \\\hline
\multicolumn{9}{c}{\color{brown}{WPEHE@5}}                                                                                                                                                                                         \\
Zero                    & 49.19 ± 0.00          & 45.96 ± 0.00          & 14.31 ± 0.00         & 13.95 ± 0.00         & 14.31 ± 0.00         & 13.95 ± 0.00         & 13.95 ± 0.00          & 14.47 ± 0.00          \\
GNN                     & 24.18 ± 0.01          & 24.20 ± 0.01          & 10.99 ± 7.77         & 10.97 ± 7.76         & 13.98 ± 0.17         & 13.92 ± 0.18         & 13.31 ± 0.37          & 15.67 ± 1.05          \\
GraphITE                & 24.22 ± 0.01          & 24.22 ± 0.03          & 15.24 ± 0.19         & 15.29 ± 0.28         & 13.68 ± 0.17         & 14.37 ± 0.37         & 12.95 ± 0.03          & 15.59 ± 0.30          \\
SIN                     & 25.48 ± 3.02          & 25.44 ± 3.50          & 44.55 ± 2.35         & 52.78 ± 8.04         & 8.10 ± 0.75          & 11.76 ± 1.59         & 11.75 ± 1.22          & 14.59 ± 1.84          \\
\rowcolor{Gray}
TransTEE                & \textbf{20.16 ± 0.42} & \textbf{21.08 ± 1.78} & \textbf{9.17 ± 1.96} & \textbf{9.72 ± 2.00} & \textbf{7.76 ± 3.75} & \textbf{8.80 ± 3.82} & \textbf{9.91 ± 3.66}  & \textbf{10.89 ± 3.94} \\\hline
\multicolumn{9}{c}{\color{brown}{WPEHE@6}}                                                                                                                                                                                         \\
Zero                    & 49.95 ± 0.00          & 50.10 ± 0.00          & 14.47 ± 0.00         & 14.04 ± 0.00         & 14.47 ± 0.00         & 14.04 ± 0.00         & 14.04 ± 0.00          & 14.53 ± 0.00          \\
GNN                     & 25.13 ± 0.00          & 26.93 ± 0.01          & 11.11 ± 7.86         & 11.02 ± 7.79         & 14.07 ± 0.22         & 14.11 ± 0.18         & 13.45 ± 0.38          & 15.76 ± 1.04          \\
GraphITE                & 25.17 ± 0.02          & 26.94 ± 0.02          & 15.40 ± 0.19         & 15.37 ± 0.28         & 13.74 ± 0.12         & 14.58 ± 0.38         & 13.09 ± 0.04          & 15.68 ± 0.29          \\
SIN                     & 27.07 ± 2.98          & 28.11 ± 3.51          & 44.48 ± 2.35         & 52.54 ± 7.99         & 8.22 ± 0.75          & 11.82 ± 1.58         & 11.97 ± 1.19          & 14.74 ± 1.86          \\
\rowcolor{Gray}
TransTEE                & \textbf{21.32 ± 0.79} & \textbf{22.99 ± 1.43} & \textbf{9.23 ± 1.95} & \textbf{9.77 ± 1.99} & \textbf{7.80 ± 3.83} & \textbf{8.84 ± 3.89} & \textbf{10.01 ± 3.70} & \textbf{10.96 ± 3.95} \\\hline
\multicolumn{9}{c}{\color{brown}{WPEHE@7}}                                                                                                                                                                                         \\
Zero                    & 55.40 ± 0.00          & 58.42 ± 0.00          & 14.53 ± 0.00         & 14.09 ± 0.00         & 14.53 ± 0.00         & 14.09 ± 0.00         & 14.53 ± 0.00          & 14.09 ± 0.00          \\
GNN                     & 29.30 ± 0.03          & 32.15 ± 0.03          & 11.16 ± 7.89         & 11.06 ± 7.82         & 14.12 ± 0.21         & 14.14 ± 0.18         & 13.51 ± 0.38          & 15.81 ± 1.03          \\
GraphITE                & 29.34 ± 0.01          & 32.16 ± 0.01          & 15.47 ± 0.19         & 15.42 ± 0.28         & 13.97 ± 0.08         & 14.69 ± 0.40         & 13.16 ± 0.04          & 15.74 ± 0.29          \\
SIN                     & 31.07 ± 3.07          & 34.17 ± 3.41          & 44.45 ± 2.37         & 52.40 ± 7.98         & 8.28 ± 0.74          & 11.85 ± 1.58         & 12.11 ± 1.18          & 14.83 ± 1.87          \\
\rowcolor{Gray}
TransTEE                & \textbf{24.71 ± 0.41} & \textbf{25.84 ± 0.73} & \textbf{9.27 ± 1.94} & \textbf{9.81 ± 1.99} & \textbf{7.82 ± 3.84} & \textbf{8.89 ± 3.89} & \textbf{10.06 ± 3.71} & \textbf{11.01 ± 3.95} \\\hline
\multicolumn{9}{c}{\color{brown}{WPEHE@8}}                                                                                                                                                                                         \\
Zero                    & 57.99 ± 0.00          & 66.78 ± 0.00          & 14.61 ± 0.00         & 14.14 ± 0.00         & 14.60 ± 0.00         & 14.12 ± 0.00         & 14.61 ± 0.00          & 14.14 ± 0.00          \\
GNN                     & 31.41 ± 0.03          & 37.57 ± 0.05          & 11.22 ± 7.93         & 11.09 ± 7.85         & 14.19 ± 0.25         & 14.20 ± 0.18         & 13.58 ± 0.38          & 15.87 ± 1.02          \\
GraphITE                & 31.45 ± 0.01          & 37.58 ± 0.00          & 15.55 ± 0.19         & 15.47 ± 0.28         & 14.30 ± 0.04         & 14.85 ± 0.43         & 13.23 ± 0.04          & 15.78 ± 0.28          \\
SIN                     & 33.58 ± 3.37          & 40.83 ± 3.64          & 44.48 ± 2.38         & 52.34 ± 7.97         & 8.33 ± 0.74          & 11.87 ± 1.57         & 12.22 ± 1.17          & 14.91 ± 1.89          \\
\rowcolor{Gray}
TransTEE                & \textbf{26.48 ± 0.27} & \textbf{32.40 ± 0.85} & \textbf{9.31 ± 1.94} & \textbf{9.85 ± 1.99} & \textbf{7.88 ± 3.84} & \textbf{8.90 ± 3.90} & \textbf{10.10 ± 3.72} & \textbf{11.04 ± 3.96} \\\hline
\multicolumn{9}{c}{\color{brown}{WPEHE@9}}                                                                                                                                                                                         \\
Zero                    & 62.52 ± 0.00          & 64.61 ± 0.00          & 14.66 ± 0.00         & 14.20 ± 0.00         & 14.61 ± 0.00         & 14.14 ± 0.00         & 14.66 ± 0.00          & 14.20 ± 0.00          \\
GNN                     & 34.13 ± 0.04          & 36.48 ± 0.04          & 11.26 ± 7.96         & 11.13 ± 7.87         & 14.21 ± 0.24         & 14.22 ± 0.17         & 13.63 ± 0.38          & 15.92 ± 1.01          \\
GraphITE                & 34.17 ± 0.02          & 36.49 ± 0.01          & 15.60 ± 0.19         & 15.53 ± 0.28         & 14.35 ± 0.04         & 14.90 ± 0.43         & 13.28 ± 0.04          & 15.83 ± 0.28          \\
SIN                     & 36.79 ± 3.35          & 40.99 ± 5.14          & 44.47 ± 2.39         & 52.31 ± 7.97         & 8.36 ± 0.74          & 11.90 ± 1.57         & 12.40 ± 1.23          & 15.08 ± 1.80          \\
\rowcolor{Gray}
TransTEE                & \textbf{28.84 ± 0.23} & \textbf{31.40 ± 0.71} & \textbf{9.34 ± 1.94} & \textbf{9.88 ± 2.00} & \textbf{7.90 ± 3.85} & \textbf{8.94 ± 3.91} & \textbf{10.14 ± 3.73} & \textbf{11.08 ± 3.97} \\\hline
\multicolumn{9}{c}{\color{brown}{WPEHE@10}}                                                                                                                                                                                        \\
Zero                    & 62.65 ± 0.00          & 65.59 ± 0.00          & 14.69 ± 0.00         & 14.23 ± 0.00         & 14.69 ± 0.00         & 14.23 ± 0.00         & 14.69 ± 0.00          & 14.23 ± 0.00          \\
GNN                     & 34.26 ± 0.04          & 37.65 ± 0.04          & 11.28 ± 7.98         & 11.16 ± 7.89         & 14.29 ± 0.22         & 14.32 ± 0.18         & 13.66 ± 0.38          & 15.96 ± 1.01          \\
GraphITE                & 34.30 ± 0.02          & 37.66 ± 0.00          & 15.64 ± 0.19         & 15.56 ± 0.28         & 14.38 ± 0.04         & 14.93 ± 0.43         & 13.31 ± 0.04          & 15.87 ± 0.27          \\
SIN                     & 37.08 ± 3.35          & 41.79 ± 5.21          & 44.49 ± 2.40         & 52.28 ± 7.96         & 8.39 ± 0.74          & 11.92 ± 1.58         & 12.49 ± 1.22          & 15.13 ± 1.81          \\
\rowcolor{Gray}
TransTEE                & \textbf{28.89 ± 0.19} & \textbf{32.25 ± 0.69} & \textbf{9.36 ± 1.93} & \textbf{9.90 ± 2.00} & \textbf{7.94 ± 3.87} & \textbf{8.95 ± 3.92} & \textbf{10.16 ± 3.74} & \textbf{11.10 ± 3.98} \\ \bottomrule
\end{tabular}}
\end{table*}

\begin{table*}[h]
\caption{\textbf{Top-$10$ samples with the maximal and minimal ATE for the effect of \textbf{Gender}.} Perturbation words in factual sentences and counterfactual sentences are colored by {\color{orange}Orange} and  {\color{magenta}Magenta} respectively.}
\label{samples_gender}
\centering 
\small
\resizebox{\textwidth}{!}{
\begin{tabular}{@{}cclc@{}}
\toprule
\multicolumn{4}{c}{Sentences with The Maximal ATEs} \\ \midrule
 &
  \textbf{Index} &
  \textbf{Sentence} &
  \textbf{ATE} \\ \midrule
 &
  1 &
  It was totally unexpected, but {\color{orange}Roger} made me feel pessimistic. &
  0.6393 \\
 &
  2 &
  We went to the restaurant, and {\color{orange}Alphonse} made me feel frustration. &
  0.578 \\
 &
  3 &
  It was totally unexpected, but {\color{orange}Amanda} made me feel pessimistic. &
  0.5109 \\
 &
  4 &
  We went to the university, and my {\color{orange}husband} made me feel angst. &
  0.4538 \\
 &
  5 &
  It is far from over, but so far i made {\color{orange}Jasmine} feel frustration. &
  0.4366 \\
 &
  6 &
  We were told that {\color{orange}Torrance} found himself in a consternation situation. &
  0.4203 \\
 &
  7 &
  We went to the university, and my {\color{orange}son} made me feel revulsion. &
  0.399 \\
 &
  8 &
  To our amazement, the conversation with my {\color{orange}aunt} was dejected. &
  0.3952 \\
 &
  9 &
  To our amazement, the conversation with my {\color{orange}aunt} was dejected. &
   0.3952\\
\multirow{-10}{*}{Factual} &
  10 &
  We went to the supermarket, and {\color{orange}Roger} made me feel uneasiness. &
  0.3752 \\ \midrule
 &
  1 &
  It was totally unexpected, but {\color{magenta}Amanda} made me feel pessimistic. &
  0.6393 \\
 &
  2 &
  We went to the school, and {\color{magenta}Latisha} made me feel frustration. &
  0.578 \\
 &
  3 &
  It was totally unexpected, but {\color{magenta}Roger} made me feel pessimistic. &
  0.5109 \\
 &
  4 &
  We went to the market, and my {\color{magenta}daughter} made me feel angst. &
  0.4538 \\
 &
  5 &
  It is far from over, but so far i made {\color{magenta}Jamel} feel frustration. &
  0.4366 \\
 &
  6 &
  We were told that {\color{magenta}Tia} found herself in a consternation situation. &
  0.4203 \\
 &
  7 &
  We went to the hairdresser, and my {\color{magenta}sister} made me feel revulsion. &
  0.399 \\
 &
  8 &
  To our amazement, the conversation with my {\color{magenta}uncle} was dejected. &
  0.3952 \\
 &
  9 &
  To our amazement, the conversation with my {\color{magenta}uncle} was dejected. &
  0.3952 \\
\multirow{-10}{*}{Counterfactual} &
  10 &
  We went to the university, and {\color{magenta}Amanda} made me feel uneasiness.&
  0.3752 \\ \bottomrule
\multicolumn{4}{c}{Sentences with The Minimal ATEs} \\ \midrule
 &
  \textbf{Index} &
  \textbf{Sentence} &
  \textbf{ATE} \\ \midrule
 &
  1 &
  \makecell[l]{
  To our amazement, the conversation with {\color{orange}Jack} was irritating, \\ no added information is given in this part.} &
  0 \\
 &
  2 &\makecell[l]{
  To our surprise, my {\color{orange}husband} found himself in a vexing situation, \\ this is only here to confuse the classifier.} &
  0 \\
 &
  3 &
  \makecell[l]{The conversation with {\color{orange}Amanda} was irritating, we could from simply looking, \\ this is only here to confuse the classifier.} &
  0 \\
 &
  4 &
  \makecell[l]{this is only here to confuse the classifier, The situation makes {\color{orange}Torrance} feel irate, \\ but it does not matter now.} &
  0 \\
 &
  5 &
  this is random noise, I made {\color{orange}Alphonse} feel irate, time and time again. &
  0 \\
 &
  6 &
  \makecell[l]{We were told that {\color{orange}Roger} found himself in a irritating situation, \\ no added information is given in this part.} &
  0 \\
 &
  7 &
 \makecell[l]{ {\color{orange}Amanda} made me feel irate whenever I came near, \\ no added information is given in this part.} &
  0 \\
 &
  8 &
  \makecell[l]{While unsurprising, the conversation with my {\color{orange}uncle} was outrageous, \\this is only here to confuse the classifier.} &
  0 \\
 &
  9 &
  It is a mystery to me, but it seems i made {\color{orange}Darnell} feel irate. &
  0 \\
\multirow{-10}{*}{Factual} &
  10 &
  \makecell[l]{The conversation with {\color{orange}Melanie} was irritating, you could feel it in the air, \\ no added information is given in this part.} &
  0 \\\midrule
 &
  1 &
  \makecell[l]{To our amazement, the conversation with {\color{magenta}Kristin} was irritating,\\ no added information is given in this part.} &
  0 \\
 &
  2 &
  \makecell[l]{To our surprise, this {\color{magenta}girl} found herself in a vexing situation, \\ this is only here to confuse the classifier.} &
  0 \\
 &
  3 &
  \makecell[l]{The conversation with {\color{magenta}Frank} was irritating, we could from simply looking, \\this is only here to confuse the classifier.} &
  0 \\
 &
  4 &
  \makecell[l]{this is only here to confuse the classifier, The situation makes {\color{magenta}Shaniqua} feel irate, \\ but it does not matter now.} &
  0 \\
 &
  5 &
  this is random noise, I made {\color{magenta}Nichelle} feel irate, time and time again. &
  0 \\
 &
  6 &
  \makecell[l]{We were told that {\color{magenta}Melanie} found herself in a irritating situation, \\no added information is given in this part.} &
  0 \\
 &
  7 &
  \makecell[l]{{\color{magenta}Justin} made me feel irate whenever I came near,\\ no added information is given in this part.} &
  0 \\
 &
  8 &
  \makecell[l]{While unsurprising, the conversation with my {\color{magenta}mother} was outrageous, \\ this is only here to confuse the classifier.} &
  0 \\
 &
  9 &
  It is a mystery to me, but it seems i made {\color{magenta}Lakisha} feel irate. &
  0 \\
\multirow{-10}{*}{Counterfactual} &
  10 &
  \makecell[l]{The conversation with {\color{magenta}Ryan} was irritating, you could feel it in the air, \\ no added information is given in this part.} &
  0 \\ \bottomrule
\end{tabular} }

\end{table*}

\begin{table*}[]
\caption{\textbf{Top-$10$ samples with the maximal and minimal ATE for the effect of \textbf{Race}.} Perturbation words in factual sentences and counterfactual sentences are colored by {\color{orange}Orange} and  {\color{magenta}Magenta} respectively.}
\label{samples_race}
\centering
\small
\resizebox{\textwidth}{!}{ 
\begin{tabular}{@{}cclc@{}}
\toprule
\multicolumn{4}{c}{Sentences with The Maximal ATEs} \\ \midrule
 &
  \textbf{Index} &
  \textbf{Sentence} &
  \textbf{ATE} \\ \midrule
 &
  1 &
   \makecell[l]{sometimes noise helps, not here, The conversation with {\color{orange}Shereen} was cry,\\ we could from simply looking.} &
   0.9976 \\
 &
  2 &
  {\color{orange}Darnell} made me feel uneasiness for the first time ever in my life.                          & 0.6853  \\
 &
  3 &
  {\color{orange}Alonzo} feels pity as he paces along to the shop.                                                 & 0.6563 \\
 &
  4 &
  {\color{orange}Adam} feels despair as he paces along to the school.                                              & 0.6066 \\
 &
  5 &
  {\color{orange}Ebony} made me feel unease for the first time ever in my life.
                                                                             & 0.592  \\
 &
  6 &
  {\color{orange}Nancy} made me feel dismay for the first time ever in my life. &
  0.548  \\
 &
  7 &
  {\color{orange}Lamar} made me feel revulsion for the first time ever in my life.
                                & 0.5074  \\
 &
  8 &
  {\color{orange}Alonzo} made me feel revulsion for the first time ever in my life.                                & 0.4911  \\
 &
  9 &
  While we were walking to the market, {\color{orange}Josh} told us all about the recent pessimistic events.       & 0.4886 \\
\multirow{-10}{*}{Factual} &
  10 &
  {\color{orange}Alonzo} made me feel unease for the first time ever in my life.                                   & 0.4877 \\ \midrule
 &
  1 &
   \makecell[l]{sometimes noise helps, not here, The conversation with {\color{magenta}Katie} was cry,\\ we could from simply looking.} &
  0.9976 \\
 &
  2 &
  {\color{magenta}Josh} made me feel uneasiness for the first time ever in my life.
 &
  0.6853 \\
 &
  3 &
  {\color{magenta}Josh} feels pity as he paces along to the shop.
 &
  0.6563 \\
 &
  4 &
   {\color{magenta}Terrence} feels despair as he paces along to the hairdresser.
 &
  0.6066 \\
 &
  5 &
  {\color{magenta}Ellen} made me feel unease for the first time ever in my life.
 &
  0.592 \\
 &
  6 &
 {\color{magenta}Latisha} made me feel dismay for the first time ever in my life. &
  0.548 \\
 &
  7 &
   {\color{magenta}Jack} revulsione me feel revulsion for the first time ever in my life. &
  0.5074 \\
 &
  8 &
 {\color{magenta} Frank} made me feel revulsion for the first time ever in my life.
 &
  0.4911 \\
 &
  9 &
 While we were walking to the college, {\color{magenta}Torrance} told us all about the recent pessimistic events.
 &
  0.4886 \\
\multirow{-10}{*}{Counterfactual} &
  10 &
   {\color{magenta}Roger} made me feel unease for the first time ever in my life.
&
  0.4877 \\ \bottomrule
\multicolumn{4}{c}{Sentences with The Minimal ATEs} \\ \midrule
 &
  \textbf{Index} &
  \textbf{Sentence} &
  \textbf{ATE} \\ \midrule
 &
  1 &
  We went to the bookstore, and {\color{orange}Alonzo} made me feel fearful, really, there is no information here. &
  0 \\
 &
  2 &nothing here is relevant, I made {\color{orange}Jack} feel angry, time and time again.
 &
  0 \\
 &
  3 &
  do not look here, it will just confuse you, {\color{orange}Jamel} feels fearful at the start.
 &
  0 \\
 &
  4 &
  We went to the bookstore, and {\color{orange}Justin} made me feel irritated.
 &
  0 \\
 &
  5 &
 As he approaches the restaurant, {\color{orange}Justin} feels irritated. &
  0 \\
 &
  6 &
Now that it is all over, {\color{orange}Andrew} feels irritated.

 &
  0 \\
 &
  7 &
 do not look here, it will just confuse you, {\color{orange}Ebony} feels fearful at the start. &
  0 \\
 &
  8 &
  do not look here, it will just confuse you, {\color{orange}Lakisha} feels fearful at the start.
 &
  0 \\
 &
  9 &
  \makecell[l]{There is still a long way to go, but the situation makes {\color{orange}Lakisha} feel irritated,\\ this is only here to confuse the classifier.}
 &
  0 \\
\multirow{-10}{*}{Factual} &
  10 &
  I have no idea how or why, but i made {\color{orange}Alan} feel irritated.
 &
  0 \\\midrule
 &
  1 &
  We went to the market, and {\color{magenta}Roger} made me feel fearful, really, there is no information here.
 &
  0 \\
 &
  2 &
  nothing here is relevant, I made {\color{magenta}Jamel} feel angry, time and time again.
 &
  0 \\
 &
  3 &
  do not look here, it will just confuse you, {\color{magenta}Harry} feels fearful at the start.
 &
  0 \\
 &
  4 &
  We went to the church, and {\color{magenta}Lamar} made me feel irritated.
 &
  0 \\
 &
  5 &
   As he approaches the shop, {\color{magenta}Malik} feels irritated. &
  0 \\
 &
  6 &
  Now that it is all over, {\color{magenta}Torrance} feels irritated.
 &
  0 \\
 &
  7 &
  do not look here, it will just confuse you, {\color{magenta}Amanda} feels fearful at the start.
 &
  0 \\
 &
  8 &
  do not look here, it will just confuse you, {\color{magenta}Amanda} feels fearful at the start.
 &
  0 \\
 &
  9 &
  \makecell[l]{There is still a long way to go, but the situation makes {\color{magenta}Katie} feel irritated, \\ this is only here to confuse the classifier.}
 &
  0 \\
\multirow{-10}{*}{Counterfactual} &
  10 &
  I have no idea how or why, but i made {\color{magenta}Darnell} feel irritated.
 &
  0 \\ \bottomrule
\end{tabular}}

\end{table*}

\section{Remarks on Interpretability}
It is fundamentally hard to evaluate the interpretability even for supervised learners, as the evaluation crucially depends on specific models, tasks, and input spaces \citep{jacovi2020towards}. 
TransTEE provides an initial step to promote causal inference model interpretability. We can see from the experimental results in fig. 4(a), 4(b), and fig. 10 that TransTEE assigns more weights to confounders as opposed to other covariates, which is a new observation that previous backbones are hard to achieve. 
We see that explaining causal inference models in this way - using the feature importance scores for each covariate can be used for benchmarking treatment effect estimators \citep{crabbe2022benchmarking}.

%% file: arxiv/sections/05_propensity.tex
We first discuss the fundamental differences and common goals between our algorithm and traditional ones: as a general approach to causal inference, \abbr~can be directly harnessed with traditional methods that estimate propensity scores by including hand-crafted features of covariates \citep{imbens2015causal} to reduce biases through covariate adjustment \citep{austin2011introduction}, matching \citep{rubin1996matching, abadie2016matching}, stratification \citep{frangakis2002principal}, reweighting \citep{hirano2003efficient}, g-computation \citep{imbens2015causal}, sub-classification \citep{rosenbaum1984reducing}, covariate adjustment \citep{austin2011introduction}, targeted regularization \citep{van2006targeted} or conditional density estimation \citep{nie2021vcnet} that create quasi-randomized experiments \citep{d1998propensity}. 
It is because the general framework provides an advantage to using an off-the-shelf propensity score regularizer for balancing covariate representations. 
Similar to the goal of traditional methods like inverse probability weighting and propensity score matching \citep{austin2011introduction}, which seeks to weigh a single observation to mimic the randomization effects with respect to the covariate from different treatment groups of interest. 

Unlike previous works that use hand-crafted features 
or directly model the conditional density 
via maximum likelihood training, 
which is prone to high variance 
when handling high-dimensional, 
structured treatments \citep{singh2019kernel} 
and can be problematic when we want to estimate 
a plausible propensity score 
from the generative model \citep{mohamed2016learning} 
(see the degraded performance of MLE in Table \ref{tab:continuous1}),
\abbr~learns a propensity score network $\pi_\phi(t|\x)$ 
via minimax bilevel optimization. 
The motivations for adversarial training 
between $\mu_\theta(\x,t)$ and $\pi_\phi(t|\x)$ are three-fold: 
(i) it enforces the independence
between treatment and covariate representations 
as shown in Proposition \ref{prop},
which serves as algorithmic randomization
in place of costly randomized controlled trials 
\citep{rubin2007design} for overcoming selection bias
\citep{d1998propensity, imbens2015causal};
(ii) it explicitly models the propensity $\pi_\phi(t|\x)$ 
to refine treatment representations 
and promote covariate adjustment \citep{kaddour2021causal}; 
and (iii) taking an adversarial domain adaptation perspective,
the methodology is effective for learning invariant representations 
and also regularizes $\mu_\theta(\x,t)$ 
to be invariant to nuisance factors 
and may perform better empirically
on some classes of distribution shifts
\citep{ganin2016domain, shalit2017tarnet, zhao2018adversarial, johansson2020generalization, wang2020continuously}. 
 
Based on the above discussion, 
when treatments are discrete, 
one might consider directly applying heuristic 
methods such as adversarial domain adaptation 
(see \citep{ganin2016domain, zhao2018adversarial} for algorithmic development guidelines).
We note the heuristic nature of domain-adversarial methods
(see \citep{wu2019domain} for clear failure cases) 
and a debunking of the common claim 
that \citep{ben2010theory} guarantees 
the robustness of such methods.
Here, we focus on continuous TEE,
a more general and challenging scenario,
where we want to estimate ADRF, 
and propose two variants of $\mathcal{L}_{\phi}$ 
as an adversary for the outcome regression objective 
$\mathcal{L}_{\theta}$ in Eq. \ref{eq:outcome} accordingly. 
The process is shown in Eq. \ref{eq:minimax} below:
\begin{equation}
    \min_\theta \max_\phi \mathcal{L}_{\theta}(\x, y, t) - \mathcal{L}_{\phi}(\x, t).
    \label{eq:minimax}
\end{equation}
We refer to the above minimax game for algorithmic randomization in place of costly randomized controlled trials. Such algorithmic randomization based on neural representations using propensity score
creates subgroups of different treated units 
as if they had been randomly assigned to different treatments 
such that conditional independence $ T \indep X \  | \ \pi(T|X) $ 
is enforced across strata and continuation,
which approximates a random block experiment
to the observed covariates \citep{imbens2015causal}.

Below, we introduce two variants of $\mathcal{L}_{\phi}(\mathbf{x}, t)$:

\textbf{Treatment Regularization (TR)}
is a standard MSE over the treatment space 
given the predicted treatment $\hat{t}_i$
and the ground truth $t_i$
\begin{equation}
    \mathcal{L}_{\phi}^{TR}(\x, t) = \sum_{i=1}^n\left(t_i - \pi_\phi(\hat{t}_i|\x_i)\right)^2.
\end{equation}
TR is explicitly matching the mean of the propensity score to that of the treatment.
In an ideal case, $\pi(t|\x)$ should be uniformly distributed given different $\x$. However, the above treatment regularization procedure
only provides matching for the mean of the propensity score,
which can be prone to bad equilibriums and treatment misalignment \citep{wang2020continuously}. Thus, we introduce the distribution of $t$ and model the uncertainty rather than predicting a scalar $t$:

\textbf{Probabilistic Treatment Regularization (PTR)} is a probabilistic version of TR that models the mean $\mu$ (with a slight abuse of notation) and the variance $\sigma^2$ of the estimated treatment $\hat{t}_i$
\begin{equation}
    \mathcal{L}_{\phi}^{PTR} = \sum_{i=1}^n \left[\frac{\left(t_i - \pi_\phi(\mu|\x_i)\right)^2}{2\pi_\phi(\sigma^2|\x_i)} + \frac{1}{2} \log \pi_\phi(\sigma^2|\x_i) \right].
\end{equation}
The PTR matches the whole distribution, i.e. both the mean and variance, of the propensity score to that of the treatment, which can be preferable in certain cases. 

\begin{table*}[t]
\caption{\textbf{Performance of individualized treatment-dose response estimation} on the TCGA (D) dataset with different numbers of treatments. We report AMSE and standard deviation over 30 repeats. The selection bias on treatment and dosage are both set to be $2.0$.}
\label{tab:dosage}
\vskip 0.15in
\begin{center}
\begin{scriptsize}
\begin{sc}
\resizebox{.95\textwidth}{!}{
\setlength{\tabcolsep}{4pt}
\renewcommand{\arraystretch}{1.05}
\begin{tabular}{@{}ccccccc@{}}
\toprule
\multirow{2}{*}{Methods} & \multicolumn{2}{c}{\#Treatment=1}                   & \multicolumn{2}{c}{\#Treatment=2}                   & \multicolumn{2}{c}{\#Treatment=3}                   \\  \cmidrule(r){2-7}
& In-Sample                & Out-Sample               & In-Sample                & Out-Sample               & In-Sample                & Out-Sample               \\ \cmidrule(r){1-7}
SCIGAN                   & 5.6966 ± 0.0000          & 5.6546 ± 0.0000          & 2.0924 ± 0.0000          & 2.3067 ± 0.0000          & 4.3183 ± 0.0000          & 4.6231 ± 0.0000          \\
TarNet(D)                & 0.7888 ± 0.0609          & 0.7908 ± 0.0606          & 1.4207 ± 0.0784          & 1.4206 ± 0.0777          & 3.1982 ± 0.5847          & 3.1920 ± 0.5746          \\
DRNet(D)                 & 0.8034 ± 0.0469          & 0.8052 ± 0.0466          & 1.3739 ± 0.0858          & 1.3738 ± 0.0853          & 2.8632 ± 0.4227          & 2.8558 ± 0.4143          \\

VCNET(D)                 & 0.1566 ± 0.0303          & 0.1579 ± 0.0301          & 0.2919 ± 0.0743          & 0.2918 ± 0.0737          & 0.6459 ± 0.1387          & 0.6493 ± 0.1397          \\
\rowcolor{Gray}
TransTEE                 & {0.0573 ± 0.0361} & {0.0585 ± 0.0358} & \textbf{0.0550 ± 0.0137} & \textbf{0.0556 ± 0.0129} & 0.2803 ± 0.0658 & 0.2768 ± 0.0639 \\ 
\rowcolor{Gray}
TransTEE + TR                 & 0.0495 ± 0.0176 & 0.0509 ± 0.0180 & 0.0663 ± 0.0268 & 0.0671 ± 0.0268 & \textbf{0.2618 ± 0.0737} & \textbf{0.2577 ± 0.0726} \\
\rowcolor{Gray}
TransTEE + PTR                 & \textbf{0.0343 ± 0.0096} & \textbf{0.0355 ± 0.0094} & 0.0679 ± 0.0252 & 0.0686 ± 0.0252 & 0.2645 ± 0.0702 & 0.2597 ± 0.0675 \\ 
\bottomrule
\end{tabular} }
\end{sc}
\end{scriptsize}
\end{center}
%\vspace{-0.4cm}
\vspace{-0.4cm}
\end{table*}

\textbf{Equilibrium of the Minimax Game.} We analyze that TR and PTR can align the first and second moment of continuous treatments at equilibrium respectively, and thus promote the independence between treatment $t$ and covariate $\x$. 
To be clear, we denote $\mu_\theta(\x, t) := w_y \circ (\Phi_x(\x), \Phi_t(t))$ and $\pi_\phi(t|\x) := w_t \circ \Phi_x(\x)$, which decompose the predictions into featurizers $\Phi_t: \mathcal{T} \rightarrow \mathcal{Z}_T, \Phi_x: \mathcal{X} \rightarrow \mathcal{Z}_X$ and predictors $w_y: \mathcal{Z}_X \times \mathcal{Z}_T \rightarrow \mathcal{Y},  w_t: \mathcal{Z}_X \rightarrow \mathcal{T}$. For example, $\Phi_x(\x)$ and $\Phi_t(t)$ can be the linear embedding layer and attention modules in our implementation. The propensity is computed on $\Phi_x(\x)$, an intermediate feature representation of $\x$. Similarly, $\mu_\theta(\x, t)$ is computed from $\Phi_t(t)$ and $\Phi_x(\x)$. For the ease of our analysis below, we assume the predictors $w_t, w_x$ are fixed.

\begin{prop}{(The optimum of propensity score model)}
In the equilibrium of the game, assuming the outcome prediction model is fixed, then the optimum of TR is achieved when $\mathbb{E}[\Phi_t(t)|\Phi_x(\x)] = \mathbb{E}[\Phi_t(t)], \forall \ \Phi_x(\x)$ via matching the mean of propensity score $\pi(\Phi_t(t)|\Phi_x(\x)) $ and the marginal distribution $p(\Phi_x(\x))$ and the optimum discriminator of PTR is achieved via matching both the mean and variance such that $\mathbb{E}[\Phi_t(t)|\Phi_x(\x)] = \mathbb{E}[\Phi_t(t)], \mathbb{V}[\Phi_t(t)|\Phi_x(\x)] = \mathbb{V}[\Phi_t(t)], \  \forall \ \Phi_x(\x)$. 
%\Phi_t(t), \Phi_t(t), See Appendix \ref{app:proof} for the proof.
\label{prop}
\end{prop}
\label{app:proof}
\begin{proof}
The proof concerns the analysis of the equilibrium of the Minimax Game. It is a special case of \citep{wang2020continuously} when there are only two players, i.e. $\mu_\theta$ and $\pi_\phi$. We represent treatments explicitly and interpret the connections with combating selection biases. Given the fixed outcome regression model $\mu_\theta$, the optimal propensity score model $\pi^*$ is
\begin{equation}
\begin{aligned}
    \pi^* & = \argmin_\pi \mathcal{L}_\phi (\Phi_x(\x), \Phi_t(t)) \\
          & = \argmin_\pi \mathbb{E}_{(\Phi_x(\x),\Phi_t(t))\sim p(\Phi_x(\x),\Phi_t(t))}
          \left(\Phi_t(t)-\pi_{\theta}
          \left(\Phi_t(\hat{t}) | \mathbf{x}\right)\right)^{2} \\
          & = \argmin_\pi \mathbb{E}_{\Phi_x(\x) \sim p(\Phi_x(\x))} \mathbb{E}_{\Phi_t(t) \sim p(\Phi_t(t)|\Phi_x(\x))} \left(\Phi_t(t)-\pi_{\theta}\left(\Phi_t(\hat{t}) | \mathbf{x}\right)\right)^{2}.
\end{aligned}
\end{equation}
The inner minimum is achieved at $\pi^*_{\theta}\left(\Phi_t(\hat{t})| \mathbf{x}\right) = \mathbb{E}_{\Phi_t(t)\sim p(\Phi_t(t)|\Phi_x(\x))}[\Phi_t(t)]$ given the following quadratic form:
\begin{equation}
\begin{aligned}
&\mathbb{E}_{(\Phi_x(\x),\Phi_t(t)) \sim p(\Phi_x(\x),\Phi_t(t))} \left(\Phi_t(t)-\pi_{\theta}\left(\Phi_t(\hat{t})| \Phi_\x(\mathbf{x})\right)\right)^{2} = \\
&\mathbb{E}_{\Phi_t(t)\sim p(\Phi_t(t)|\Phi_x(\x))}[\Phi_t(t)^2] - 2\pi_{\theta}\left(\Phi_t(\hat{t})| \mathbf{x}\right)\mathbb{E}_{\Phi_t(t)\sim p(\Phi_t(t)|\Phi_x(\x))}[\Phi_t(t)] + \pi_{\theta}\left(\Phi_t(\hat{t})| \mathbf{x}\right)^2.
\end{aligned}
\end{equation}
We assume the above optimum condition of the propensity score model always holds with respect to the outcome regression model during training, then the minimax game in Eq. \ref{eq:minimax} can be converted to maximize the inner loop:
\begin{equation}
\begin{aligned} %\min_{\theta}
& \max _{\phi} -\mathcal{L}_{\phi}(\mathbf{x}, \Phi_t(t)) = \mathcal{L}_{\phi^*}(\Phi_x(\x),\Phi_t(t)) \\
&= \mathbb{E}_{(\Phi_x(\x),\Phi_t(t)) \sim p(\Phi_x(\x),\Phi_t(t))} \left(\Phi_t(t)-\mathbb{E}_{\Phi_t(t)\sim p(\Phi_t(t)|\Phi_x(\x))} [\Phi_t(t)]\right)^{2} \\
&= \mathbb{E}_{\Phi_x(\x)\sim p(\Phi_x(\x))} \mathbb{E}_{\Phi_t(t)\sim p(\Phi_t(t)|\Phi_x(\x)) \sim p(\Phi_x(\x),\Phi_t(t))} \left(\Phi_t(t)-\mathbb{E}_{\Phi_t(t)\sim p(\Phi_t(t)|\Phi_x(\x))}[\Phi_t(t)]\right)^{2} \\
&= \mathbb{E}_{\Phi_x(\x)\sim p(\Phi_x(\x))}\mathbb{V}_{\Phi_t(t)\sim p(\Phi_t(t)|\Phi_x(\x))}[\Phi_t(t)] = \mathbb{E}_{\Phi_x(\x)} \mathbb{V}[\Phi_t(t)|\Phi_x(\x)].
\label{eq:inner}
\end{aligned}
\end{equation}
Next we show the difference between Eq. \ref{eq:inner} and the variance of the treatment $\mathbb{V}[\Phi_t(t)]$:
\begin{equation}
\begin{aligned}
     & \mathbb{E}_{\Phi_x(\x)\sim p(\Phi_x(\x))}\mathbb{V}_{\Phi_t(t)\sim p(\Phi_t(t)|\Phi_x(\x))}[\Phi_t(t)] - \mathbb{V}[\Phi_t(t)] \\
    =& \mathbb{E}_{\Phi_x(\x)\sim p(\Phi_x(\x))}[\mathbb{E}[\Phi_t(t)^2|\Phi_x(\x)] - \mathbb{E}[\Phi_t(t)|\Phi_x(\x)]^2] - (\mathbb{E}[\Phi_t(t)^2] - \mathbb{E}[\Phi_t(t)]^2) \\
    =& \mathbb{E}[\Phi_t(t)]^2 - \mathbb{E}_{\Phi_x(\x)}[\mathbb{E}[\Phi_t(t)|\Phi_x(\x)]^2] = \mathbb{E}_{\Phi_x(\x)}[\mathbb{E}[\Phi_t(t)|\Phi_x(\x)]]^2 - \mathbb{E}_{\Phi_x(\x)}[\mathbb{E}[\Phi_t(t)|\Phi_x(\x)]^2]\\
    \leq & \mathbb{E}_{\Phi_x(\x)}[\mathbb{E}[\Phi_t(t)|\Phi_x(\x)]^2] - \mathbb{E}_{\Phi_x(\x)}[\mathbb{E}[\Phi_t(t)|\Phi_x(\x)]^2] = 0
\end{aligned}
\end{equation}
where the last inequality is determined by Jensen's inequality and the convexity of $\Phi_t(t)^2$. The optimum is achieved when $\mathbb{E}[\Phi_t(t)|\Phi_x(\x)]$ is constant w.r.t $\Phi_x(\x)$ and so $\mathbb{E}[\Phi_t(t)|\Phi_x(\x)]=\mathbb{E}[\Phi_t(t)], \ \forall \Phi_x(\x)$. 

The proof process for PTR is similar but includes the derivation of variance matching.

%\resizebox{\textwidth}{!}{
\begin{equation}
\begin{split}
    \pi^* & = \argmin_\pi \mathcal{L}_\phi (\Phi_x(\x), \Phi_t(t)) \\
          & = \argmin_\pi \mathbb{E}_{(\Phi_x(\x),\Phi_t(t))\sim p(\Phi_x(\x),\Phi_t(t))} \left( \frac{(\mathbb{E}[\Phi_t(t) | \Phi_x(\x)]-\Phi_t(t))^{2}}{2 \mathbb{V}[\Phi_t(t) | \Phi_x(\x)]} + \frac{\log \mathbb{V}[\Phi_t(t)|\Phi_x(\x)]}{2} \right) \\
          & = \argmin_\pi \mathbb{E}_{\Phi_x(\x)} \mathbb{E}_{\Phi_t(t)} 
          \bigg(\frac{(\mathbb{E}[\Phi_t(t) | \Phi_x(\x)]-\Phi_t(t))^{2}}{2 \mathbb{V}[\Phi_t(t) | \Phi_x(\x)]} + \frac{\log \mathbb{V}[\Phi_t(t)| \Phi_x(\x)]}{2} \bigg),
\end{split}
\end{equation} 

where $\mathbb{E}_{\Phi_x(\x)}$ and $\mathbb{E}_{\Phi_t(t)}$ denote $\mathbb{E}_{\Phi_x(\x) \sim p(\Phi_x(\x))}$ and $\mathbb{E}_{\Phi_t(t) \sim p(\Phi_t(t)|\Phi_x(\x))}$ respectively for brevity.
The first term can be reduced to a constant given the definition of variance:
%\resizebox{\textwidth}{!}{
\begin{equation}
\begin{aligned}
& \mathbb{E}_{\Phi_x(\x) \sim p(\Phi_x(\x))} \mathbb{E}_{\Phi_t(t) \sim p(\Phi_t(t)|\Phi_x(\x))} \left( \frac{(\mathbb{E}[\Phi_t(t) | \mathbf{x}]-\Phi_t(t))^{2}}{2 \mathbb{V}[\Phi_t(t) | \mathbf{x}]} \right) \\
& = \mathbb{E}_{\Phi_x(\x) \sim p(\Phi_x(\x))} \left( \frac{\mathbb{V}[\Phi_t(t) | \mathbf{x}]}{2\mathbb{V}[\Phi_t(t) | \mathbf{x}]} \right) = \frac{1}{2}.
\end{aligned}
\label{eq:comp1}
\end{equation}
%}

The second term can be upper bounded by using Jensen's inequality:
%\resizebox{\textwidth}{!}{
\begin{equation}
\begin{aligned}
& \mathbb{E}_{\Phi_x(\x) \sim p(\Phi_x(\x))} \mathbb{E}_{\Phi_t(t) \sim p(\Phi_t(t)|\Phi_x(\x))} \left( \frac{\log \mathbb{V}[\Phi_t(t)|\mathbf{x}]}{2}  \right) \\
& \le \frac{1}{2}\log \left( \mathbb{E}_{\Phi_x(\x) \sim p(\Phi_x(\x))}[\mathbb{V}[\Phi_t(t)|\Phi_x(\x)]] \right) \\
& \le \frac{1}{2} \log \left(\mathbb{V}[\Phi_t(t)]\right).
\end{aligned}
\label{eq:comp2}
\end{equation} %}

Combining Eq. \ref{eq:comp1} and Eq. \ref{eq:comp2}, the optimum $\frac{1}{2} + \frac{1}{2} \log \left(\mathbb{V}[\Phi_t(t)]\right)$ is achieved when $\mathbb{E}[\Phi_t(t)|\Phi_x(\x)]$, $\mathbb{V}[\Phi_t(t)|\Phi_x(\x)]$ is constant w.r.t $\Phi_x(\x)$ and so $\mathbb{E}[\Phi_t(t)|\Phi_x(\x)]=\mathbb{E}[\Phi_t(t)], \mathbb{V}[\Phi_t(t)|\Phi_x(\x)]=\mathbb{V}[\Phi_t(t)], \ \forall \Phi_x(\x)$ according to the equality conditions of the first and second inequality in Eq. \ref{eq:comp2}, respectively. 
\end{proof}